\documentclass[oldfontcommands,a4paper,11pt,twoside,toc,page]{memoir} 

\usepackage{geometry}
\usepackage{hyperref}
\usepackage[english]{babel}
\usepackage{amsfonts}
\usepackage{amsmath}
\usepackage{amssymb}
\usepackage{amsthm} 
\usepackage{thmtools}
\usepackage{mathtools}
\usepackage{booktabs}
\usepackage{wasysym}
\usepackage{xspace}
\usepackage{afterpage}
\usepackage{covington}  
\usepackage[svgnames]{xcolor}
\usepackage{graphicx}
\usepackage{latexsym}
\usepackage{pgf}
\usepackage{tikz}
\usepackage{pgfplots}
\usepackage{pgfplotstable}
\usepackage{natbib}
\usepackage[]{appendix}
\usetikzlibrary{arrows,automata}
\usepgfplotslibrary{external}
\pgfplotsset{compat=newest}
\usepackage{csvsimple}
\usepackage{kbordermatrix}
\usepackage{tkzkiviat}
\usepackage{framenet}
\usepackage{float}
\usepackage{url}
\usepackage{numprint}
\usepackage{enumerate}
\usepackage{paralist}
\usepackage{subcaption}
\usepackage{minitoc}
\usepackage{array}
\usepackage{textcomp} 
\ifxetex
  \usepackage{fontspec}
\fi

\usepackage[T1]{fontenc}
\usepackage[libertine,cmintegrals,cmbraces,vvarbb]{newtxmath}
\usepackage[scaled=0.95]{inconsolata}

\usepackage{color}
\usepackage{calc}
\usepackage[scaled=.92]{helvet}

\usepackage{rotating}
\usepackage{multirow}

\usepackage[toc,section=chapter,nopostdot,xindy]{glossaries}
\makeglossaries

\usepackage{makeidx}
\makeindex

\newcommand{\romNum}[1]{\uppercase\expandafter{\romannumeral #1\relax}}

\newcommand{\smallurl}[1]{{\footnotesize\url{#1}}}

\renewcommand{\paragraph}[1]{\vspace{2EM} \noindent \small \textbf{#1}\normalsize}

\definecolor{CBb}{HTML}{56B4E9}
\definecolor{CBdb}{HTML}{0072B2}
\definecolor{CBg}{HTML}{009E73}
\definecolor{CBo}{HTML}{D55E00}

\definecolor{HL1}{HTML}{6B944D}
\definecolor{HL2}{HTML}{966BAF}
\definecolor{HL3}{HTML}{BC5A47}
\newcommand\myshade{20}
\colorlet{mylinkcolor}{HL1}
\colorlet{mycitecolor}{HL2}
\colorlet{myurlcolor}{HL3}
\hypersetup{
  linkcolor  = mylinkcolor!\myshade!black,
  citecolor  = mycitecolor!\myshade!black,
  urlcolor   = myurlcolor!\myshade!black,
  colorlinks = true,
  pdfauthor = {Alexandre Kabbach},
  unicode = true,
}

\colorlet{lightred}{red!60!}
\colorlet{lightgreen}{green!60!}
\colorlet{lightorange}{orange!70!}

\newsavebox{\ChpNumBox}
\definecolor{ChapColor}{rgb}{0.80,0.00,0.00}
\makeatletter
\newcommand*{\thickhrulefill}{%
\leavevmode\leaders\hrule height 1\p@ \hfill \kern \z@}
\newcommand*\BuildChpNum[2]{%
    \begin{tabular}[t]{@{}c@{}}
        \makebox[0pt][c]{#1\strut} \\[.5ex]
        \colorbox{ChapColor}{%
            \rule[-10em]{0pt}{0pt}%
            \rule{2.5ex}{0pt}\color{white}#2\strut
        \rule{1ex}{0pt}}%
\end{tabular}}
\makechapterstyle{BlueBox}{%
    \renewcommand{\chapnamefont}{\large\scshape}
    \renewcommand{\chapnumfont}{\Huge\bfseries}
    
    \setlength{\beforechapskip}{20pt}
    \setlength{\midchapskip}{26pt}
    \setlength{\afterchapskip}{40pt}

    \renewcommand{\printchapternum}{%
        \sbox{\ChpNumBox}{%
            \BuildChpNum{\chapnamefont\@chapapp}%
    {\chapnumfont\thechapter}}}
    \renewcommand{\printchapternonum}{%
        \sbox{\ChpNumBox}{%
            \BuildChpNum{\chapnamefont\vphantom{\@chapapp}}%
    {\chapnumfont\hphantom{\thechapter}}}}

}

\makeatletter

\makechapterstyle{VZ14}{

\renewcommand\thickhrulefill{\leavevmode \leaders \hrule height 1ex \hfill \kern \z@}

\renewcommand\chapnamefont{\Large\scshape}
\renewcommand\printchapternum{%
\chapnamefont\null\thickhrulefill\quad
\@chapapp\space\thechapter\quad\thickhrulefill}
\renewcommand\printchapternonum{%
\par\thickhrulefill\par\vskip\midchapskip
\hrule\vskip\midchapskip
}

}

\addto\captionsenglish{
  \renewcommand{\contentsname}%
    {Table of Contents}%
}

\newcommand{\semafor}{\textsc{Semafor}\xspace}
\newcommand{\rofames}{\textsc{Rofames}\xspace}
\newcommand{\noframenet}{\textsc{NoFrameNet}\xspace}
\newcommand{\valencer}{\textsc{Valencer}\xspace}
\newcommand{\myvalencer}{\textsc{myValencer}\xspace}
\newcommand{\pfn}{\texttt{pFN}\xspace}
\newcommand{\pyfn}{\texttt{pyFN}\xspace}
\newcommand{\dfn}{\texttt{dFN}\xspace}

\newcommand{\sem}{\texttt{SEM}\xspace}
\newcommand{\acl}{\texttt{ACL}\xspace}
\newcommand{\febar}{\texttt{febar}\xspace}

\newcommand\blankpage{%
    \null
    \thispagestyle{empty}%
    \addtocounter{page}{-1}%
    \newpage}

\makeatother

\chapterstyle{BlueBox} 

\setsecnumdepth{subsubsection}
\mtcselectlanguage{french}
\dominitoc

\selectfont 

\DeclareMathOperator*{\argmin}{arg\,min}

\makepagestyle{myruled}
\makeheadrule{myruled}{\textwidth}{\normalrulethickness}
\makefootrule{myruled}{\textwidth}{\normalrulethickness}{\footruleskip}
\makeevenhead{myruled}{}{\textsc{\leftmark}}{}
\makeoddhead{myruled}{}{\textsc{\rightmark}}{}
\makeevenfoot{myruled}{\normalcolor\thepage}{}{}
\makeoddfoot{myruled}{}{}{\normalcolor\thepage}
\makeatletter 
\makepsmarks{myruled}{
\nouppercaseheads
}
\makeatother

\title{
    Debugging Frame Semantic Role Labeling\\
    \vspace{0.5em}
    {\Large \emph{Towards robust error analysis of statistical models for automatic frame semantic structure extraction}}
}
\author{Alexandre Kabbach}
\date{2017}

\begin{document}

\newgeometry{body={160mm,250mm}, bottom=20mm, right=25mm, left=25mm,top=20mm, headheight=0mm, headsep=0mm, marginparsep=0mm, marginparwidth=0mm}
\makeatletter
\thispagestyle{empty}

\begin{figure}[h!tpb]
 \centering
 \includegraphics{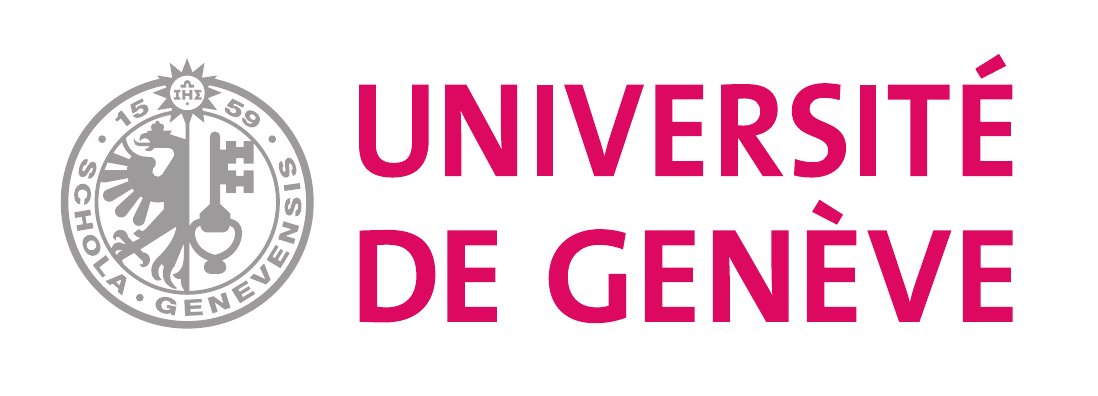}
\end{figure}
\begin{vplace}
 \begin{center}

  \vspace{2\baselineskip}

  {\huge\@title}\par

  \vspace{5\baselineskip}

  {\huge\bfseries\@author}\par

  \vspace{2\baselineskip}

  {\Large Memoir for the Certificate of Specialization in Linguistics}\par

  \vspace{6\baselineskip}

  {\Large University of Geneva}\par
  \vspace{1\baselineskip}
  {\Large Faculty of Humanities}\par
  \vspace{1\baselineskip}
  {\Large Department of Linguistics}\par

  \vspace{2\baselineskip}

  \vspace{2em}

  {\large December 2017}\par
 \end{center}
\end{vplace}
\afterpage{\blankpage}
\setcounter{page}{0}
\makeatother
\restoregeometry
\clearpage
\pagestyle{myruled}
\section*{Abstract}
We propose a quantitative and qualitative analysis of the performances
of statistical models for frame semantic structure extraction.
We report on a replication study on FrameNet 1.7 data and show that
preprocessing toolkits play a major role in argument identification
performances, observing gains similar in their order of magnitude to those
reported by recent models for frame semantic parsing.
We report on the robustness of a recent statistical classifier for
frame semantic parsing to lexical configurations of predicate-argument structures,
relying on an artificially augmented dataset generated using a rule-based
algorithm combining valence pattern matching and lexical substitution.
We prove that syntactic preprocessing plays a major role in the performances
of statistical classifiers to argument identification, and discuss
the core reasons of syntactic mismatch between dependency parsers output
and FrameNet syntactic formalism. Finally, we suggest new leads for
improving statistical models for frame semantic parsing, including
joint syntax-semantic parsing relying on FrameNet syntactic formalism,
latent classes inference via split-and-merge algorithms and neural network
architectures relying on rich input representations of words.

\clearpage
\tableofcontents
\clearpage
\listoffigures
\mtcaddchapter
\clearpage
\listoftables
\mtcaddchapter
\mainmatter
\clearpage
\pagenumbering{arabic}
\chapter{Introduction}

Frame semantics
\citep{fillmore1982}
is an approach to the study of lexical meaning where schematic representations
of events, relations or entities called \emph{frames} provide semantic
background for interpreting the meaning of words and illustrating their
syntactic valence. Its implementation for English through the FrameNet lexicon
has proven greatly beneficial for the field of natural language processing
over the past twenty years.
Indeed, FrameNet \citep{baker1998berkeley}
provides a fine-grained classification of predicate-argument structures in English,
relying on an exhaustive taxonomy of inter-connected semantic classes and roles,
which is particularly suited for drawing useful semantic inferences from text while
abstracting away from surface syntactic structures.
FrameNet has been used in tasks as diverse as
paraphrase recognition \citep{pado2005cause}, question answering \citep{shen2007using},
stock price movements prediction \citep{xie2013}, student essay clarity modeling
\citep{persing2013}, knowledge extraction from Twitter \citep{sogaard2015using}
and event detection \citep{liuetal2016,spilioetal2017}.

In order to make use of FrameNet annotation for the tasks at hand,
the aforementioned systems
have typically relied on statistical models to perform
automatic extraction of frame semantic structures
\textendash\ a task also referred to as \emph{frame semantic parsing} \textendash\
so as to project
frame semantic annotation on custom data for which no FrameNet annotation was available.
However, the low
performances of frame semantic parsers, especially in out-of-domain settings
\citep{croceetal2010,hartmannetal2017}, have persistently limited the
exploitation of FrameNet's full potential.

Past research~\citep{dasetal2014,fitzgerald2015,hartmann2016} have argued
that frame semantic parsing suffers from the sparsity of FrameNet annotated data.
Yet, previous attempts at augmenting training data, either manually
~\citep{kshirsagaretal2015}, or through automatic generation
via annotation projection ~\citep{furstenau2012} and transfer from linked
lexical resources
~\citep{hartmann2016} have resulted in limited improvements.
The limited improvements of \cite{kshirsagaretal2015}
are particularly worrisome, as they relied on previously unused
gold FrameNet annotated data, which augmented training data by an order of
magnitude. It is all the more concerning that they did try to
compensate for out-of-domainness effects
\textendash\
the fact that data used to evaluate a statistical model may contain
patterns absent from the training data used to generate the model
\textendash\
following remarks of \cite{dasetal2014}, but to no avail.
Two questions arise from those observations:
\begin{enumerate}
 \item Are prediction failures due to a
       \emph{lack of annotation}, which translates to the absence in the
       training data of linguistic patterns necessary for \emph{any}
       machine learning model to properly predict the frame semantic
       structures included in the evaluation data?
 \item Does failure to capture structural generalization
       principles necessary to correctly predict frame semantic structures
       lie with the probabilistic models used so far?
\end{enumerate}
%
Past results already offer preliminary leads which can guide reflection:
several research have successfully attempted to incorporate features relating
to information hard-coded in FrameNet \textendash\ such as frame relations
\citep{kshirsagaretal2015},
or frame element semantic types \citep{rothlapata2015} \textendash\
to statistical models
for frame semantic parsing, suggesting that improvements can still be made
on the modeling side, as current statistical models fail to capture
structural information available in FrameNet.
It is therefore more than ever crucial to understand exactly \emph{what} statistical
models learn, and
\emph{what} they \emph{fail} to learn. In addition, it is important
to understand whether statistical models'
failures to capture structural generalizations correlate with properties of
language and linguistic phenomena at large, or whether they lie in
engineering approximations and optimization constraints. The question
of what statistical models do learn in the end is all the more relevant today
as recent approaches to frame semantic parsing have relied on neural networks
\citep{fitzgerald2015,roth2016,swayamdipta2017,yangmitchell2017}, which are
notoriously hard to debug. Careful
error analysis are still lacking to understand \emph{qualitatively} rather than
\emph{quantitatively} what those models have contributed to and why they
do seem to learn better than previous models.

Although it is likely true that statistical models for frame semantic
parsing can still be improved,
it is evenly likely that the performances of those models remain
bounded by domain-specific limitations and annotation redundancies
\citep{croceetal2010,hartmannetal2017}.
Redundancies are particularly problematic given
the cost of FrameNet annotation which requires extensive
linguistic expertise and time.
If such redundancies do exist, they should
be clearly identified to prevent annotators from wasting time and energy.
Given those considerations, the purpose of this memoir is threefold:
\begin{enumerate}
 \item Demonstrate our understanding of frame semantics theory, FrameNet and
       the literature on frame semantic parsing;
 \item Formalize the challenges posed by frame semantic parsing,
       replicate past results and rationalize experimental setups;
 \item Report on the preliminary results of a model used
       to artificially augment FrameNet annotated data in order to ultimately
       perform qualitative analysis of
       statistical models for frame semantic parsing.
\end{enumerate}

More specifically, we propose a model to artificially augment FrameNet annotated
data
with paraphrastic examples generated via a rule-based algorithm combining
lexical substitution and
valence pattern\footnote{defined as a specific combination of syntactic
 realizations of the arguments of a predicate, see Section~\ref{framenet}}
matching.
For example, given
the sentence \emph{John \target{gave} an apple to Mary},
our system could generate the sentence
\emph{John \target{handed out} an apple to Mary}
given
that both \emph{give} and
\emph{hand out} predicates bear semantic and argumental similarities.
The core idea of the model
is to generate data reflecting at the surface level\footnote{i.e.
at the sentence level, versus, encoded as a high-level feature}
structural information encoded in FrameNet through valence patterns. In theory,
such artificial data, highly consistent with the underlying logic of FrameNet
annotation, should not contribute significantly (if not at all) to improving
performances of (good) statistical models for frame semantic parsing.
Indeed, our model would only provide \emph{new} information in terms of
\emph{lexical configurations} of predicate-argument structures \textendash\
the association of a given predicate with a given set of arguments in a given
syntactic configuration, as previously exemplified in the sentence
\emph{John handed out an apple to Mary}.
It \emph{would not} introduce predicates, arguments or predicate-argument
configurations previously \emph{absent} from the data.
Therefore, if the performances of the aforementioned
statistical models were to improve
when trained on an artificially augmented dataset, it would only demonstrate
that such models fail to properly generalize predicate-argument structures
in FrameNet when exposed to a limited set of gold data. Our
paraphrastic data augmentation approach could offer an easy alternative
to compensate for the
shortcomings of such models, and a simple way to understand under which
conditions those models fail to generalize and what kind of
additional data, in terms of lexical predicate-argument configurations,
those models require to better \emph{learn}.

The structure of the memoir is as follows: in Chapter \ref{task} we introduce
in more details frame semantics theory, FrameNet and the frame semantic parsing task.
In Chapter \ref{relatedwork}, we go over the literature on frame semantic parsing and
critically analyze challenges posed by current approaches to frame semantic parsing
as well as the limitations of past error analysis. In Chapter \ref{modelxpsetup},
we introduce our paraphrastic data augmentation model and
the experimental setup used to generate artificial data.
In Chapter \ref{baseline}
we replicate the results of several past studies on frame semantic parsing
and detail the baseline used for quantifying the contribution
of our data augmentation model.
We report our results in Chapter \ref{results}, and analyze them
using different metrics in Chapter \ref{analysis}.
Finally, we discuss possible improvements in Chapter~\ref{discussion} and
conclude in Chapter \ref{conclusion}.

\chapter{Frame Semantic Parsing}
\label{task}
\emph{Frame semantics}
\citep{fillmore1982,backgroundfn2003,fillmore2009,fillmoreinterview2010}
is an approach to the study of lexical meaning where schematic representations
of events, relations or entities called \emph{frames} provide semantic
background for interpreting the meaning of words and illustrating their
syntactic valence.
\emph{Frame semantic parsing} is the task of automatically predicting predicate-argument
structures following the classification framework of frame semantics.
Pioneered by \cite{srlgildeajurafsky2002}, it was
more formally defined during the SemEval 2007 shared task 19 \citep{fnsemeval2007}
on \emph{Frame Semantic Structure Extraction}.

In this Chapter we present an introduction to the theory of frame semantics,
its implementation into the FrameNet lexicon, and the frame semantic parsing
task.
In Section~\ref{history} we introduce the theory of frame
semantics, its historical development, and its relation
to other linguistic theories such as \emph{case grammar}.
In Section~\ref{framenet} we introduce the FrameNet lexicon,
an implementation of frame semantics theory for English. We present its
core concepts, statistics
on annotated data, and several implementation specifications crucial to the
task of frame semantic parsing.
In Section~\ref{comparison}
we briefly compare FrameNet to other lexical resources for English
such as WordNet, PropBank or VerbNet.
Finally, in Section~\ref{taskparsing}, we
introduce the frame semantic parsing task as defined during the SemEval 2007
shared task, its datasets and evaluation protocols.

\section{A brief history of frame semantics theory}
\label{history}
\subsection{Historical motivations}
The motivations underpinning the frame semantics enterprise can be traced back
to Fillmore's early work on the classification of English verbs
\citep{fillmore1961,fillmore1963} and to the fundamental idea that discoveries
in the behavior of particular classes of words could lead to discoveries
in the structure of the grammar of English.
Fillmore's approach at the time was
resolutely transformationalist \textendash\ influenced by work such as
\citep{chomsky1957},
\citep{lees1960} or \citep{chomsky1965} \textendash\ in that it attempted to
classify
verbs according to \emph{surface-syntactic frames} and
\emph{grammatical behavior} defined
as sensitivity of certain classificatory structures to particular grammatical
transformations.

\subsection{The case for case}
\label{thecaseforcase}
Later on, in his seminal work on \emph{case grammar} \citep{case1968,fillmore1977a},
Fillmore proposed a more semantically-grounded theory where verbs
could be classified according to the semantic roles of
their associated arguments. Fillmore's work was then
largely influenced by \emph{dependency grammar} and \emph{valence theory}
\citep{tesniere1959}, notably in its treating all arguments of the
predicate equally, without special consideration for the subject, contrary
to previous transformationalist approaches.
Still motivated by considerations over the
universality of the deep syntactic structure of clauses, he introduced
a set of six (potentially)
universal semantic role categories called
\emph{deep cases}.\footnote{Agentive, instrumental, dative, factitive,
locative and objective}
Shared \emph{semantic valence} \textendash\ defined as a set of co-occurrence
restrictions
over the arguments of
predicates according to their semantic roles \textendash\ allowed for grouping
verbs into what Fillmore called \emph{case frames}, which provided
the necessary background for understanding the formation
of valid minimal clauses across languages. Abstracting over surface syntactic
configurations, case grammar made it possible to account for cases where
semantic roles occupy different syntactic positions, as in:\footnote{
 In all our examples containing semantic roles annotation, we follow
 FrameNet annotation style: we rely on a phrase structure grammar
 rather than a dependency grammar, i.e., we tag the whole constituents
rather than just the head words of the constituents}
\begin{examples}
 \item \label{exa} \feanno{John}{Agent} broke the window
 \item \label{exb} \feanno{A hammer}{Instrument} broke the window
 \item \label{exc} \feanno{John}{Agent} broke the window \feanno{with a hammer}{Instrument}
\end{examples}
In sentence \ref{exa} the item marked with the \fe{Agent} case is the
subject whereas in sentence \ref{exb}, the \fe{Instrument} is the subject.
Both \fe{Agent} and \fe{Instrument}  cases may appear in
the same sentence as in \ref{exc} but only the \fe{Agent} is subject.

The theory allowed for interesting predictions: the fact that the
subject in (\ref{exa}) and (\ref{exb})
have different deep cases explain why the combined
meaning of the two sentences could not be produced by co-joining the subjects,
as in:
\begin{example}
 *John and a hammer broke the window.
\end{example}
According to Fillmore, additional restrictions which may hold between cases
and lexical features (e.g.
between \fe{Agent} and animateness) could also explain why certain sentences are
unacceptable, such as:
\begin{example}
 *A hammer broke the glass with a chisel.
\end{example}

Both co-occurrence restrictions over deep cases and constraints between
cases and lexical features are fundamental properties of language that
every parser aiming at automatically predicting predicate-argument
structure must capture in order to properly identify and classify semantic roles.
As we will see in the following chapters, this poses major challenges to
the task of frame semantic parsing in general.

\subsection{From case grammar to frame semantics}
The theory of \emph{frame semantics} was formalized in 1982
to overcome the shortcomings of case grammar which, according to Fillmore,
failed to provide the details needed
for semantic description, especially for verbs in particular limited domains.
The concept of a (cognitive) semantic frame which would provide background for
semantic analysis was already salient in case grammar,
where Fillmore considered \emph{case frames} to characterize small abstract
\emph{scenes} or \emph{situations}, so that to understand the semantic structure
of a
verb it was necessary to understand the properties of such schematized scenes.

Some of the core concepts of frame semantics theory were introduced in
\citep{fillmore1977b} focusing on the characterization of the cognitive
\emph{scene} related to the notion of a \emph{commercial event}.
In his work, Fillmore argued that an important set of English verbs could
be seen as semantically related to each other given that they \emph{evoked}
the same general scene. The \emph{commercial event} schematic scene, or \emph{frame},
were to include concepts such as \fe{Buyer}, \fe{Seller}, \fe{Goods} and \fe{Money}.
Verbs such as \lexunit{buy}{v} would be said to focus on the actions of the
\fe{Buyer} with respect to the \fe{Goods}, backgrounding the
\fe{Seller} and the \fe{Money}, allowing sentences such as:
\begin{examples}
 \item \feanno{John}{Buyer} \target{bought} \feanno{a car}{Goods}
 \item \feanno{John}{Buyer} \target{bought} \feanno{a car}{Goods} \feanno{from Mary}{Seller}
 \item \feanno{John}{Buyer} \target{bought} \feanno{a car}{Goods} \feanno{from Mary}{Seller} \feanno{for \$5,000}{Money}
\end{examples}
while disallowing sentences such as:
\begin{examples}
 \item *\feanno{John}{Buyer} \target{bought} \feanno{from Mary}{Seller}
 \item *\feanno{John}{Buyer} \target{bought} \feanno{for \$5,000}{Money}
\end{examples}

Similarly, the verb \lexunit{sell}{v} would be said to focus on the actions
of the \fe{Seller} with respect to the \fe{Goods},
backgrounding the \fe{Buyer} and the \fe{Money}, while the verb \lexunit{pay}{v}
would be said to focus on the actions
of the \fe{Buyer} with respect to both the \fe{Money} and the \fe{Seller},
backgrounding the \fe{Goods}.
The point of the description was to argue that nobody could
be said to know the meanings of these verbs who did not know the
details of the kind of scenes that these words could represent.\\

In the following section we introduce the FrameNet implementation of
frame semantics theory for English.

\section{Introduction to FrameNet}
\label{framenet}

\subsection{Core concepts}

FrameNet \citep{baker1998berkeley,backgroundfn2003,fillmore2009,rupetal2016}
is a computational
lexicography project whose purpose is to provide reliable descriptions
of the syntactic and semantic combinatorial properties of each word in the
lexicon, and to assemble information about alternative ways of expressing
concepts in the same conceptual domain.
Its output takes the form of a database of corpus-extracted sentences annotated
in terms of frame semantics.
FrameNet is articulated around the following core concepts:
\begin{itemize}
 \item \textbf{frames}: schematic representations of events, relations or
       entities.
       They provide semantic background for understanding the meaning of words;
 \item \textbf{lexical units}: words paired with meaning. A lexical unit
       corresponds
       to a specific word sense and is formalized as a link between a \emph{lemma}
       and a \emph{frame}. The lexical unit is said to \emph{evoke} the frame it
       belongs to.
       A polysemous word will be formalized as a single lemma with references to
       multiple lexical units in distinct frames;
 \item \textbf{frame elements}: frame-specific semantic roles. Frame elements
       are said to be \emph{core} if they are obligatorily expressed,
       and \emph{non-core} if they are not (see examples~\ref{core}
       and~\ref{non-core} as well
       as Section~\ref{fecatego} for additional details on frame elements
       categorization
       in FrameNet). Frame element names generalize over frames, i.e. different
       frames with similar frame elements are considered to share semantic information.
       Frame elements can also be connected via frame
       element-to-frame element
       relations, to indicate semantic relations between them
       (see Section~\ref{fferelations});
 \item \textbf{valence units}: syntactic realizations of frame elements. A
       valence unit is represented as a triplet FE.PT.GF of frame
       element (FE), phrase type (PT) and grammatical function (GF);
 \item \textbf{valence patterns}: the range of combinatorial
       possibilities of valence units for each lexical unit. A single valence pattern
       is usually represented as a string of multiple valence units FE.PT.GF
       separated by whitespaces.
\end{itemize}
Let us now turn to a concrete example illustrating the aforementioned concepts.
In FrameNet, the \semframe{Commerce\_buy} frame is defined as follows:
\begin{quote}
 These are words describing a basic commercial transaction involving a Buyer
 and a Seller exchanging Money and Goods, taking the perspective of the Buyer.
 The words vary individually in the patterns of frame element realization they
 allow. For example, the typical pattern for the verb BUY: Buyer buys Goods
 from Seller for Money.
\end{quote}
It contains six lexical units: \lexunit{buy}{v}, \lexunit{buyer}{n},
\lexunit{client}{n}, \lexunit{purchase [act]}{n}, \lexunit{purchase}{v} and
\lexunit{purchaser}{n}.
It also contains two core frame elements, \fe{Buyer} and \fe{Goods}, which are
necessary to form grammatical clauses with lexical units belonging to the
frame, as in:
\begin{example}
 \label{core}
 \feanno{John}{Buyer} \target{bought} \feanno{a car}{Goods}
\end{example}
It also contains thirteen non-core frame elements, such as, e.g., \fe{Money},
\fe{Place},
\fe{Seller}, \fe{Recipient} or \fe{Time} which may or may not be expressed
in a given clause without altering its grammaticality, as in:
\begin{example}
 \label{non-core}
 \feanno{John}{Buyer} \target{bought} \feanno{Susan}{Recipient}
 \feanno{a car}{Goods}
 \feanno{yesterday}{Time} \feanno{for \$5,000}{Money}
\end{example}
The two core frame elements \fe{Buyer} and \fe{Goods}, occur in valence patterns
such as:\footnote{In the following examples, `NP' is a noun phrase, `Ext'
 an external argument (the subject), `Obj' an object, `PP'
 a prepositional phrase,
`AJP' and adjective phrase and `Dep' a dependent}
\begin{itemize}
 \item \vpattern{Buyer.NP.Ext Goods.NP.Obj}, as in the simple direct object
       construction:
       \begin{example}
        \vuanno{John}{Buyer}{NP}{Ext} \target{bought}
        \vuanno{a car}{Goods}{NP}{Obj}
       \end{example}
 \item \vpattern{Buyer.PP.Dep Goods.NP.Ext}, for the passive construction:
       \begin{example}
        \vuanno{A car}{Goods}{NP}{Ext} was \target{bought}
        \vuanno{by John}{Buyer}{PP}{Dep}
       \end{example}
 \item \vpattern{Buyer.NP.Ext Goods.AJP.Dep} for the adjectival construction:
       \begin{example}
        \vuanno{John}{Buyer}{NP}{Ext} \target{bought}
        \vuanno{American}{Goods}{AJP}{Dep}
       \end{example}
\end{itemize}

\subsection{Data and statistics}
FrameNet mostly annotates the British National Corpus (BNC) and
the American National Corpus (ANC). FrameNet annotation is divided into two sets:
\begin{enumerate} \item \textbf{exemplar}: also referred to as
 \emph{lexicographic annotation}. Exemplars are the core of the FrameNet
 lexicographic project. Each exemplar sentence exemplifies the use of a valence
 pattern for a single lexical unit; \item \textbf{fulltext}: developed with NLP
 applications in mind, fulltext annotations cover FrameNet annotation for entire
 corpus-extracted texts where all possible lexical units and subsequent valence
 patterns of each sentence are annotated. \end{enumerate}
 Table~\ref{rawfndatastats} and Table~\ref{compfndatastats} below provide
 interesting statistics regarding three major releases of the English FrameNet
 data: 1.3, 1.5 and 1.7.\footnote{See the release notes on \url{https://framenet.icsi.berkeley.edu}}
 Release 1.3 formed
 the basis of the SemEval 2007 shared task (see Section~\ref{taskparsing}),
 release 1.5 was used in most recent state-of-the-art work on frame semantic
 parsing (see Section~\ref{fsparsinghistory}), while release 1.7, the latest, is
 used throughout our work (see Chapter~\ref{baseline}).

 \begin{table}[h!tpb]
  \centering
  \begin{tabular}{lcrrr}
   \toprule
                              &            & R1.3              & R1.5              & R1.7              \\
   \midrule
   \multicolumn{2}{l}{Frames} & \numprint{795}  & \numprint{1019} & \numprint{1221}  \\
   Frame Elements             & (FEs)      & \numprint{7124}   & \numprint{8884}   & \numprint{10503}  \\
   Lexical Units              & (LUs)      & \numprint{10195}  & \numprint{11829}  & \numprint{13572}  \\
   Valence Units              & (VUs)      & \textendash       & \numprint{22734}  & \numprint{26563}  \\
   Valence Patterns           & (VPs)      & \textendash       & \numprint{44499}  & \numprint{51470}  \\
   \midrule
   Annotation Sets            & (AnnoSets) & \numprint{151110} & \numprint{173018} & \numprint{202225} \\
   \hspace{20px} in exemplars &            & \numprint{139439} & \numprint{149931} & \numprint{174017} \\
   \hspace{20px} in fulltext  &            & \numprint{11671}  & \numprint{23087}  & \numprint{28208}  \\
   \bottomrule
  \end{tabular}
  \caption{Raw FrameNet data statistics for releases 1.3, 1.5 and 1.7}
  \label{rawfndatastats}
 \end{table}

 \begin{table}[h!tpb]
  \centering
  \begin{tabular}{llrrr}
   \toprule
     &   & R1.3 & R1.5 & R1.7 \\
   \midrule
   \multicolumn{2}{l}{FEs per Frame} & \numprint{9.88}  & \numprint{9.78} & \numprint{9.7}  \\
   \multicolumn{2}{l}{LUs per Frame} & \numprint{14.14} & \numprint{13.03} & \numprint{12.5} \\
   \midrule
   \multicolumn{2}{l}{LUs with AnnoSets} & \numprint{66.8}\% & \numprint{65.2}\% & \numprint{62}\%      \\
   \midrule
   \multicolumn{2}{l}{AnnoSets per LU}  & \numprint{20.5} & \numprint{19.5} & \numprint{20.73}      \\
   \bottomrule
  \end{tabular}
  \caption{Computed FrameNet data statistics for releases 1.3, 1.5 and 1.7}
  \label{compfndatastats}
 \end{table}

 FrameNet is also being developped in other languages such as Spanish \citep{subirats2003},
 Swedish \citep{borin2010} or Japanese \citep{ohara2004}, for which
 we provide comparative annotation
 statistics in Table~\ref{multifndatastats}.

 \begin{table}[h!tpb]
  \centering
  \begin{tabular}{lrrrr}
   \toprule
                         & Spanish         & Swedish          & Japanese        & English          \\
   \midrule
   Nouns                 & \numprint{271}  & \numprint{28891} & \numprint{2043} & \numprint{5299}  \\
   Verbs                 & \numprint{856}  & \numprint{5398}  & \numprint{908}  & \numprint{5141}  \\
   Adjectives            & \numprint{99}   & \numprint{3293}  & \numprint{134}  & \numprint{2347}  \\
   Adverbs               & \numprint{16}   & \numprint{322}   & \numprint{89}   & \numprint{214}   \\
   Other                 & \numprint{26}   & \numprint{124}   & \numprint{231}  & \numprint{420}   \\
   \midrule
   Total Lexical Units   & \numprint{1268} & \numprint{38028} & \numprint{3405} & \numprint{13421} \\
   Total Annotation Sets & \numprint{11}k  & \numprint{9}k    & \numprint{73}k  & \numprint{200}k  \\
   \bottomrule
  \end{tabular}
  \caption{Multilingual FrameNet data statistics according to~\cite{bakeretal2015}}
  \label{multifndatastats}
 \end{table}

 \subsection{Frame and frame element relations}
 \label{fferelations}
 Frames in FrameNet form a network of interconnected entities,
 connected to each other via eight types of
 \emph{frame-to-frame relations}. Frame elements are similarly connected to
 each other via \emph{frame element-to-frame element relations}.
 Frame relations include:
 \begin{enumerate}
  \item \textbf{Inheritance}: a \emph{child} frame inherits from a
        \emph{parent} frame if its equally of more specific than its parent.
        For example, the \semframe{Commerce\_buy} frame inherits from
        the \semframe{Getting} frame,
        which can be understood intuitively by the fact that \emph{buying} something
        is \emph{getting} it, with the additional assumption that the thing
        received was exchanged
        for money. Hence, the \fe{Buyer} frame element of the \semframe{Commerce\_buy}
        frame is related to the \fe{Recipient} frame element of the \semframe{Getting}
        frame but is semantically more specific.
        \begin{examples}
         \item \feanno{John}{Recipient} \target{acquired} \feanno{a house}{Theme}
         \item \feanno{John}{Buyer} \target{bought} \feanno{a house}{Goods}
        \end{examples}

  \item \textbf{Perspective\_on}: a given frame is in a Perspective\_on relationship
        to a \emph{neutral} frame if it provides a specific perspective on the neutral frame.
        For example, both the \semframe{Commerce\_buy} frame and
        the \semframe{Commerce\_sell} frame are in a Perspective\_on
        relationship to the \semframe{Commerce\_goods-transfer} frame: the
        \semframe{Commerce\_buy}
        frame provides the point-of-view of the buyer while the
        \semframe{Commerce\_sell} frame
        provides the point-of-view of the seller, although both frames relate
        to the same \emph{scene}, as shown by the following sentences with
        the lexical units \semlexunit{Commerce\_buy}{buy}{v} and
        \semlexunit{Commerce\_sell}{sell}{v}:
        \begin{examples}
         \item \feanno{John}{Buyer} \target{bought} \feanno{a car}{Goods}
         \feanno{from Mary}{Seller} \feanno{for \$5,000}{Money}
         \item \feanno{John}{Seller} \target{sold} \feanno{a car}{Goods}
         \feanno{to Mary}{Buyer} \feanno{for \$5,000}{Money}
        \end{examples}

  \item \textbf{Using}: a \emph{target} frame is in a Using relationship to a
        \emph{source}
        frame if a part of the scene evoked by the target frame refers to the source frame.
        Note that a frame can be in a Using relationship with multiple other frames.
        For example, the \semframe{Judgment\_communication} frame is in a Using
        relationship to both
        the \semframe{Judgment} frame and
        the \semframe{Statement} frame. It does not Inherit from the
        \semframe{Judgment} frame
        as it is not a simple sub-type of a purely cognitive state. Similarly,
        it does not Inherit from the \semframe{Statement} frame as its
        frame element corresponding to the \semframe{Statement} frame's \fe{Message}
        frame element
        is actually split in two separate frame elements: \fe{Evaluee} and \fe{Reason};

  \item \textbf{SubFrame}: a \emph{target} frame is in a SubFrame relationship to
        a \emph{source} frame if the source frame refers to a sequence of \emph{states}
        and \emph{transitions}, of which the target frame is a particular instance.
        For example, the \semframe{Arrest}, \semframe{Arraignment}, \semframe{Trial},
        \semframe{Sentencing}
        and \semframe{Appeal} frames are all in a SubFrame relationship with
        the \semframe{Criminal\_process} frame;

  \item \textbf{Precedes}: a \emph{target} frame is in a Precedes relationship to
        a \emph{source} frame if both target and source frames refer to specific states
        in a sequence of states and if the target frame has temporal or ordinal
        precedence to the source frame.
        For example, the aforementioned \semframe{Arrest} frame is in a Precedes
        relationship to the \semframe{Arraignment} frame;

  \item \textbf{Causative\_of and Inchoative\_of}: a \emph{target} frame is in a
        Causative\_of, Inchoative\_of relationship to a stative \emph{source} frame if it
        describes the causative / inchoative view on a given stative scene.
        Causative frames will typically also inherit from the
        \semframe{Transitive\_action}
        frame and contain an \fe{Agent} frame element
        as in ``John broke the window'', and inchoative frames will typically
        inherit from the \semframe{Event}, \semframe{State} or
        \semframe{Gradable\_attribute} frame;

  \item \textbf{Metaphor}: a \emph{target} frame is in a Metaphor relationship
        to a \emph{source} frame if many of the lexical units in the target frame
        are to be understood at least partially in terms of the source frame.
        Consider the Metaphor relation between the \semframe{Cause\_motion} frame and
        the \semframe{Suasion} frame, as in:
        \begin{example}
         \label{movedmetaphor}
         The judge was not \target{moved} by the lawyer's argument
        \end{example}
        In \ref{movedmetaphor}, ``moved'' would be annotated in the
        \semframe{Cause\_motion} frame,
        with the annotation set marked with the Metaphor label;

  \item \textbf{See\_also}: a \emph{target} frame is in a See\_also relationship to
        a \emph{source} frame if it is similar to the source frame but should be
        carefully differentiated,
        compared and contrasted. This is intended to help human users better make
        sense of the FrameNet frame differentiation.
 \end{enumerate}
 We will detail in Section~\ref{fsparsinghistory} and
 Section~\ref{challengesfsparsing} how frame and frame element relations have
 been successfully used to improve automatic predictions of frame element labels,
 in work such as \citep{kshirsagaretal2015} or \citep{matsubayashi2014}.

 \subsection{Frame element attributes}

 \subsubsection{Core types}
 \label{fecatego}
 FrameNet distinguishes between core and non-core frame elements,
 following the traditional distinction between core arguments and peripheral
 adjuncts (such as \emph{time}, \emph{place} and \emph{manner}).
 Those frame element categories are referred to in FrameNet as \textbf{core} and
 \textbf{peripheral}.
 FrameNet also contains two additional frame element
 categories:
 \begin{enumerate}
  \item \textbf{Extra-thematic} frame elements, which are frame elements
        that introduce
        information which is not a necessary part of the description of
        the central frame
        they belong to. For example, the \fe{Recipient} frame element in
        the \semframe{Commerce\_buy} frame is extra-thematic: it introduces a
        concept
        incorporated from the \semframe{Giving} (and subsequent) frame
        which is not core in the \semframe{Commerce\_buy}
        frame, as shown in the following sentences:
        \begin{examples}
         \item \feanno{Mary}{Buyer} \target{bought} \feanno{a car}{Goods}
         \item \feanno{Mary}{Buyer} \target{bought} \feanno{a car}{Goods} \feanno{for John}{Recipient}
        \end{examples}
        The difference between extra-thematic and peripheral frame elements is that
        peripheral frame elements do not uniquely characterize a frame, and can be
        instantiated in any semantically appropriate frame;

  \item \textbf{Core-unexpressed} frame elements, which are frame elements that
        behave like core frame elements in the frame they belong to but are not
        necessarily listed among the frame elements in the descendant frames.
        This can be explained by cases where information bear by the frame
        element is \emph{included} in
        the lexical units in the descendant frame, and cannot be
        separately expressed.
        This is exemplified in the \semframe{Choosing} frame which inherits from
        the \semframe{Intentionally\_act} frame containing a core-unexpressed
        \fe{Act} frame element, which cannot be passed on to the \semframe{Choosing}
        frame
        without forming ungrammatical sentences as in~\ref{actungramm}:
        \begin{examples}
         \item \feanno{I}{Agent} will \target{do} \feanno{the cooking}{Act}
         \item \label{actungramm} *\feanno{I}{Cognizer} will \target{choose} \feanno{decision}{Act} \feanno{to cook}{Chosen}
        \end{examples}
 \end{enumerate}

 \subsubsection{Null instantiations}
 There are many occurrences in corpora where core frame elements do not appear
 in given sentences. Those cases are referred to in FrameNet as
 \emph{null instantiations}.
 FrameNet categorizes three cases of null instantiation:
 \begin{enumerate}
  \item \textbf{Indefinite Null Instantiation} (INI): refers to cases where
        an argument is omitted for intransitive lexical units. FrameNet marks transitive
        and intransitive lexical units as both transitive and, in intransitive cases,
        marks the core argument as INI, as in:
        \begin{examples}
         \item \feanno{I}{Ingestor} ate \feanno{a cake}{Ingestibles}
         \item \feanno{I}{Ingestor} ate \hspace{114px} \fe{Ingestibles}.INI
        \end{examples}
  \item \textbf{Definite Null Instantiation} (DNI): refers to cases where an
        argument is omitted under lexically licensed zero anaphora, where all parties
        in a conversation are assumed to know what the argument is, as in:
        \begin{examples}
         \item \feanno{We}{Competitor} won! \hspace{80px} \fe{Competition}.DNI
         \item When did \feanno{they}{Theme} arrive?  \hspace{31px} \fe{Goal}.DNI
        \end{examples}
  \item \textbf{Construction Null Instantiation} (CNI): refers to cases where
        the grammar considerations require or permit the omission of some arguments
        in particular structures. This is typically the case for imperative sentences
        which omit the subject, and passive sentences which omit the agent, as in:
        \begin{examples}
         \item Get out! \hspace{130px} \fe{Theme}.CNI
         \item We have been robbed \hspace{71px} \fe{Perpetrator}.CNI
        \end{examples}
 \end{enumerate}
 Null instantiations exhibit different linguistic phenomena which have been shown
 to be particularly hard to learn by machine learning algorithms used for
 frame semantic parsing. Indeed, frame semantic parsers have a tendency to
 wrongly predict core frame elements when confronted to null instantiations,
 negatively affecting recall metrics (see \citep{chen2010} detailed in
 Section~\ref{challengesfsparsing}).

 \subsubsection{Semantic types}
 Frame element semantic types in FrameNet are designed primarily
 to aid frame parsing and automatic frame element recognition.
 They provide a general and limited set of categories which characterize
 the basic typing of fillers of frame elements. For example, the \fe{Cognizer}
 frame element is marked as `Sentient', the \fe{Place} frame element
 as `Locative\_relation', and the
 \fe{Purpose} or \fe{Means} frame elements as `State\_of\_affairs'.
 We will detail in Section~\ref{challengesfsparsing} how
 \cite{matsubayashi2014} successfully used
 frame elements semantic types to improve automatic predictions of
 frame element labels.

\section{FrameNet and alternative lexical resources}
\label{comparison}

\subsection{WordNet}
WordNet \citep{miller1995wordnet} is a large electronic lexical database of
English, which was inspired by pycholinguistics research investigating how and
what type of information is stored in the human mental lexicon.
WordNet groups words into sets of synonyms called \emph{synsets} and records
a number
of relations among lexical entries, such as synonymy/antonymy,
 hyponymy/hypernymy, meronymy/holonymy, troponymy or entailment.
 It has a large coverage, with more than 200,000 word-sense pairs documented in
  four morphosyntactic categories: nouns, verbs, adjectives and adverbs.
Contrary to FrameNet, WordNet contains little syntagmatic information. Its purpose
is not to document predicate-argument structures in general. It does, however, offer more
fine-grained distinctions across lexical entries, especially for nouns, and is
therefore complementary to FrameNet.

\subsection{PropBank}
PropBank \citep{propbank2005} is a resource which annotates predicate-argument
structures on top of the phrase structure annotation of the
Penn TreeBank \citep{marcus1993}. Predicates senses are classified according to
frames, as in FrameNet, but, unlike FrameNet, PropBank postulates
predicate-specific roles
which are then generalized to a set of 5 core roles classes
(ARG0 to ARG4)\footnote{ARG0: agent, ARG1: patient, ARG2: instrument,
benefactive, attribute,
ARG3: starting point, benefactive, attribute, ARG4: ending point} and
18 adjunct/modifier classes (ARGM-*)\footnote{Such as ARGM-LOC for
location and ARGM-TMP for time}.
PropBank annotates verbs exclusively and does not offer the same granularity of
semantic roles and frames as in
FrameNet. It also does not provide semantic relations across frames and roles.
Its generalized roles set does however make the task of automatic semantic role
labeling simpler and it has been made it a popular resource in parsing and
computational linguistics consequently.

\subsection{VerbNet}
VerbNet \citep{verbnet2005} is a digital database that provides predicate-argument
structures documentation for English verbs, relying on a hierarchical
classification of verbs inspired by Levin's verb classes \citep{levin1993}.
Arguments in VerbNet consist of a list of 23 generic thematic roles. Each (syntactic)
frame groups together verbs that have similar semantic and syntactic
characterizations, and VerbNet provides documentation for frame-specific
thematic role syntactic realizations and co-occurence constraints.
VerbNet has a much lower coverage than FrameNet, with only 4526 senses
of 3769 verbs annotated, and, unlike FrameNet, does not encompass (semantic)
relations across
its entities.

\subsection{Others}
Several other resources also aim at documenting predicate-argument structures
 for English: the Valency Dictionary of English \citep{herbst2004valency},
 which, unlike previously mentioned resources, is not free and not entirely
 machine-readable;
 VALEX \citep{korhonen2006large}, a large valency subcategorization lexicon for
 English verbs, which is, unlike previously mentioned resources,
 partially automatically annotated and therefore somehow noisy; and
 Abstract Meaning Representation (AMR) \citep{amr2013}, the newest resources
 of all. AMR offers a graph-based representation of lexical concepts and
 typed relations between
 those concepts, denoted by an English sentence. It covers PropBank
 predicate-argument semantics, entity linking, coreference, modality, negation,
 questions, relations between nominals and canonicalization of content words.
 Its graph-based representation allows for abstracting away grammatical
 specificities, and its single-structure representation is designed to support
 rapid
 corpus annotation and data-driven natural language processing.

\section{The frame semantic parsing task}
\label{taskparsing}

\subsection{Definition}
\label{taskdefinition}
Frame semantic parsing is the task of automatically predicting frame-evoking
words, frames, and corresponding frame element spans and labels
in a given sentence.
The task, formally defined in the SemEval 2007 shared task 19 on
\emph{Frame Semantic Structure Extraction} \citep{fnsemeval2007}, was divided
into three subtasks:
\begin{enumerate}
 \item \textbf{target identification}: identify all frame evoking words
       in a given sentence;
 \item \textbf{frame identification}: label all frames of pre-identified
       targets in a given sentence;
 \item \textbf{argument identification}: identify all frame-specific frame
       element spans and labels for pre-identified lexical units in a given sentence.
\end{enumerate}

\subsection{Datasets}
Data were based on an extended version of the FrameNet 1.3 data release.
Training data comprised about 2,000 annotated sentences with about 11,000
annotation sets in fulltext documents.
Testing data comprised about 3 annotated fulltexts composed of 120 sentences
and about 1,000 annotation sets (See details in \citep{dasetal2014}).
Texts were extracted from the American National Corpus (ANC) and the
NTI website.\footnote{\url{http://www.nti.org/}}

In this work, we first attempt to
replicate past studies and therefore rely on the original FrameNet 1.5-based
training and testing sets of \cite{dasetal2014} (see Chapter~\ref{baseline} and
Section~\ref{xpsetupdatasets}).
However, we also report results on
FrameNet 1.7-based datasets, where the testing set contains the same list of
fulltext documents as the original list of \cite{dasetal2014},
but contains more annotated data as those documents were further annotated
in the 1.7 release (see Table~\ref{compfndatastats}) and therefore produce
a more robust testing set.

\subsection{Evaluation}
\label{taskevaluation}

\subsubsection{Scoring}
The original task was evaluated globally, each subtask taking as input the predicted
output of the previous subtask and global precision recall and $F_1$ scores
were computed for frames and arguments microaveraged across a concatenation of
the test set sentences.

\begin{figure}
 \includegraphics[scale=0.55]{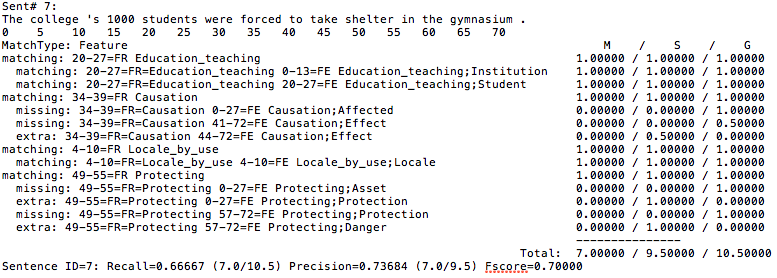}
 \caption{SemEval evaluation script output with frames scores}
 \label{semevalevalsnippet}
\end{figure}

\begin{figure}
 \includegraphics[scale=0.55]{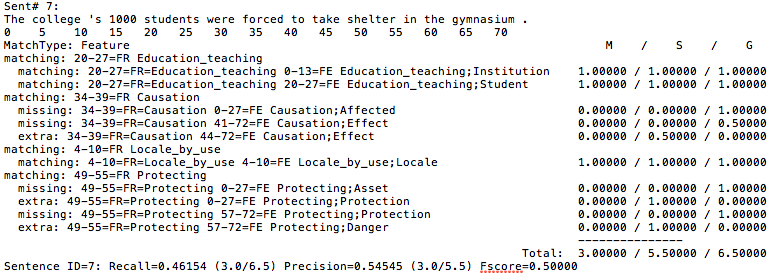}
 \caption{Modified SemEval evaluation script output without frames scores}
 \label{aclevalsnippet}
\end{figure}

In this work, we focus mostly on the subtask of argument identification and
therefore evaluate predictions of arguments spans and labels
given gold targets and gold frames.
We use the modified evaluation script of the SemEval shared task introduced
by \cite{kshirsagaretal2015}. The modified evaluation script does not give
extra credit for gold frames, unlike the standard SemEval script. This allows for
better evaluation of parsers' improvements in argument identification, as it does
not dilute argument identification scores by blending them with (gold) frame
scores, necessarily 100\% accurate. For example, Figure~\ref{semevalevalsnippet}
which shows the original SemEval evaluation script
output for the sentence \emph{The college's students were forced to take
shelter in the gymnasium}, contains full M/S/G scores on the fourth line of
the output script, corresponding to the frame (FR)
\semframe{Education\_teaching} evoked by the target ranging from character
20 to character 27. A contrario, Figure~\ref{aclevalsnippet},
which shows the modified
evaluation script output for the same sentence, does not contain any M/S/G scores
for the same line.

The evaluation script scores each frame and argument independently. In our
configuration, argument spans and labels are scored jointly.
Core frame elements
are given one point and non-core frame elements are given half a point.
The `Match' column M is assigned a value each time a correct argument is
identified (true positive); the `Score' column `S' is assigned a value
each time
an argument is wrongly predicted (false positive); and the `Gold' column G is
assigned a value each time the corresponding span and label is gold annotation.

Precision is then computed by summing over all true positives and false positives
for all predicted labels in a given sentence:
\begin{equation}
 P = \frac{\sum{t_p}}{\sum{t_p} + \sum{f_p}} = \frac{\sum{M}}{\sum{S}}
\end{equation}

Recall is computed in a similar fashion with true positives and false negatives:
\begin{equation}
 R = \frac{\sum{t_p}}{\sum{t_p} + \sum{f_n}} = \frac{\sum{M}}{\sum{G}}
\end{equation}

$F_1$ score is then given by:
\begin{equation}
 F_1 = 2 \cdot \frac{P \cdot R}{P + R}
\end{equation}

\subsubsection{Statistical significance}
Finally, we measure the statistical significance of all our results relying on the
bootstrapping approach of \cite{bergkirkpatrick2012}: let us consider
a system $A$ which we compare to a baseline system $B$. Let us
consider additionally a test set
$x = x_1,\ldots,x_n$ on which $A$ beats $B$ by $\delta(x)$.
$b$ bootstrap samples $x^{(i)}$ of size $n$ are drawn by sampling with
replacement from $x$. For each $x^{(i)}$, a variable $s$, initialized at 0,
is incremented if and only if $\delta(x^{(i)}) > 2\delta(x)$. The $p$
value is then given by:
\begin{equation}
 p(x) = \frac{s}{b}
\end{equation}

\chapter{Related Work}
\label{relatedwork}

\section{A historical overview of frame semantic parsing}
\label{fsparsinghistory}

Early work on frame semantic parsing has relied almost exclusively
on statistical classifiers based on an exhaustive list of carefully
handcrafted features
\citep{srlgildeajurafsky2002,johansson2007,dasetal2014,hermann2014,kshirsagaretal2015,rothlapata2015,tack2015}.
Such approaches usually require extensive feature engineering which can prove particularly
time-consuming, in addition to risking biasing systems for the task at hand \citep{rothlapata2016}.
Moreover, those approaches have typically made use of heavy preprocessing,
be it through part-of-speech tagging, dependency-parsing, or coreference
resolution \citep[e.g.][]{dasetal2014,rothlapata2015}. However, preprocessing
has been shown to significantly impact the overall performances of statistical
models for frame semantic parsing
(see Chapter~\ref{baseline} for details), and even to strictly bound the
performances of those models below an unsatisfactory threshold \citep{dasetal2014,tack2015}.
On top of this, previous research have relied on different preprocessing toolkits, and
those inconsistent experimental setups have prevented fair comparison between models,
as they do not allow discounting the contribution of the preprocessing toolkit
from the contribution of the statistical model itself.

Given those considerations, recent neural-network-based approaches to frame
semantic parsing appear very promising \citep{fitzgerald2015,roth2016,swayamdipta2017,yangmitchell2017},
given the limited number of core features they require as input. In this regard,
the most exciting approach is probably that of \cite{swayamdipta2017}, which
proposed two very interesting models: one requiring no syntactic preprocessing
and one acquiring syntactic representations alongside other distributional
representations in a multi-task setting. Both
models yielded competitive if not above state-of-the-art results.

Recent approaches have also called into question how the frame semantic parsing
task itself has been formalized, confirming what intuition long suggested:
frame and argument identification performed jointly yield better results
than performed sequentially~\citep{yangmitchell2017}.

State-of-the-art results on frame identification are reported in
\citep{hermann2014} at 88.41\% $F_1$ score; on argument identification with gold frames,
in \citep{swayamdipta2017} at 68.9\% $F_1$ and on argument identification
with predicted frames in \citep{yangmitchell2017} at 76.6\% $F_1$.
Readers can refer to Appendix~\ref{detailedoverviewfsparsing}
for a more exhaustive historical overview of
supervised and semi-supervised methods for frame semantic parsing.

\section{Challenges of frame semantic parsing}
\label{challengesfsparsing}
Some important challenges posed by frame semantic parsing are deeply rooted
in its underlying theoretical formalization. Indeed, frame semantics theory
operates many fine-grained distinctions within predicate-argument
structures classification, many of which are semantically motivated
and do not translate into linguistic phenomena that can easily be captured
at the surface syntactic level \citep{croceetal2010}.
Concretely, those considerations have turned FrameNet into an exhaustive
taxonomy
which has rendered the task of frame semantic structure prediction
very complex (see Section~\ref{richannotation}).

The richness of FrameNet's classification, which is also its main asset compared
to other lexical resources, has lead researchers to operate multiple approximations
in order to be able to solve the task at hand. Those approximations dealt with
the definition of the frame semantic parsing task itself, and with its
resolution, both in terms of modeling and
optimization (see Section~\ref{necessaryapproximations}).

In addition, the fined-grained nature of FrameNet annotation
has rendered its production extremely costly, resulting
in limited annotated data compared to what is usually required
for machine learning algorithms to learn the kind of generalizations necessary
to properly predict frame semantic structures, within and across domains
(see Section~\ref{scarcedomainspec}).

In this section we skim through the existing literature on each of those challenges
and detail some of the solutions that have been proposed to tackle them.

\subsection{A rich underlying taxonomy}
\label{richannotation}

The FrameNet taxonomy, albeit rich and complex,
is also well structured. From this observation, \cite{matsubayashi2014}
proposed to exploit the structural information available in FrameNet \textendash\
such as frame and frame element relations and semantic types \textendash\
combined with other lexical resources such as PropBank and VerbNet, to infer
generalizations on frame elements which would reduce the number of classes for
 the argument identification task, and thereby its complexity.
 Although they did show
significant improvements, with a 19.16\% error reduction in terms of total
accuracy, their approach can be viewed as a simplification of the task of frame
semantic parsing in general, a simplification which discards many
interesting distinctions made
by the FrameNet framework. It does however support in an indirect fashion
the underlying annotation of FrameNet, as it shows that generalizations motivated
by theoretical considerations from frame semantics theory generate systematic
linguistic phenomena in as they can be captured by a machine learning system.

Other forms of linguistic knowledge annotated in FrameNet have proven harder to
capture through automatic means, whether directly or indirectly.
\cite{dasetal2014} for instance, showed that a traditional log-linear model relying
on a latent variable over frame and argument representations failed to capture
the kind of structural constraints on frame elements which are hard-coded in FrameNet.
Such constraints included a vast majority of non-overlapping frame element labels,
\textendash\ which the system violated 441 times\footnote{About 10\% of the time, given the size
of the test set} \textendash\ and other forms of constraints, such as argument
pairwise requirement/exclusion, a constraint exemplified in the following sentences:
\begin{examples}
  \item \feanno{A blackberry}{Entity\_1} resembles \feanno{a loganberry}{Entity\_2}
  \item \feanno{Most berries}{Entities} resemble each other
\end{examples}
The frame elements \fe{Entity\_1} and \fe{Entity\_2} \emph{require} one another
and \emph{exclude} the \fe{Entities} frame element and vice versa.

In a similar vein \cite{chen2010} showed that the frame semantic parser of
\cite{dasetal2014} often failed to correctly identify null instantiations of
core frame elements. As for \cite{kshirsagaretal2015}, they showed that
the same parser also failed to capture linguistic knowledge encoded in
frame relations, and that such information could only be adequately captured
through active supervision.

\subsection{Necessary approximations}
\label{necessaryapproximations}
In order to simplify automatic frame semantic structure extraction, past research
have proceeded to multiple approximations.

The first approximation
deals with the definition of the task itself,
which has been artificially split into the three subtasks of
\emph{target identification},
\emph{frame identification} and \emph{argument identification}.
Nothing in the theory of frame
semantics requires this subdivision of labor, which is more commanded by
requirements in terms of evaluation for natural language processing.
On the contrary, it is often useful in order to perform frame identification for
a given predicate, to process the arguments of the predicates \emph{in context}.
The recent work of \cite{yangmitchell2017} has confirmed this intuition,
achieving a significant improvement in both frame and argument identification
tasks by performing those two tasks jointly rather than sequentially.

The second approximation deals with input processing, which has systematically
been performed at the sentence level, despite theoretical considerations in
frame semantics stressing the importance of processing language at a higher
level than the mere sentence \citep{fillmore1982}. Again, this
 constraint is due to computational
considerations, and the recent work of \cite{rothlapata2015} who
improved argument identification performance by incorporating
discourse features through, e.g., coreference resolution, confirms the
limitations
posed by this approximation.

Other approximations have been performed to ease the modeling and better fit
it into a machine learning pipeline. Such approximations include that of
\cite{dasetal2014} on the nature and the systematicity of structural constraints
which operate on frame elements. Those approximations, which often have no
theoretical grounds, allow formulating the problem of argument identification
as a constrained optimization problem, turning a complex structured prediction
problem into a more standard optimization problem, easier to process with, e.g.,
linear programming solvers.

Finally, work such as \citep{dasetal2014,tack2015,fitzgerald2015} have demonstrated
how the performances of traditional approaches to frame semantic parsing, relying on heavy
syntactic preprocessing, could be bounded by the performances of syntactic parsers.
This is particularily striking in the case of argument span identification methods,
which, in the aforementioned work, have relied on heuristics-based algorithms
exploiting syntactic features derived from syntactic parsers, and which have
bounded arguments span identification recall to a low 80\%.

The impact of preprocessing on frame semantic parsing will be further detailed
in Chapter \ref{baseline}, where we will discuss how inconsistent preprocessing
pipelines across research and experimental setups may pose problems for
replicating past results and quantifying systems contributions.

\subsection{Annotation data scarcity and domain specificity}
\label{scarcedomainspec}

It has frequently been argued that FrameNet lacked the necessary amount
of annotated data for machine learning systems to properly predict
frame semantic structure \citep{palmercoverage2010,dasetal2014,das2014,fitzgerald2015},
or generalize across domains \citep{croceetal2010,dasetal2014,hartmannetal2017}.

The negative impact of coverage gaps has been extensively documented
by \cite{palmercoverage2010}, albeit limited to the task of frame identification,
which had a knock-on effect on argument identification for frame semantic parsing
pipelines processing tasks sequentially.

While \cite{johansson2007} noted a 20 $F_1$ loss on argument identification when
tested on a different domain than that of the training set, \cite{dasetal2014} argued
further that
exemplar sentences could not successfully be used for training frame semantic parsers
on the SemEval data. According to them, possible explanations included
the fact that exemplars were not representative as a sample, did not have complete
annotations, and were not from a domain similar to the fulltext test data.

Yet, recent work of \cite{kshirsagaretal2015} and \cite{yangmitchell2017} have
shown a few $F_1$ points gain when training on both fulltexts and exemplars
of the FrameNet 1.5 dataset, with and without domain adaptation techniques.
Those improvements, however, remain limited considering that the exemplar sentences
represent more than three times the amount of annotated data available in fulltext data.

Many approaches have been tried to overcome the problems of data sparsity,
coverage gaps and domain specificities.
\cite{croceetal2010} and \cite{hartmannetal2017} proposed to rely on distributional
representations of lexical features to better generalize in out-of-domain
settings. \cite{furstenau2012} augmented the FrameNet
annotation set by projecting FrameNet role annotation to syntactically
similar sentences, while \cite{hartmann2016} augmented the set of annotation for both
FrameNet senses and roles using distant supervision to transfer annotation for
linked lexical resources.
\cite{rastogiandvandurme2014} and \cite{pavlicketal2015} proposed to augment
FrameNet annotation automatically via paraphrasing, while \cite{hongbaker2011},
\cite{fossatietal2013}, \cite{chang2015scaling} and \cite{chang2016linguistic}
proposed to do so via crowdsourcing.

However, the aforementioned methods have either yielded limited improvements,
either not been tested in traditional frame semantic parsing setups
for proper comparison, either not been tested at all. It is therefore difficult
to measure the true impact of FrameNet data augmentation on frame semantic parsing.
Even if such improvements were to be attested, its impact would remain limited
without a proper error analysis which would help us understand exactly \emph{what
kind of linguistic knowledge} additional data helped capture.

\section{Shortcomings of past error analysis}
Although much work has been done over the past ten years in the field of
frame semantic parsing, very little has been done in terms of error analysis
to understand exactly the contribution of each new models to the tasks
at hand, as well as what systems still fail to capture in terms of linguistic
phenomena.

The seminal work of \cite{srlgildeajurafsky2002} provided extensive error
analysis, including performance of the system broken down by feature and roles,
measuring the contribution of each feature to the identification and labeling
of each semantic role. Their work also provided accuracy, recall and $F_1$ scores
for most frequent frame elements, as well as performance measures of each
learning algorithm used. This provided a detailed overview of what their
system was good at (e.g., identifying and labeling \fe{Agent} frame elements),
and what it was bad at (e.g.,
identifying \fe{Manner} and \fe{Location} frame elements). Their analysis of
each feature's coverage and contribution to frame element identification and
labeling helped craft a carefully selected set of features (highly motivated and backed
by experimental data and analysis) which has been
used almost as-is (though extended) in subsequent studies of semantic role
labeling.

Since then, however, very little has been done in terms of error analysis.
If work such as \citep{palmercoverage2010} or \citep{kshirsagaretal2015} do
provide coverage and frequency metrics of lexical units and/or frame elements,
potentially exhibiting correlations between coverage gaps, low frequency
frame elements and low performance of system, we still lack the kind of analysis
done on PropBank work \citep[e.g.][]{woodsend2014}, in order to identify linguistic
knowledge which proves systematically hard to predict by frame semantic parsers.
This is especially true for recent approaches relying on neural networks.
If work such as \cite{fitzgerald2015} or \cite{rothlapata2016} have provided
very useful 2D-graphics of the representations learned by their model,
projections of learned representations
exemplifying how models could operate classification of semantic roles and
group together similar semantico-syntactic configurations, such as nested agents,
relative clauses, complements of nominal predicates, etc., error analysis are
often absent, or limited to the mere comparison of performance based on
sentence length \citep{yangmitchell2017}.

A notable exception to this absence of error analysis is work by
\cite{rothlapata2015}. They were indeed among the only ones to provide
an exhaustive qualitative analysis of the improvements of their system, based
on discourse and context features. They showed for instance that adding contextual
features helped the model capture semantic properties of unseen tokens, such
as being able to properly label `Dec. 1' with the \fe{Time} frame element.
Similarly, their system proved able to capture certain semantic categorizations,
such as the fact that \emph{aunt}, \emph{uncle} and \emph{grandmother} were all
of the \emph{human} semantic type and therefore could more likely fill roles
such as \fe{Recipient} than that of \fe{Goal}. They additionally showed that
their system, and more specifically the contextual features on semantic types
and the discourse feature on salience, better captured the \emph{agency} property
of certain frame elements, which translated into an increase in recall from
56\% to 78\%. They also showed that contextual word representations
lead to better identification of \fe{Time} frame elements, especially
when the label was filled by an infrequent adverb.

\section{Using FrameNet for automatic paraphrase generation}
A major question that arises from previous sections
is whether data in FrameNet contain
many \emph{redundancies} and whether they are \emph{well-balanced} for
parsers to
properly learn useful generalizations, or whether the fault lies in machine
learning systems used so far which prove unable to capture useful generalizations
from the available training data.

In order to test the robustness and ability of statistical models for
frame semantic parsing to extract information from FrameNet annotated data,
we propose, in the next chapters, a model to artificially augment FrameNet
annotation with \emph{paraphrastic}
examples
generated via a rule-based
algorithm combining lexical substitution and valence pattern matching.
The core idea of the model is to generate data reflecting at the surface level
\textendash\ via lexical configurations of predicate-argument structures
\textendash\
structural information encoded in FrameNet through valence patterns.
By recombining information already latently present in FrameNet,
our paraphrastic data augmentation approach should provide a
simple way to understand under which conditions statistical models
for frame semantic parsing fail
to generalize and what kind of additional data, in terms of lexical
predicate-argument configurations, those models require to better learn.

Our paraphrastic data augmentation model uses FrameNet-internal knowledge
to generate near-paraphrases which are in turn used to try and improve
frame semantic parsing. It is worth mentioning that both ideas have antecedents
in the literature: the potential of FrameNet for paraphrasing has already been
thoroughly documented
\citep{ellsworth2007,coyne2009lexpar,hasegawaetal2011}, and several work
have argued that paraphrasing could prove to be beneficial to frame semantic
parsing \citep{rastogiandvandurme2014,pavlicketal2015}.
However, no open-source system exists today to generate paraphrases with
FrameNet, and no previous work have actually \emph{proven} paraphrases to
quantifiably improve
frame semantic parsing performances.

Several other work have attempted to augment training data, sometimes relying
on paraphrase
generation, be it for Propbank \citep{woodsend2014}, or
FrameNet \citep{furstenau2012},
but none of those approaches used resource-internal knowledge to generate
paraphrases and/or augment the training set.
Moreover, they have both exhibited limited improvements,
while significantly increasing the size of the training set.
With our approach, we hope that a system relying on FrameNet-internal
logic to augment the training set will prove more consistent with the
gold annotation and lead to better performances ultimately.

\chapter{Models and Experimental Setup}
\label{modelxpsetup}
In this chapter we introduce two models: (1) a rule-based model for
\textbf{paraphrastic data augmentation}
used for augmenting
our training set of FrameNet annotated data with artificial exemples generated
from the FrameNet database via a valence patterns matching algorithm; and (2)
a statistical model for \textbf{argument identification} as originally defined by
\cite{dasetal2014} and refined by \cite{kshirsagaretal2015}.

In Section~\ref{paraphrasticmodel} we introduce the paraphrastic data augmentation model,
detail its core motivations and implementation specifications, and discuss
its robustness. In Section~\ref{argidentificationmodel} we introduce
the argument identification model, its features, and its learning and decoding
strategies. Additionally, in Section~\ref{xpsetup}, we detail the experimental
setup used throughout this work, in terms of training, development and testing
datasets, toolkits and hyperparameters used for all the models.

\section{Model}
\label{model}

\subsection{Paraphrastic data augmentation}
\label{paraphrasticmodel}

\subsubsection{Philosophy}

Our data augmentation model is grounded in frame semantics theory which
argues that lexical units sharing similar valence
patterns should be considered as semantically equi\-valent~\citep{fillmore1982,rupetal2016}.
This fundamental idea is at the core
of past research which have relied on FrameNet to generate sentential paraphrases
\citep{ellsworth2007,hasegawaetal2011}.
For example, both \lexunit{buy}{v} and
\lexunit{purchase}{v} lexical units in the \semframe{Commerce\_buy} frame
are compatible with the
\vpattern{Buyer.NP.Ext Goods.NP.Obj} valence pattern. Therefore, the following
sentences should be considered as \emph{grammatically} correct and
\emph{semantically} equivalent:\footnote{In all the following examples
 boldface indicates \emph{targets}, i.e. frame-evoking words in context}

\begin{examples}
 \item \vuanno{John}{Buyer}{NP}{Ext} \target{bought} \vuanno{a car}{Goods}{NP}{Obj}
 \item \vuanno{John}{Buyer}{NP}{Ext} \target{purchased} \vuanno{a car}{Goods}{NP}{Obj}
\end{examples}

Note that the semantic equivalence is not strict, as FrameNet may classify
in the same frame lexical units with different semantics \textendash\
sometimes even antonyms \textendash\ provided that those lexical units
share some commonalities
in the semantico-syntactic realizations of their arguments.
For example, both \lexunit{make}{v} and \lexunit{lose}{v} lexical units
belong to the same
\semframe{Earnings\_and\_losses} frame and are compatible with the
\vpattern{Earner.NP.Ext Earnings.NP.Obj} valence pattern, as in:

\begin{examples}
 \item \vuanno{I}{Earner}{NP}{Ext} like working and \target{making}
 \vuanno{money}{Earnings}{NP}{Obj}
 \item \vuanno{I}{Earner}{NP}{Ext} like working and \target{losing}
 \vuanno{money}{Earnings}{NP}{Obj}
\end{examples}

As lexical units with diverging (or even opposite) meanings may share common
valence patterns in FrameNet,
the lexical units generated by our valence pattern matching
algorithm will not
necessarily form \emph{true} or \emph{valid} paraphrases of the original
target. Hence, we call them \emph{paraphrastic} candidates and describe
our data augmentation method as \emph{paraphrastic data augmentation}.

\subsubsection{Implementation}

Our paraphrastic data augmentation is implemented in a system called
\pfn, which workflow is as follows:
\begin{enumerate}
 \item Extract all annotation sets from FrameNet XML files and unmarshall to
       a list of standardized annotation sets for easy manipulation;
 \item Filter the list of standardized annotation sets according to the
       filtering options (detailed in Section~\ref{filteringoptions});
 \item For each filtered annotation set, extract the corresponding
       valence pattern and fetch all the lexical units in training data compatible
       with the source valence pattern.
       The list of lexical units output by the system, enriched with
       the set of annotation labels extracted
       from the original annotation set, will constitute the list of paraphrastic
       candidates;
 \item Filter the list of paraphrastic candidates according to the
       filtering options (detailed in Section~\ref{filteringoptions});
 \item Generate all possible sentences from the list of filtered candidates,
       project FrameNet annotation label to the new sentences, and export
       all newly generated sentences to FrameNet XML format.
\end{enumerate}

\begin{figure}
 \resizebox{\linewidth}{!}{
  \includegraphics{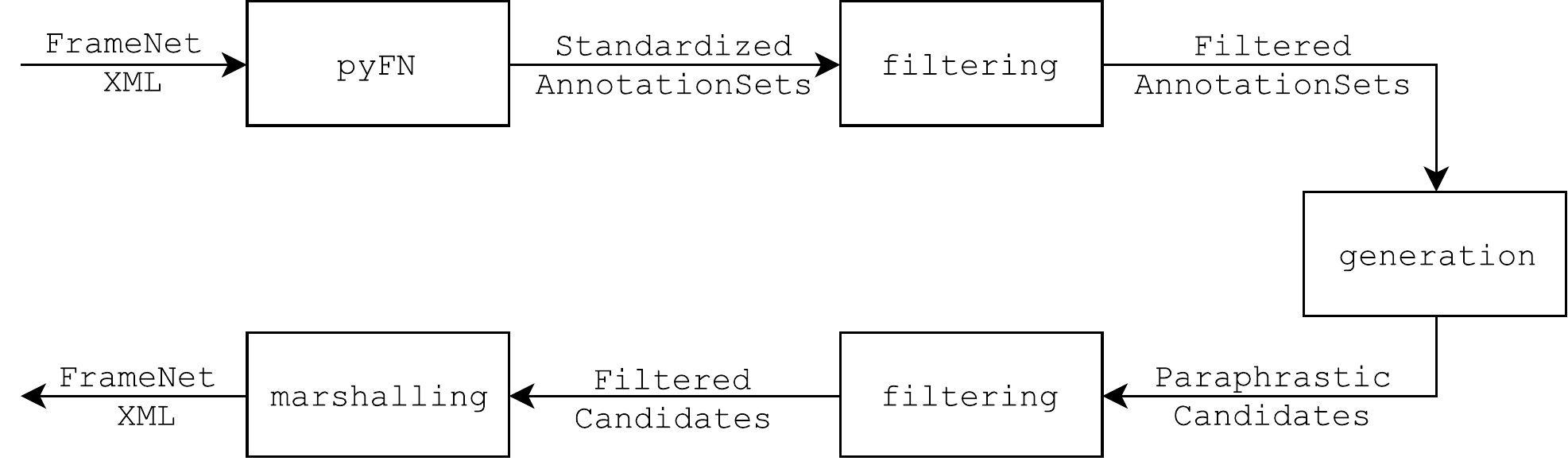}}
 \caption{\pfn paraphrastic data generation pipeline}
 \label{pfnpipeline}
\end{figure}

\begin{figure}
 \resizebox{\linewidth}{!}{
 \includegraphics{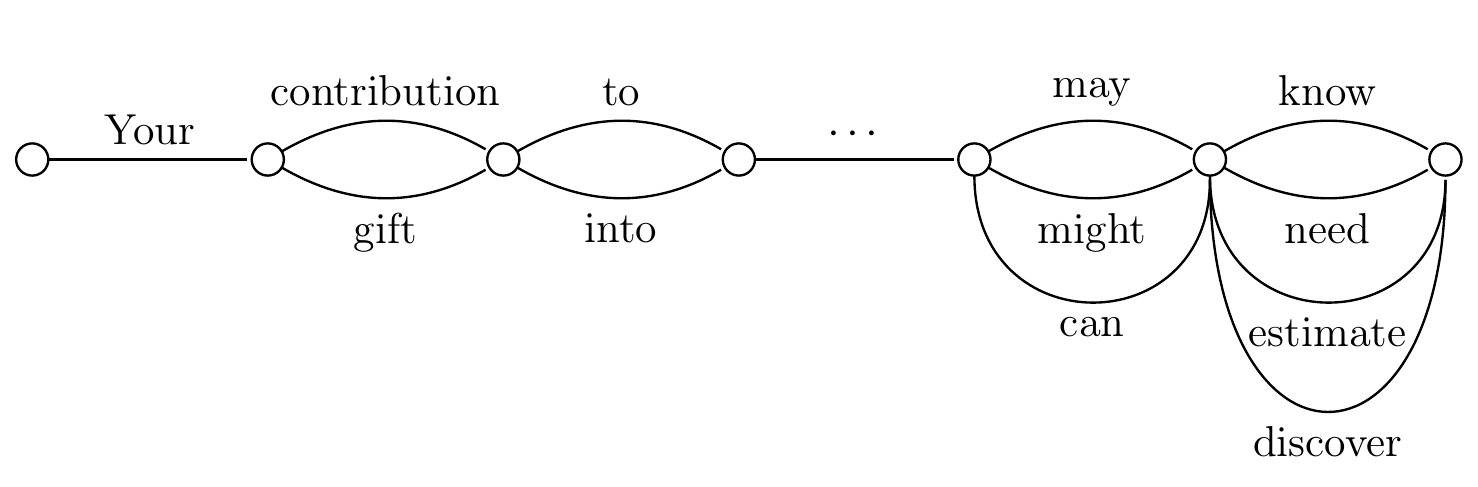}
  %
  %
 }
 \caption{A word lattice showing all paraphrasing candidates output by the
  \pfn system for the sentence: \emph{You contribution to Goodwill will mean more than
  you may know}. Original targets are represented in the upper edges and paraphrastic
 candidates are represented in the lower edges}
 \label{wordlattice}
\end{figure}

Figure~\ref{pfnpipeline} presents the overall pipeline of the \pfn system. The
\texttt{generation} step is where the list of paraphrastic candidates
is generated for a given source annotation set.
To provide a concrete example, consider the following sentence:
\begin{example}
 \label{goodwill}
 Your contribution to Goodwill will mean more than you may know
\end{example}
This sentence has exactly six annotation sets in FrameNet
(one per target):\footnote{for clarity we only show the frame element labels,
omitting phrase types and grammatical functions}
\begin{examples}
 \item \feanno{Your}{Donor} \target{contribution} \feanno{to Goodwill}{Recipient} will mean more than you may know
 \item \feanno{Your contribution}{Trajector} \target{to} \feanno{Goodwill}{Landmark} will mean more than you may know
 \item \feanno{Your contribution to Goodwill}{Means} will \target{mean} \feanno{more than you may know}{Value}
 \item Your contribution to Goodwill will mean \feanno{\target{more} \feanno{than you may know}{Class}}{Added\_set}
 \item Your contribution to Goodwill will mean more than \feanno{you}{Hypothetical\_event} \target{may} \feanno{know}{Hypothetical\_event}
 \item Your contribution to Goodwill will mean more than \feanno{you}{Cognizer} may \target{know}
\end{examples}
The \pfn system will return paraphrastic candidates for only four
of those targets: \emph{contribution}, \emph{to}, \emph{may} and \emph{know}.
Targets for which no paraphrastic candidates are returned usually indicate that
they are the sole lexical units found in the input valence patterns.
The list of paraphrastic
candidates can be represented as a word lattice, as shown in Figure~\ref{wordlattice},
following traditional studies on sentencial paraphrase~\citep[e.g.][]{madnanidorr2010}.

\pfn will perform \emph{loose} valence pattern matching, i.e., two valence patterns
will be considered matching as long as all their \emph{core} valence units\footnote{
the syntactic realizations of the core frame elements} do match.
For example, the valence patterns \vpattern{Earner.NP.Ext Earnings.NP.Obj} and
\vpattern{Earner.NP.Ext Earnings.NP.Obj Time.AVP.Dep} will be considered
matching as the \fe{Time} frame element is peripheral in the
\vpattern{Earner.NP.Ext Earnings.NP.Obj Time.AVP.Dep} valence pattern.
This approach has the advantage of augmenting the pool of candidates generated
by the system when only a limited number of valence patterns has been annotated for a given
lexical unit and when those valence patterns contain peripheral frame elements.

The number of possible paraphrastic \textbf{sentences} $N_p$ output by the
\pfn system for a given
sentence $s$ containing $N_a$ annotation sets is given by:
\begin{equation}
 N_p = \prod_{j}^{N_{a}} \binom{N_{c_j} + 1}{1} - 1
\end{equation}
where $N_{c_{j}}$ is the number of paraphrastic candidates generated
by \pfn for the the $j$-th annotation set. $N_p$ simplifies to:
\begin{equation}
 N_p = \prod_{j}^{N_a} (N_{c_j} + 1) - 1
\end{equation}
For sentence \ref{goodwill}, and without any filtering, \pfn will therefore generate
47 paraphrastic sentences:
\begin{equation}
 N_p = 2 \cdot 2 \cdot 1 \cdot 1 \cdot 3 \cdot 4 - 1 = 47
\end{equation}

\subsubsection{Robustness}
\label{robustness}

Our approach for augmenting FrameNet annotation via paraphrastic sentences
is more robust than what previous approaches to automatic FrameNet data
augmentation have attempted, as it guarantees strong consistency with
the underlying FrameNet annotation framework.
Indeed, past research have often combined word-based semantic similarity
with dependency-tree-based syntactic similarity to project annotation
to new data, be it on FrameNet \citep{furstenau2012},
or PropBank \citep{woodsend2014}.

However, the challenge of such approaches is to be able to reproduce
the structure of FrameNet annotation.
Consider first the following example:
\begin{examples}
 \item \vuanno{John}{Buyer}{NP}{Ext} \target{bought}
 \vuanno{a car}{Goods}{NP}{Obj} this morning
 \item Mary \target{acquired} a house yesterday
\end{examples}
Although both \semlexunit{Commerce\_buy}{buy}{v}  and
\semlexunit{Getting}{acquire}{v} lexical units share some apparent syntactic
and semantic similarities, FrameNet clearly distinguishes them
and assigns them to different frames with distinct
frame elements in order to account explicitly for the semantic specificity of
the lexical unit \lexunit{buy}{v} which implies an exchange of money for
the thing acquired. Consider then the following example:
\begin{examples}
 \item \vuanno{I}{Ingestor}{NP}{Ext} \target{ate} \vuanno{cheese}{Ingestibles}{NP}{Obj}
 \item \vuanno{I}{Ingestor}{NP}{Ext} \target{drank} \vuanno{wine}{Ingestibles}{NP}{Obj}
\end{examples}
Although both \lexunit{eat}{v} and \lexunit{drink}{v} lexical units share
syntactic similarities, they appear somehow slightly more distant semantically
than \lexunit{buy}{v} and \lexunit{acquire}{v}. They are nonetheless annotated
within the same \semframe{Ingestion} frame
in FrameNet, and are both compatible with the
\vpattern{Ingestor.NP.Ext Ingestibles.NP.Obj} valence pattern.

The dilemma for systems combining word-based semantic similarity
and dependency tree-based syntactic similarity is therefore to be able to
handle \emph{both} eat/drink and buy/\\acquire cases, by including one while
excluding the other. However, a quick test with
a cosine similarity calculated on distributional representations
computed with \texttt{word2vec}\footnote{See Section~\ref{filteringoptions}
for details}
gives us a 0.57 similarity score between \emph{buy} and \emph{acquire} and
a 0.5 similarity score between \emph{eat} and \emph{drink}. We see directly how
such a measure would lead to filtering out both cases or including them both
in the augmented data, given that the output hierarchy of similarity is the
inverse of that of FrameNet.

Although our approach does overcome those limitations,
as it generates artificial data fully compliant with FrameNet annotation
\emph{by construction}, it still faces two major shortcomings:
\begin{enumerate}
 \item as previous approaches, and without filtering, it generates a huge
       number of paraphrastic sentences, which tend to explode the training set
       beyond what is computationally reasonable;
 \item due to FrameNet's annotating dissimilar
       lexical units with similar valence patterns,
       the \pfn system may generate sentences which are unlikely to be seen in
       a corpus, as:
       \begin{example}
        ? I ate wine
       \end{example}
\end{enumerate}

The key question is therefore whether or not the statistical classifier used
for performing argument identification is sensitive to lexical diversity
at the argument level \textendash\ that is, to lexical co-ocurrence (as suggested by the
lexical features detailed in Figure~\ref{semaforfeatures}) \textendash\
which is what our paraphrastic data augmentation brings.
We hypothesized two different scenarios:
\begin{enumerate}
 \item the statistical classifier relies primarily on a latent representation of
       the arguments, and is less sensitive to predicate-argument lexical configurations,
       in which case we should be able to observe an improvement in global argument identification
       scores when training on our augmented training set, even without filtering;
 \item the statistical classifier is sensitive to predicate-argument lexical configurations,
       to the point that bias introduced by unlikely lexical configurations does not compensate
       for the lexical diversity brought at the argument level. In this case we
       should observe a decrease in global argument identification
       scores when training on our augmented training set without filtering.
\end{enumerate}
Following our first results which confirmed scenario 2 (see Chapter~\ref{results}),
we decided to implement an extra layer of filtering
in order to filter out unlikely lexical configurations.

\subsubsection{Filtering}
\label{filteringoptions}
The \pfn pipeline includes two different kinds of filtering:
\begin{enumerate}
 \item \textbf{annotation sets filtering}: source annotation sets are
       filtered based on the part of speech or number of tokens of their
       lexical units. Filtering out items according to part of speech allows for
       evaluating the contribution of paraphrastic data augmentation for each
       morphosyntactic category separately. Filtering out multi-lexeme lemmas
       allows removing configurations not supported by the \semafor parser
       which we use for argument identification;\footnote{
        \semafor does not support distant multi-lexeme
        lemmas such as phrasal verbs separated by nouns or adjectives,
        e.g. ``\target{give} the answer \target{away}''}
 \item \textbf{paraphrastic candidates filtering}: paraphrastic candidates
       are filtered based on the cosine similarity measure between their
       distributional
       representation and the distributional representation of the source
       lexical unit.
\end{enumerate}
The aforementioned \emph{distributional representations} refer to
vectorial representations of the meaning of words where the dimensions of
the vectors record contextual information about the co-occurrence
of target words with their neighboring words as observed in large text corpora.
The distributional representations used for paraphrastic candidates filtering
are computed from \emph{predictive models} and \emph{count-based models}.
Predictive models estimate word vectors via a supervised task where the
weights of the word
vector are set to maximize the probability of the contexts in which the word
is observed in the corpus. Count-based models compute word vectors by extracting
co-occurrence counts of target words and their contexts. The
\emph{cosine similarity}
measure used is defined as the dot
product of two vectors normalized by the product of vectors length.
Given two distributional representations $\overrightarrow{u} = \langle
u_1,\dots, u_n \rangle$ and $\overrightarrow{v} = \langle v_1, \dots, v_n
\rangle$, the cosine similarity between $\overrightarrow{u}$ and
$\overrightarrow{v}$ is formally defined as:
\begin{equation}
 cos(\overrightarrow{u},\overrightarrow{v}) =
 \frac{\sum_{i}{u_i\cdot v_i}}{||\overrightarrow{u}|| \cdot ||\overrightarrow{v}||}
\end{equation}
The semantic filter component of the paraphrastic candidates filter has three
possible configurations:
\begin{enumerate}
 \item \textbf{random} which selects $n$ random paraphrastic candidates in the list;
 \item \textbf{word2vec} which filters out paraphrastic candidates based
       on a \texttt{word2vec} predictive model \citep{mikolovetal2013} trained on the GoogleNews
       corpus;\footnote{See details at \url{https://code.google.com/archive/p/word2vec/}}
 \item \textbf{dissect} which filters out paraphrastic candidates using
       the \texttt{dissect} toolkit \citep{dissect2013} with two different models:
       a \texttt{word2vec} predictive model trained on a concatenation
       of the ukWaC, English Wikipedia, and BNC corpora, and a (reduced) count-based
       model generated from the same concatenated corpus
       \citep[see][for details]{baronietal2014}.
\end{enumerate}
Both \textbf{word2vec} and \textbf{dissect} configurations have two additional
setups:
\begin{enumerate}
 \item \textbf{top}: which selects the top $n$ elements from a list of
       paraphrastic candidates ranked in a decreasing order
       according to the cosine similarity between their distributional
       representations
       and the distributional representation of the original target;
 \item \textbf{threshold} which selects all paraphrastic candidates for which
       the cosine similarity between their distributional representations
       and the distributional representation of the original target is above a
       specified value.
\end{enumerate}

\subsection{Argument identification}
\label{argidentificationmodel}
For argument identification we rely on the log-linear model of \cite{dasetal2014},
modified by \cite{kshirsagaretal2015} to reduce training time and enable training
on exemplar sentences.

\subsubsection{Model}
The model performs argument span identification and argument role labeling
jointly: all possible spans are pre-identified with a rule-based algorithm
and the system then assigns a role to each span from
the set of frame elements belonging to the
target predicate's frame, to which has been added the null role $\varnothing$
in case no role is to be assigned to the span.

\noindent
\textbf{Argument span identification}
is performed by identifying all continuous spans in a sentence
that contain a single word or
comprise a valid subtree of a word and all its descendants in the dependency
parse produced by the dependency parser. This bounds recall on argument identification
as it covers only 80\% of arguments in the FrameNet dataset.

\noindent
\textbf{Argument role labeling}
is performed with a log-linear model as follows:
Let $x$ be a dependency-parsed sentence, $p$ the target predicate and $f$
the frame evoked by $p$. Let $spans(x,p,f)$ be the set of candidate spans
previously identified. For each candidate argument $a \in spans(x,p,f)$ and
each role (frame element) $r$, a binary feature vector $\phi(a,x,p,f,r)$ is
extracted. Let $w$ be the weight matrix. Each
argument $a$ is given a real-valued score by a linear model:
\begin{equation}
 score_w(a|x,p,f,r) = w^{\top}\phi(a,x,p,f,r)
\end{equation}

\subsubsection{Features}

\begin{figure}
 \resizebox{\linewidth}{!}{
  \includegraphics{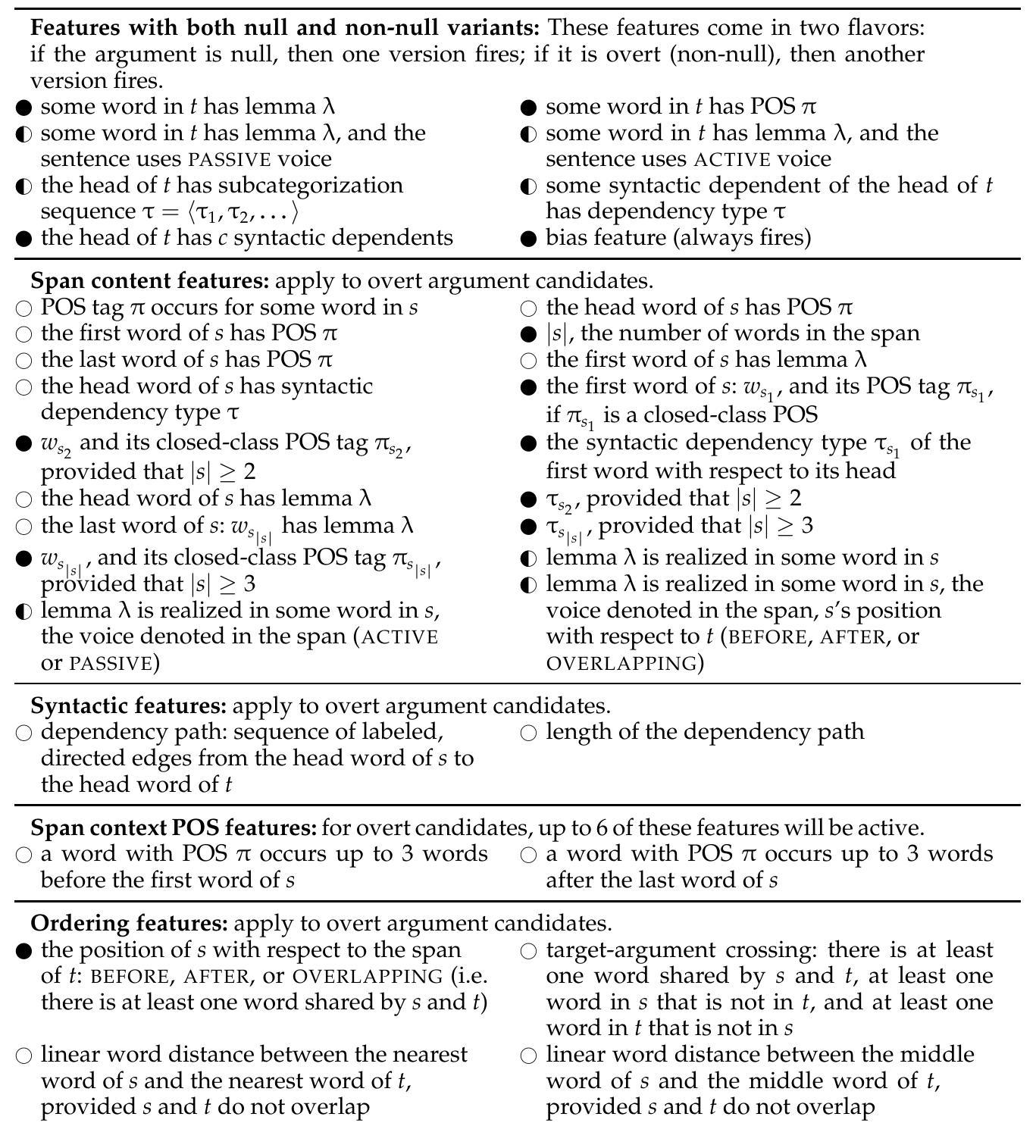}}
 \caption{Argument identification feature templates as of \cite{dasetal2014}}
 \label{semaforfeatures}
\end{figure}

Figure~\ref{semaforfeatures} shows the list of feature templates used by the model.
Every feature template has a $\Circle$ version that does not take into account
the role being
filled (so as to incorporate overall biases). The $\LEFTcircle$ symbol indicates that
the feature template also has a variant that is conjoined with the frame
element name being filled, and $\CIRCLE$ indicates that the feature
template additionally has a variant that is conjoined with both the frame element
name and the frame name.

\subsubsection{Learning}
The model is trained using a local objective where each span-role pair is
treated as an independent training instance.
Consider the following squared structured hinge loss function $L_w$ given
for a training example $i$:
\begin{equation}
 L_w(i) = \bigg(\max_{a'}(w^{\top}\cdot\phi(a', x, p, f, r) + 1\{a' \neq a \})
 - w^{\top}\cdot\phi(a, x, p, f, r) \bigg)^2
\end{equation}
The model parameters are learned by minimizing the $l_2$-regularized average loss using
the online optimization method AdaDelta \citep{zeiler2012}:

\begin{equation}
 w^{\star} = \argmin_w \frac{1}{N} \sum_{i=1}^{N}L_w(i) + \frac{1}{2}\lambda \|w\|_2^2
\end{equation}

\subsubsection{Decoding}
Decoding is done using beam search which produces a set of k-best hypotheses
of sets of span-role pairs. The approach enforces multiple constraints including
the fact that a frame element label may be assigned to at most one span and
that spans of overt arguments must not overlap.

\section{Experimental setup}
\label{xpsetup}

\subsection{Datasets}
\label{xpsetupdatasets}

In this work we diverge from previous approaches in that we rely on the latest
FrameNet 1.7 data release instead of FrameNet 1.5.
We split data in three sets: train, dev, test for
training, development and testing. For testing, we use the same list of
fulltext documents as \cite{dasetal2014} used for testing on FrameNet 1.5.
Using FrameNet 1.7 has the benefit of almost doubling the size of the testing set,
providing thereby a more robust set to test model performances.
Table~\ref{tabledatasetsdev} shows the number of sentences and annotation sets
in the train/dev/test setting on FrameNet 1.7. Table~\ref{tablecomparisondatasets}
shows the same metrics on FrameNet 1.5 and 1.7 in a train/test setting where
the content of the development set is included in the training set.
Additional data such as counts and coverage metrics for lexical units,
frame elements, valence units and valence patterns are provided
in Chapter~\ref{analysis}.

\begin{table}[h!tpb]
 \centering
 \begin{tabular}{lrr}
  \toprule
  \multicolumn{3}{c}{\textsc{Framenet 1.7 dataset metrics}}\\
  \midrule
                                       & \#sentences       & \#annotationsets   \\
  \midrule
  \textsc{train (fulltext + exemplar)} & \numprint{168237} & \numprint{184 050} \\
  \textsc{train (fulltext)}            & \numprint{2804}   & \numprint{16215}   \\
  \textsc{dev}                         & \numprint{887}    & \numprint{5715}    \\
  \textsc{test}                        & \numprint{1247}   & \numprint{6517}    \\
  \bottomrule
 \end{tabular}
 \caption{FrameNet 1.7 datasets metrics with development and testing sets.
  The training
 set does not include the development set}
 \label{tabledatasetsdev}
\end{table}

\begin{table}[h!tpb]
 \centering
 \begin{tabular}{lrr}
  \toprule
  \multicolumn{3}{c}{\textsc{Framenet 1.5 and 1.7 datasets metrics}}\\
  \midrule
                           & \#sentences       & \#annotationsets  \\
  \midrule
  \textsc{train FT 1.5}    & \numprint{2983}   & \numprint{18228}  \\
  \textsc{train FT 1.7}    & \numprint{3691}   & \numprint{20886}  \\
  \textsc{train FT+EX 1.5} & \numprint{150735} & \numprint{168357} \\
  \textsc{train FT+EX 1.7} & \numprint{169124} & \numprint{189553} \\
  \textsc{test 1.5}        & \numprint{875}    & \numprint{3702}   \\
  \textsc{test 1.7}        & \numprint{1246}   & \numprint{6040}   \\
  \bottomrule
 \end{tabular}
 \caption{Metrics comparison between FrameNet 1.5 and FrameNet 1.7 datasets
 with no development set}
 \label{tablecomparisondatasets}
\end{table}

\subsection{Toolkits}
Over the course of our research, we had to develop and entire toolkit
so as to replicate past results, implement our models and compare ourselves
to previous approaches. The following sections provide a short description of
this toolkit.

\subsubsection{Querying valence patterns}
As valence patterns were not accessible out-of-the-box in FrameNet, due to
their being deeply embedded in the structure of FrameNet XML and split into
lists of valence units,
we had to implement a set of tools in order to be able to efficiently
search for valence patterns
in the FrameNet dataset. For performance reasons, we decided to rely on a NoSQL
MongoDB database which provided an optimal architecture with indexed
valence units and valence patterns against which FrameNet data could be queried directly.
We named our application to import FrameNet XML data to MongoDB
\noframenet.\footnote{available at: \url{https://github.com/akb89/noframenet}}
We then implemented a search engine API to query FrameNet valence patterns in
MongoDB. The API would receive HTTP GET requests and
return JSON-formatted FrameNet
data. We named this API \valencer\footnote{available at: \url{https://github.com/akb89/valencer}}
and published it as a system demonstration during the COLING 2016 conference
\citep[see][]{kabbach2016}.
To be able to easily visualize the output of our API, we also implemented
a web client on top of the \valencer API. We named it
\myvalencer\footnote{available at: \url{https://github.com/akb89/myvalencer}}
and published it at the eLex 2017 conference as a system demonstration
\citep[see][]{kabbach2017}.

\subsubsection{From \semafor to \rofames}
We forked the \texttt{acl2015} branch of the \semafor\ parser of \cite{dasetal2014},
corresponding to
the version of the parser used in~\cite{kshirsagaretal2015}.
We re-implemented the whole
preprocessing pipeline to generate the train, dev and test splits from FrameNet
XML data as it was no longer available in that version of \semafor. This
enabled us to train and test on FrameNet 1.7 data in addition to 1.5.
We added a branch for the system where we removed the hierarchy feature
introduced by~\cite{kshirsagaretal2015}
in order to have a version of
\semafor\ corresponding to previous baselines but we kept the changes in
training as it reduced training time by
about one order of magnitude and made it possible to train on exemplar sentences
as well as fulltext.
We named our forked version \rofames, the semordnilap of \semafor.

\subsubsection{Generating paraphrastic examples and debugging \rofames}
We implemented a Python wrapper, named \pyfn, to handle FrameNet XML data in a more systematic
fashion. This proved particularly helpful as we discovered that FrameNet XML
data contained many corrupted entries (such as annotation with partially missing
label indexes), which bugged our paraphrastic data augmentation system
\pfn (see Chapter~\ref{baseline} for details). In addition to \pyfn and \pfn,
we also implemented a debugging application named \dfn to compute several metrics
such as coverage and precision/recall/$F_1$ scores per frame element given
FrameNet data and the model output by \rofames.

\subsection{Hyperparameters}

In all our experiments, \rofames is used with $\lambda = 1e^{-6}$ and a beam
of 100. Data are tokenized and part-of-speech tagged with NLP4J \citep{choi2016}
and dependency-parsed with BMST \citep{kiperwasser2016}. The \valencer API
throws requests at a MongoDB database containing the exact same data as in
the training set. We experiment \pfn configurations with semantic filtering with
\texttt{word2vec} predict
and \texttt{dissect} reduced count models, all in random mode with random values 2,3,4,
in top mode 2,3,4 and in threshold mode 0.5 and 0.7. We filter part of speech
to experiment data augmentation on nouns only, verbs only, adjectives only
and nouns plus verbs.

\chapter{Baseline and Replication}
\label{baseline}

In this chapter we report on our attempts to replicate several past results
of argument identification
with our FrameNet 1.7 datasets. We also discuss several crucial experimental
considerations regarding the production of a robust baseline
for argument identification on FrameNet.
In Section~\ref{validationprepro}
we report on results validating our preprocessing implementation in \rofames.
In Section~\ref{impactpostagger} and Section~\ref{impactdepparser},
we discuss the impact of part-of-speech
taggers and dependency parsers on argument identification.
Next, in Section~\ref{impactexemplar} and Section~\ref{impacthierarchy}
we report on our attempts
at replicating results from \citep{kshirsagaretal2015} with exemplar data and
the hierarchy feature.
Finally, in Section~\ref{baselinescoressection} we summarize our results and
compile them into
a list of recommendations providing a robust baseline against which we will be
able to compare our paraphrastic data augmentation approach in
Chapter~\ref{results}.

\section{Validating the preprocessing pipeline}
\label{validationprepro}
As we re-implemented the preprocessing pipeline of \rofames, we first
tested the robustness of our implementation by replicating previous baseline
results based on the FrameNet 1.5 splits of \cite{dasetal2014}.
We trained \rofames on our train split but tested on the test split
of \cite{dasetal2014}. Our train split differed from that of \cite{dasetal2014}
in that we applied strict filtering of potentially erroneous annotation in order
not to bias the parser: we removed all annotation sets which
contained missing or invalid start/end indexes (except for null instantiations).
Splits were tokenized using Robert MacIntyre's sed script, part-of-speech tagged
with MXPOST~\citep{mxpost1996} and dependency-parsed with the
MSTParser~\citep{mst2006}, following the original pipeline of \cite{dasetal2014}.
For scoring, we used the two evaluation scripts mentioned in Section~\ref{taskevaluation}:
\begin{enumerate}
 \item \textbf{SEM}: the SemEval 2007 evaluation script as-is~\citep{fnsemeval2007}
 \item \textbf{ACL}: the modified SemEval 2007 evaluation script used by ~\cite{kshirsagaretal2015},
       which does not give extra credits for (gold) frames and therefore better accounts
       for argument identification performances (refer to Section~\ref{taskevaluation} for details)
\end{enumerate}

Scores, detailed in Table~\ref{preprocessingvalidation}, validate our
implementation.
Results output by the \sem scoring script are consistent with the
baseline reported in~\citep{rothlapata2015}.
Regarding results output by the \acl script,
we found a 1 point $F_1$ gain over the baseline reported
in~\citep{kshirsagaretal2015}. Both approaches used the train and test splits
of \cite{dasetal2014}.
Note that we found the decrease in precision to be consecutive to the increase
in recall.

\begin{table}[h!tpb]
 \centering
 \begin{tabular}{l@{\hskip 5cm}ccc}
  \toprule
  \multicolumn{4}{c}{\textsc{Validating \rofames preprocessing pipeline on FrameNet 1.5}}\\
  \midrule
  \textsc{SEM}               & P    & R    & F$_1$ \\
  \midrule
  \#1 ROFAMES + MXPOST + MST & 76.9 & 74.4 & 75.6  \\
  \cite{rothlapata2015}      & 78.4 & 73.1 & 75.7  \\
  \midrule
  \textsc{ACL}               & P    & R    & F$_1$ \\
  \midrule
  \#1 ROFAMES + MXPOST + MST & 64.3 & 56.4 & 60.1  \\
  \cite{kshirsagaretal2015}  & 65.6 & 53.8 & 59.1  \\
  \bottomrule
 \end{tabular}
 \caption{Argument identification results with gold
  frames output by the \sem\ and \acl\ scoring scripts. Train and test splits are
  generated from FrameNet 1.5 fulltext XML data. Train split is produced
  by \rofames, test split is that of \cite{dasetal2014} as-is.
  All data are part-of-speech tagged with MXPOST and
 dependency parsed with the MSTParser}
 \label{preprocessingvalidation}
\end{table}

\section{Impact of part-of-speech taggers}
\label{impactpostagger}
Past studies on frame semantic parsing have used various preprocessing
toolkits. In this section, we focus on the impact of part-of-speech tagger
performances on argument identification, comparing MXPOST~\citep{mxpost1996}
originally used in \citep{dasetal2014} with
NLP4J~\citep{choi2016} used in \citep{roth2016}.
Results, reported in Table~\ref{postaggers}, show a systematic $F_1$ gain
in argument identification when relying on the NLP4J part-of-speech tagger.
Improvements gains vary from 0.3 $F_1$ in test to 0.7 $F_1$ in dev, with a systematic
improvement in recall. In subsequent experiments, we rely systematically on
NLP4J for part-of-speech tagging.

\begin{table}[h!tpb]
 \centering
 \begin{tabular}{ll@{\hskip 5cm}ccc}
  \toprule
  \multicolumn{4}{c}{\textsc{Comparing pos taggers on FrameNet 1.7}}\\
  \midrule
      & \textsc{dev}  & P    & R    & F$_1$          \\
  \midrule
  \#7 & MXPOST + MST  & 56.3 & 51.0 & 53.5           \\
  \#8 & NLP4J + MST   & 56.6 & 52.0 & $54.2^{\dagger}$ \\
  \midrule
      & \textsc{test} & P    & R    & F$_1$          \\
  \midrule
  \#7 & MXPOST + MST  & 57.8 & 54.1 & 55.9           \\
  \#8 & NLP4J + MST   & 57.7 & 54.7 & $56.2^{\ast}$    \\
  \bottomrule
 \end{tabular}
 \caption{Argument identification results with gold frames.
  Train, dev and test splits are part-of-speech tagged with
  MXPOST or NLP4J
  and dependency parsed with the MSTParser. Statistical significance is indicated
  by $\dagger$ for $p < 0.05$ and $\ast$ for $p < 0.10$}
 \label{postaggers}
\end{table}

\section{Impact of dependency parsers}
\label{impactdepparser}
Similarly, past studies on frame semantic parsing have used various
dependency parsers. Here, we compare the MSTParser originally
used \cite{dasetal2014} to the BIST parser~\citep{kiperwasser2016} in two of its
variants: the transition-based BARCH parser and the graph-based BMST parser.
The BIST parser was originally used by \cite{roth2016} for frame semantic
parsing.
Results, presented in Table~\ref{dependencyparsers}, show a systematic
improvement on argument identification when preprocessing data with
the BIST parser rather than the MSTParser. $F_1$ gains vary from 1.3 to
2.6 depending on the scoring set and the BIST parser variant used.
In subsequent experiments, we rely systematically on the BMST variant of
the BIST parser for dependency parsing. We favor it over the BARCH variant
as it is faster and provides only slighly lower performances.

\begin{table}[h!tpb]
 \centering
 \begin{tabular}{ll@{\hskip 5cm}ccc}
  \toprule
  \multicolumn{5}{c}{\textsc{Comparing dependency parsers on FrameNet 1.7}}\\
  \midrule
       & \textsc{dev}  & P    & R    & F$_1$ \\
  \midrule
  \#8  & NLP4J + MST   & 56.6 & 52.0 & 54.2  \\
  \#9  & NLP4J + BMST  & 59.4 & 52.6 & 55.8  \\
  \#10 & NLP4J + BARCH & 58.9 & 52.5 & 55.5  \\
  \midrule
       & \textsc{test} & P    & R    & F$_1$ \\
  \midrule
  \#8  & NLP4J + MST   & 57.7 & 54.7 & 56.2  \\
  \#9  & NLP4J + BMST  & 60.5 & 55.6 & 58.0  \\
  \#10 & NLP4J + BARCH & 60.9 & 56.8 & 58.8  \\
  \bottomrule
 \end{tabular}
 \caption{Argument identification results with gold frames.
  Train, dev and test splits are
  part-of-speech tagged with NLP4J
  and dependency parsed with the MSTParser or the BIST
 parser in its BMST or BARCH variant}
 \label{dependencyparsers}
\end{table}

\section{Impact of exemplar annotation}
\label{impactexemplar}
In the next two sections, we report on our attempts to replicate
results from \cite{kshirsagaretal2015}. We first include exemplar
data in the training set.
Results, displayed in Table~\ref{exemplars}, show a systematic 1.2 $F_1$ gain
over training on fulltext data only. This is, however, significantly lower
that the 2.8 $F_1$ gain on FrameNet 1.5 data reported by~\cite{kshirsagaretal2015}.
In all the following sections, we found $F_1$ gains on our FrameNet 1.7 splits
to be systematically lower that those reported in \citep{kshirsagaretal2015}.
We hypothesize that those may be due to the increased size of the development
and testing sets, the reduced size of the training set when testing with the
development set, and the increased robustness of our train, dev and test splits
which do not contain duplicate annotation sets likely to bias scores, and no
annotation sets with missing or invalid indexes.

\begin{table}[h!tpb]
 \centering
 \begin{tabular}{ll@{\hskip 5cm}ccc}
  \toprule
  \multicolumn{5}{c}{\textsc{Adding exemplar data on FrameNet 1.7}}\\
  \midrule
       & \textsc{dev}  & P    & R    & F$_1$ \\
  \midrule
  \#9  & FT            & 59.4 & 52.6 & 55.8  \\
  \#12 & FT + EX       & 58.4 & 55.7 & 57.0  \\
  \midrule
       & \textsc{test} & P    & R    & F$_1$ \\
  \midrule
  \#9  & FT            & 60.5 & 55.6 & 58.0  \\
  \#12 & FT + EX       & 59.5 & 58.9 & 59.2  \\
  \bottomrule
 \end{tabular}
 \caption{Argument identification results with gold frames. \rofames is
 trained on fulltext (FT) and exemplar (EX) data}
 \label{exemplars}
\end{table}

\section{Impact of the hierarchy feature}
\label{impacthierarchy}
Additionally, still following \cite{kshirsagaretal2015}, we trained \rofames on
FrameNet 1.7 data with the hierarchy feature comprising Inheritance and
SubFrame relations. Results, displayed
in Table~\ref{hierarchyandexemplars}, show a $F_1$ gain of 0.5 to 0.7
when using the hierarchy feature, again lower than the 1.3 gain reported
by \cite{kshirsagaretal2015} on FrameNet 1.5. When incorporating both
the hierarchy feature and exemplar data to the setup, we show a
2.1 to 2.4 $F_1$ gain, again significantly lower than the 4 points gain reported
by \cite{kshirsagaretal2015}.

\begin{table}[h!tpb]
 \centering
 \begin{tabular}{ll@{\hskip 5cm}ccc}
  \toprule
  \multicolumn{5}{c}{\textsc{Adding the hierarchy feature on FrameNet 1.7}}\\
  \midrule
       & \textsc{dev}  & P    & R    & F$_1$           \\
  \midrule
  \#9  & FT            & 59.4 & 52.6 & 55.8            \\
  \#13 & FT + H        & 60.0 & 53.5 & $56.5^{\dagger}$  \\
  \#14 & FT + H + EX   & 59.2 & 57.2 & $58.2^{\ddagger}$ \\
  \midrule
       & \textsc{test} & P    & R    & F$_1$           \\
  \midrule
  \#9  & FT            & 60.5 & 55.6 & 58.0            \\
  \#13 & FT + H        & 61.0 & 56.3 & $58.5^{\ast}$     \\
  \#14 & FT + H + EX   & 60.3 & 60.0 & $60.1^{\ddagger}$ \\
  \bottomrule
 \end{tabular}
 \caption{Argument identification results with gold frames.
  \rofames is trained on fulltext (FT) and exemplar (EX) data, with and without
  the hierarchy feature (H) of~\cite{kshirsagaretal2015}.
  Statistical significance is indicated
  by $\ddagger$ for $p < 0.01$, $\dagger$ for $p < 0.05$
  and $\ast$ for $p < 0.10$}
 \label{hierarchyandexemplars}
\end{table}

\section{Baseline}
\label{baselinescoressection}
All subsequent experiments in Chapter~\ref{results} rely on the
baseline scores shown in Table~\ref{baselinescores}, computed
from FrameNet 1.7 splits. We built on results
of previous sections in order to produce the most robust
baseline:
\begin{itemize}
 \item for each train/dev/test split, all annotation sets with
       missing or invalid start/end indexes are removed (except for null instantiations);
 \item all duplicate annotation sets and sentences are removed from the test split;
 \item incomplete annotation \textendash\ targets with annotated frames but no
       annotated frame elements \textendash\ are removed from the training set;
 \item data are part-of-speech tagged with the NLP4J tagger~\citep{choi2016};
 \item data are dependency-parsed with the BIST BMST parser~\citep{kiperwasser2016}.
\end{itemize}

\begin{table}[h!tpb]
 \centering
 \begin{tabular}{ll@{\hskip 7cm}ccc}
  \toprule
  \multicolumn{5}{c}{\textsc{Argument identification baseline on FrameNet 1.7}}\\
  \midrule
      &               & P    & R    & F$_1$ \\
  \midrule
  \#9 & \textsc{dev}  & 59.4 & 52.6 & 55.8  \\
  \#9 & \textsc{test} & 60.5 & 55.6 & 58.0  \\
  \bottomrule
 \end{tabular}
 \caption{Baseline scores for argument identification with gold frames
  on dev and test splits, scored with the \acl script.
  Train, dev and test splits are
  part-of-speech tagged with NLP4J
  and dependency parsed with the BIST BMST
 parser}
 \label{baselinescores}
\end{table}

\chapter{Results}
\label{results}

In this chapter, we report on the contribution of our paraphrastic data
augmentation model described in Chapter~\ref{modelxpsetup} to argument
identification, evaluated against the baseline described in
Chapter~\ref{baseline}. Our preliminary results, obtained by augmenting nouns
and verbs without filtering paraphrastic candidates (see
Section~\ref{nofilteringnv}), confirm the need for additional (semantic)
filtering: both approaches lead to a systematic decrease in $F_1$ score compared
to the baseline, despite the significant increase in size of the datasets
produced, almost one order of magnitude higher than the baseline in the case of
augmented verbs.

Our semantic filtering approach,
detailed in Section~\ref{semanticfileringall}, shows mixed results:
on augmented nouns and adjectives, it yields no significant improvements,
but it does not lead to significant decreases in performance either.
On augmented verbs, it does yield limited improvements, in the order of
.5 $F_1$ gain in the best scenario (XP\#41). However, none of the reported
improvements prove statistically significant in the end, with $p$ values
systematically above 0.1, except for XP\#29 ($p = 0.08$).

If, overall, our results tend to demonstrate the smartness of the
\rofames parser,
the marginal improvements shown upon training on augmented verbs,
albeit not statistically significant, suggest that
there could be specific
patterns that \rofames fails
to extract from gold data alone.

Similarly, our results tend to support the robustness of our paraphrastic
data augmentation approach: almost none of the approaches relying on filtering
of paraphrastic candidates lead to a decrease in performances, suggesting that
the core \pfn logic, combined with the right candidates filtering,
rarely introduces noise in the data. 

However, those results should again lead us to tread carefully, as no-filtering
approaches do show significant decreases in performance, suggesting
that paraphrastic approaches relying on FrameNet annotation logic alone
have a natural tendency to generate (noisy) unlikely predicate-argument lexical
configurations, to which the \rofames parser is not robust.

It would be tempting at first to conclude from our results
that semantic filtering of
paraphrastic candidates, motivated by semantic proximity measures computed
from distributional representations of words, is capable of compensating
for the natural tendency of \pfn to generate unlikely lexical configurations.
However, semantic filtering approaches do not actually prove any better
than random filtering approaches, as shown in Section~\ref{randomversusall}.

What is to be made of the observations hereof? We postulate two interpretations,
motivated by two distinct phenomena: (1) there exists a latent variable, not
necessarily semantically motivated, or at least not in ways distributional
models can capture, that determines, among the set of unfiltered \pfn candidates,
what is noise from what is not; or (2) data generated by \pfn are actually
always somehow noisy, but the \rofames parser is robust enough that it can
compensate for the bias introduced by noisy artificial data,
provided that those noisy data remain limited in size.

All in all, the results and preliminary explanations presented in this chapter
do not suffice to arbitrate, in light of the intricacies
that arise. Further systematic analysis of datasets and errors
are required in order to
properly guide reflection. Those analysis
will be presented in Chapter~\ref{analysis}, and further discussed in
Chapter~\ref{discussion}.

\section{No filtering setup}
\label{nofilteringnv}
In the first two experiments we augmented the FrameNet fulltext training set
with \pfn-generated data produced with minimal filtering:
we kept only noun and verb annotation sets and did not apply any semantic
filtering on candidates.
Our preliminary results for augmented nouns and verbs are shown in
Table~\ref{nofilteringresults}. In all subsequent tables,
FT refers to FrameNet fulltext data and EX to FrameNet exemplar data.
\begin{table}[h!tpb]
 \centering
 \begin{tabular}{ccccrrrrr}
  \toprule
  \multicolumn{9}{c}{\textsc{No filtering setup}}\\
  \midrule
  &    \multicolumn{2}{c}{train}       & filters          &            &            &     &   & \\
  XP       & FN & pFN         & POS         & \#sent           & \#anno                     & P    & R    & F$_1$         \\
  \midrule
  baseline & FT & \textendash & \textendash & \numprint{2804}  & \numprint{15389}           & 59.4 & 52.6 & 55.8          \\
  \midrule
  \#16     & FT & FT          & N           & \numprint{10386} & \numprint{34734}           & 60.3 & 51.5 & \textbf{55.6} \\
  \#18     & FT & FT          & V           & \numprint{36234} & \textbf{\numprint{126436}} & 60.4 & 50.9 & \textbf{55.3} \\
  \bottomrule
 \end{tabular}
 \caption{Argument identification results with a model trained on FrameNet
  fulltext (FT)
  augmented with \pfn data generated from FT filtered by part-of-speech (POS).
  Results are given
  for nouns (N) and verbs (V) without additional filtering on \pfn candidates}
 \label{nofilteringresults}
\end{table}

Results confirm the second scenario formulated in Section~\ref{robustness}:
the statistical classifier used for argument identification seems to
be sensitive to predicate-argument
lexical configurations,
to the point that bias introduced by unlikely
lexical configurations does not compensate
for the lexical diversity brought at the argument level. Note, in addition,
that the size of the training set increases significantly with
the paraphrastic data augmentation approach,
due to the number of paraphrastic candidates
extracted by the valence pattern matching algorithm.

\section{Introducing multi-tokens filtering}
The \rofames\ system fails to process cases where annotated lexical units
contain non-continuous tokens, such as phrasal verbs separated by a noun (e.g.
\emph{He \target{gave} the answer \target{away}}).
We therefore tried and filtered out all multi-token targets
and candidates to see whether or not it could positively
impact the performances of the classifier.
Results, displayed in Table~\ref{mutlitokensfilteringresults}, show a marginal
difference compared to the no-filtering approach. We nonetheless
applied mutli-tokens (or \emph{multiword expressions}) filtering in all subsequent experiments in order to
limit the pool of candidates and generate higher quality artificial data.
\begin{table}[h!tpb]
 \centering
 \begin{tabular}{cccccrrrrr}
  \toprule
  \multicolumn{10}{c}{\textsc{Multi-tokens filtering}}\\
  \midrule
  &    \multicolumn{2}{c}{train}       & \multicolumn{2}{c}{filters}          &            &            &     &   & \\
  XP       & FN & pFN         & POS         & MWE         & \#sent           & \#anno            & P    & R    & F$_1$ \\
  \midrule
  baseline & FT & \textendash & \textendash & \textendash & \numprint{2804}  & \numprint{15389}  & 59.4 & 52.6 & 55.8  \\
  \midrule
  \#18     & FT & FT          & V           & false       & \numprint{36234} & \numprint{126436} & 60.4 & 50.9 & 55.3  \\
  \#19     & FT & FT          & V           & true        & \numprint{34119} & \numprint{119418} & 60.9 & 50.5 & 55.2  \\
  \bottomrule
 \end{tabular}
 \caption{Impact of multi-tokens filtering (MWE = true) on
  argument identification with a model trained on FrameNet fulltext (FT)
  augmented with \pfn data generated from verbs}
 \label{mutlitokensfilteringresults}
\end{table}

\section{Introducing semantic filtering}
\label{semanticfileringall}
\subsection{Semantic filtering on nouns}
Following the observations of Section~\ref{nofilteringnv}, we first
experimented on filtering noun candidates with a
\texttt{word2vec} semantic filter.
Overall results, displayed in Table~\ref{wvnounsemfilters}, show
a marginal, though not statistically significant ($p = 0.3$), improvement
in one scenario (XP\#21), and a systematic decrease in recall otherwise,
suggesting that the \pfn system mostly
introduced noise in the training set.
\begin{table}[h!tpb]
 \centering
 \begin{tabular}{ccccccrrrrr}
  \toprule
  \multicolumn{11}{c}{\textsc{Semantic filtering on nouns}}\\
  \midrule
  &    \multicolumn{2}{c}{train}       & \multicolumn{3}{c}{filters}          &            &            &     &   & \\
  XP       & FN & pFN         & POS         & MWE         & SEM           & \#sent           & \#anno           & P    & R    & F$_1$         \\
  \midrule
  baseline & FT & \textendash & \textendash & \textendash & \textendash   & \numprint{2804}  & \numprint{15389} & 59.4 & 52.6 & 55.8          \\
  \midrule
  \#16     & FT & FT          & N           & \textendash & \textendash   & \numprint{10386} & \numprint{34734} & 60.3 & 51.5 & 55.6          \\
  \#20     & FT & FT          & N           & \textendash & random-3      & \numprint{6615}  & \numprint{24209} & 59.7 & 52.0 & 55.6          \\
  \#21     & FT & FT          & N           & \textendash & top-3         & \numprint{6537}  & \numprint{24050} & 60.4 & 52.0 & \textbf{55.9} \\
  \#22     & FT & FT          & N           & \textendash & threshold-0.5 & \numprint{3021}  & \numprint{15666} & 59.2 & 52.5 & 55.7          \\
  \#23     & FT & FT          & N           & \textendash & threshold-0.7 & \numprint{2844}  & \numprint{15434} & 60.0 & 52.2 & 55.8          \\
  \bottomrule
 \end{tabular}
 \caption{Impact of various \texttt{word2vec} semantic filtering of
  \pfn candidates on
  argument identification with a model trained on FrameNet fulltext (FT)
  augmented with \pfn data generated from nouns (N)}
 \label{wvnounsemfilters}
\end{table}

\subsection{Semantic filtering on adjectives}
Similar observations apply for adjectives, which exhibit not significant
improvements overall, regardless of the filtering configuration used
(see Table~\ref{wvsemfiltersadj}).
\begin{table}[h!tpb]
 \centering
 \begin{tabular}{ccccccrrrrr}
  \toprule
  \multicolumn{11}{c}{\textsc{Semantic filtering on adjectives}}\\
  \midrule
  &    \multicolumn{2}{c}{train}       & \multicolumn{3}{c}{filters}          &            &            &     &   & \\
  XP       & FN & pFN         & POS         & MWE         & SEM           & \#sent          & \#anno           & P    & R    & F$_1$ \\
  \midrule
  baseline & FT & \textendash & \textendash & \textendash & \textendash   & \numprint{2804} & \numprint{15389} & 59.4 & 52.6 & 55.8  \\
  \midrule
  \#33     & FT & FT          & A           & \textendash & random-3      & \numprint{6000} & \numprint{21811} & 59.7 & 52.3 & 55.8  \\
  \#34     & FT & FT          & A           & \textendash & top-3         & \numprint{5984} & \numprint{21799} & 59.5 & 52.6 & 55.8  \\
  \#35     & FT & FT          & A           & \textendash & threshold-0.5 & \numprint{3490} & \numprint{16221} & 59.4 & 52.3 & 55.6  \\
  \bottomrule
 \end{tabular}
 \caption{Impact of various \texttt{word2vec} semantic filtering of
  \pfn candidates on
  argument identification with a model trained on FrameNet fulltext (FT)
  augmented with \pfn data generated from adjectives (A)}
 \label{wvsemfiltersadj}
\end{table}

\subsection{Semantic filtering on verbs}
Experiments with semantic filtering on verbs,
reported in Table~\ref{wvsemfiltersverbs}, show a relative increase in recall
over the baseline, especially compared to no-filtering approaches.
Overall, results tend to show that filtering candidates by keeping those
\emph{closest semantically} to the original target is beneficial, especially
when keeping two to three candidates per target only. No significant different
is observed between \emph{top} and \emph{threshold} filtering.
The most interesting experiment is probably XP\#30 which shows a .4 $F_1$ gain
over the baseline with a less that doubled dataset.
Note, however, that none of the experiments proved statistically significant
($p > 0.1$) and that the best improvements
in precision seen in XP\#31 are
incidental to the decrease in recall.

\begin{table}[h!tpb]
 \centering
 \begin{tabular}{ccccccrrrrr}
  \toprule
  \multicolumn{11}{c}{\textsc{Semantic filtering on verbs}}\\
  \midrule
  &    \multicolumn{2}{c}{train}       & \multicolumn{3}{c}{filters}          &            &            &     &   & \\
  XP            & FN          & pFN         & POS         & MWE           & SEM            & \#sent                   & \#anno                    & P             & R             & F$_1$         \\
  \midrule
  bas.          & FT          & \textendash & \textendash & \textendash   & \textendash    & \numprint{2804}          & \numprint{15389}          & 59.4          & 52.6          & 55.8          \\
  \midrule
  \#19          & FT          & FT          & V           & true          & \textendash    & \numprint{34119}         & \numprint{119418}         & 60.9          & 50.5          & 55.2          \\
  \textbf{\#30} & \textbf{FT} & \textbf{FT} & \textbf{V}  & \textbf{true} & \textbf{top-2} & \textbf{\numprint{8213}} & \textbf{\numprint{28772}} & \textbf{60.2} & \textbf{52.8} & \textbf{56.2} \\
  \#25          & FT          & FT          & V           & true          & top-3          & \numprint{11099}         & \numprint{37491}          & 60.1          & 52.7          & 56.1          \\
  \#31          & FT          & FT          & V           & true          & top-4          & \numprint{13548}         & \numprint{45227}          & 61.1          & 52.0          & 56.2          \\
  \#26          & FT          & FT          & V           & true          & threshold-0.5  & \numprint{4720}          & \numprint{18826}          & 59.2          & 52.7          & 55.8          \\
  \#27          & FT          & FT          & V           & true          & threshold-0.7  & \numprint{2969}          & \numprint{15568}          & 59.7          & 52.5          & 55.9          \\
  \bottomrule
 \end{tabular}
 \caption{Impact of various \texttt{word2vec} semantic filtering of
  \pfn candidates on
  argument identification with a model trained on FrameNet fulltext (FT)
  augmented with \pfn data generated from verbs (V). None of the improvements
 reported proved statistically significant ($p > 0.1$)}
 \label{wvsemfiltersverbs}
\end{table}

\subsection{Semantic filtering on nouns and verbs}
We also attempted to augment data in two dimensions at once, namely
nouns and verbs, in order to increase lexical diversity and potentially
bring new information to the statistical classifier.
Experiments were all performed with a semantic filter on top, to reduce
the pool of candidates, for both quality and feasability reasons.
Results, displayed in Table~\ref{wvsemfiltersnvboth} show a significant drop
in recall (XP\#32) in a top-n setting, with an explosion of the size
of the augmented dataset generated, suggesting very noisy and low quality
data. Results in a threshold setting (XP\#36) did not bring statistically
significant improvements, and generated a limited number of
paraphrastic sentences.

\begin{table}[h!tpb]
 \centering
 \begin{tabular}{ccccccrrrrr}
  \toprule
  \multicolumn{11}{c}{\textsc{Semantic filtering on nouns and verbs}}\\
  \midrule
  &    \multicolumn{2}{c}{train}       & \multicolumn{3}{c}{filters}          &            &            &     &   & \\
  XP   & FN & pFN         & POS         & MWE         & SEM           & \#sent           & \#anno            & P    & R    & F$_1$ \\
  \midrule
  bas. & FT & \textendash & \textendash & \textendash & \textendash   & \numprint{2804}  & \numprint{15389}  & 59.4 & 52.6 & 55.8  \\
  \midrule
  \#32 & FT & FT          & N+V         & true        & top-3         & \numprint{34367} & \numprint{153293} & 60.4 & 49.4 & 54.3  \\
  \#36 & FT & FT          & N+V         & true        & threshold-0.5 & \numprint{5118}  & \numprint{19807}  & 60.0 & 52.5 & 56.0  \\
  \bottomrule
 \end{tabular}
 \caption{Impact of various \texttt{word2vec} semantic filtering of
  \pfn candidates on
  argument identification with a model trained on FrameNet fulltext (FT)
  augmented with \pfn data generated from nouns and verbs (N +V).
 In both settings (top and threshold), $p > 0.1$}
 \label{wvsemfiltersnvboth}
\end{table}

\subsection{Comparing random vs. top filtering}
\label{randomversusall}
We then tried to quantify the real benefits brought by a semantic filter
based on proximity measures between two distributional representations, compared
to a control filter selecting candidates randomly.
The top-n filter, choosing the $n$ paraphrastic candidates closest to the
original target from a given list output by \pfn, did not actually perform
better than the random-n filter
choosing n candidates randomly (see Table~\ref{wvtoprandom}).

\begin{table}[h!tpb]
 \centering
 \begin{tabular}{lcccccrrrrr}
  \toprule
  \multicolumn{11}{c}{\textsc{Random vs. top semantic filtering}}\\
  \midrule
  &    \multicolumn{2}{c}{train}       & \multicolumn{3}{c}{filters}          &            &            &     &   & \\
  XP   & FN & pFN & POS & MWE  & SEM      & \#sent           & \#anno           & P    & R    & F$_1$ \\
  \midrule
  \#28 & FT & FT  & V   & true & random-2 & \numprint{8278}  & \numprint{28885} & 60.3 & 52.6 & 56.1  \\
  \#30 & FT & FT  & V   & true & top-2    & \numprint{8213}  & \numprint{28772} & 60.2 & 52.8 & 56.2  \\
  \#24 & FT & FT  & V   & true & random-3 & \numprint{11188} & \numprint{37632} & 60.4 & 52.5 & 56.2  \\
  \#25 & FT & FT  & V   & true & top-3    & \numprint{11099} & \numprint{37491} & 60.1 & 52.7 & 56.1  \\
  \#29 & FT & FT  & V   & true & random-4 & \numprint{13563} & \numprint{45143} & 60.9 & 52.2 & 56.2  \\
  \#31 & FT & FT  & V   & true & top-4    & \numprint{13548} & \numprint{45227} & 61.1 & 52.0 & 56.2  \\
  \bottomrule
 \end{tabular}
 \caption{Impact of top-n \texttt{word2vec} semantic filters \emph{versus}
  n-random filters for \pfn data generated from verbs. Improvements reported
  are not statistically significant ($p > 0.1$ except for XP\#29 with $p = 0.08$)}
 \label{wvtoprandom}
\end{table}

\subsection{Comparing predictive and count-based distributional models}
The previous experiment comparing \emph{top} and \emph{random} semantic filters
raised the question of the quality of the underlying \texttt{word2vec}
distributional model and toolkit used to measure semantic proximity between
paraphrastic candidates and targets. Past research \citep{baronietal2014}
have shown
that the choice of distributional model can significantly impact
performances on tasks such as \emph{semantic relatedness} and
\emph{synonym detection}.
We therefore experimented with two additional
distributional models and toolkits, namely \texttt{dissect-pred} and
\texttt{dissect-rcount}, on top of \texttt{word2vec}.\footnote{see Section~\ref{filteringoptions} for details}
However, our experimental results, displayed in Table~\ref{dissectwv},
showed only marginal differences across models and toolkits.

\begin{table}[h!tpb]
 \centering
 \begin{tabular}{lccccrrrrr}
  \toprule
  \multicolumn{10}{c}{\textsc{Predictive vs. count-based distributional models}}\\
  \midrule
  XP & POS & MWE & \multicolumn{2}{c}{SEM} & \#sent            & \#anno & P & R & F$_1$ \\
  \midrule
  \#25 & V & true & word2vec       & top-3         & \numprint{11099} & \numprint{37491} & 60.1 & 52.7 & 56.1 \\
  \#37 & V & true & dissect-pred   & top-3         & \numprint{11006} & \numprint{37259} & 59.8 & 52.9 & 56.1 \\
  \#38 & V & true & dissect-rcount & top-3         & \numprint{11026} & \numprint{37281} & 60.4 & 52.3 & 56.0 \\
  \#26 & V & true & word2vec       & threshold-0.5 & \numprint{4720}  & \numprint{18826} & 59.2 & 52.7 & 55.8 \\
  \#39 & V & true & dissect-pred   & threshold-0.5 & \numprint{3758}  & \numprint{16855} & 59.5 & 52.7 & 55.9 \\
  \#40 & V & true & dissect-rcount & threshold-0.5 & \numprint{6350}  & \numprint{23656} & 60.1 & 52.4 & 56.0 \\
  \bottomrule
 \end{tabular}
 \caption{Comparing predictive and count-based distributional models
  for paraphrastic candidates filtering. Impact is measured on argument identification.
  \texttt{word2vec} is the traditional predictive model trained on Google News
  and used with the \texttt{word2vec} toolkit. \texttt{dissect-pred} is
  a \texttt{word2vec} predictive model extracted from a different dataset than
  Google News, and used with the \texttt{dissect} toolkit. \texttt{dissect-rcount}
  is a count-based model extracted from the same dataset than \texttt{dissect-pred}
  and used with the \texttt{dissect} toolkit}
 \label{dissectwv}
\end{table}

\subsection{Measuring the impact of exemplar data}
\label{measuringimpactexemplar}
Finally, we experimented with a \pfn model performing valence pattern
matching on the FrameNet fulltext plus exemplar
dataset. Our purpose was to analyze whether such a model could actually beneficially
incorporate information available in the exemplar dataset, or, failing that,
be able to create an augmentated artificial dataset of higher quality given
that paraphrastic candidates would be chosen from a larger pool.
Results, displayed in Table~\ref{semexemplars}, showed no significant
improvements over the \pfn model based on fulltext alone. Note that improvements
reported in XP\#30 and XP\#41 did not prove statistically significant ($p > 0.1$),
while improvements on XP\#12 did ($p < 0.01$). This shows that training on
gold fulltext and exemplar data leads to much more robust improvements than training
on our artificial dataset.

\begin{table}[h!tpb]
 \centering
 \begin{tabular}{ccccccrrrrr}
  \toprule
  \multicolumn{11}{c}{\textsc{Impact of exemplar data}}\\
  \midrule
  &    \multicolumn{2}{c}{train}       & \multicolumn{3}{c}{filters}          &            &            &     &   & \\
  XP    & FN    & pFN         & POS         & MWE         & SEM         & \#sent            & \#anno            & P    & R    & F$_1$         \\
  \midrule
  base. & FT    & \textendash & \textendash & \textendash & \textendash & \numprint{2804}   & \numprint{15389}  & 59.4 & 52.6 & 55.8          \\
  \midrule
  \#12  & FT+EX & \textendash & \textendash & \textendash & \textendash & \numprint{168237} & \numprint{184050} & 58.4 & 55.7 & \textbf{57.0} \\
  \#30  & FT    & FT          & V           & true        & top-2       & \numprint{8213}   & \numprint{28772}  & 60.2 & 52.8 & 56.2          \\
  \#41  & FT    & FT+EX       & V           & true        & top-2       & \numprint{14588}  & \numprint{53536}  & 60.3 & 52.9 & 56.3          \\
  \bottomrule
 \end{tabular}
 \caption{Impact of using exemplar data for training \rofames (XP\#12),
  or for generating artificial data via \pfn (XP\#41). A top-2 \texttt{word2vec}
  semantic filter is applied to limit the size of the paraphrastic dataset generated
  by \pfn.
  Improvements in XP\#30 and XP\#41 are not statistically significant ($p > 0.1$),
  while improvements in XP\#12 are ($p < 0.01$)}
 \label{semexemplars}
\end{table}

\chapter{Error analysis}
\label{analysis}

In this chapter we discuss both \emph{quantitative} and
\emph{qualitative} performances
of the statistical classifier used for argument identification.
We focus on trying to identify (linguistic) phenomena and/or specific (linguistic)
patterns which may account for the performances reported in Chapter~\ref{results}.
Our analysis is articulated around two axis: in Section~\ref{errorbaseline},
we present an
exhaustive error analysis of the baseline system, proposing several angles
to tackle the key question of what constitutes \emph{easy} and \emph{hard}
frame elements, from the perspective of statistical learning.
In Section~\ref{comparativeanalysis} we discuss the respective contributions
of incorporating
exemplar and paraphrastic data into the training set, both in terms of
coverage and how it affects data distribution.
In the following sections we use \emph{exemplar approach} to refer to the
approach consisting in training on fulltext and exemplar data, and
\emph{paraphrastic approach} to refer to the approach corresponding to
experiment 30, augmented on verbs with a \texttt{word2vec} top-2 filter.

\section{Baseline error analysis}
\label{errorbaseline}
The purpose of this section is to discuss both qualitative and quantitative
performances of the statistical classifier at a more fined-grained level
than the micro-averaged results presented in Chapter~\ref{results}.
We report on the robustness of the parser to frequency effects
(Section~\ref{errorfrequencyeffects}), analyze the impact of
\emph{frame elements bearing targets} on argument identification
(Section~\ref{framelementsbearingtargets}),
and discuss both lexical (Section~\ref{lexicalchallenges})
and syntactic challenges
(Section~\ref{syntacticchallenges}) posed to statistical learning.

\subsection{Frequency effects on Frame Elements}
\label{errorfrequencyeffects}
The first observation that can be made when looking at precision, recall and
$F_1$ measure \emph{per} frame element, is that the \rofames parser seems
relatively robust to frequency effects.
Indeed, Table~\ref{tenmostfrequentfes} shows that, among the ten most
frequent frame
elements in the development set, $F_1$ measures do not seem to correlate
well with frequency and/or distribution in either training or development sets.
The lack of pattern is confirmed when classifying frame elements by
$F_1$ measure, such as in Table~\ref{top10bestframelements} or Table~\ref{top10worstframelements}:
we find certain frequent frame elements to have relatively low $F_1$ scores, such as
\fe{Descriptor}, \fe{Place} or \fe{Figure}, and conversely, certain less
frequent frame elements to have higher $F_1$ scores, such a \fe{Count}, \fe{Part},
or \fe{Material}.

\begin{table}[!htb]
 \centering
 \begin{tabular}{lrrrrr}
  \toprule
  \multicolumn{6}{c}{\textsc{Top 10 Most frequent FEs}}\\
  \midrule
             & train           & dev & P    & R    & F$_1$ \\
  \midrule
  Agent      & \numprint{1391} & 347 & 68.5 & 64.6 & 66.5  \\
  Entity     & \numprint{1204} & 318 & 59.4 & 55.7 & 57.5  \\
  Locale     & 957             & 317 & 87.3 & 89.0 & 88.1  \\
  Weapon     & 608             & 308 & 67.8 & 66.2 & 67.0  \\
  Time       & 791             & 275 & 54.3 & 34.5 & 42.2  \\
  Theme      & 796             & 232 & 56.2 & 56.5 & 56.3  \\
  Message    & 541             & 215 & 60.3 & 53.0 & 56.4  \\
  Event      & 670             & 211 & 44.9 & 31.3 & 36.9  \\
  Speaker    & 598             & 201 & 72.0 & 71.6 & 71.8  \\
  Descriptor & 527             & 172 & 35.3 & 14.0 & 20.0  \\
  \bottomrule
 \end{tabular}
 \caption{Top 10 most frequent frame elements, with occurrence counts in training
 and development sets}
 \label{tenmostfrequentfes}
\end{table}

\begin{table}[!htb]
 \begin{minipage}[t]{.48\linewidth}
  \centering
  \resizebox{\linewidth}{!}{
   \begin{tabular}{l@{\hskip 2cm}rrrrr}
    \toprule
    \multicolumn{6}{c}{\textsc{Top 10 Best FEs}}\\
    \midrule
              & train & dev & P    & R    & F$_1$ \\
    \midrule
    Number    & 207   & 59  & 96.6 & 96.6 & 96.6  \\
    Substance & 239   & 125 & 91.7 & 88.8 & 90.2  \\
    Locale    & 957   & 317 & 87.3 & 89.0 & 88.1  \\
    Person    & 304   & 70  & 84.9 & 88.6 & 86.7  \\
    Count     & 164   & 82  & 73.5 & 74.4 & 73.9  \\
    Part      & 116   & 72  & 73.2 & 72.2 & 72.7  \\
    Unit      & 382   & 146 & 64.2 & 82.2 & 72.1  \\
    Speaker   & 598   & 201 & 72.0 & 71.6 & 71.8  \\
    Material  & 106   & 61  & 84.4 & 62.3 & 71.7  \\
    Cognizer  & 537   & 129 & 72.5 & 67.4 & 69.9  \\
    \bottomrule
   \end{tabular}
  }
  \caption{Top 10 best frame elements among the top 50 most frequent
  frame elements in dev set, ranked by $F_1$ score}
  \label{top10bestframelements}
 \end{minipage}%
 \hspace{.04\linewidth}
 \begin{minipage}[t]{.48\linewidth}
  \centering
  \resizebox{\linewidth}{!}{
   \begin{tabular}{lrrrrr}
    \toprule
    \multicolumn{6}{c}{\textsc{Top 10 Worst FEs}}\\
    \midrule
                        & train & dev & P    & R    & F$_1$ \\
    \midrule
    Name                & 246   & 54  & 27.3 & 11.1 & 15.8  \\
    Manner              & 251   & 89  & 29.4 & 11.2 & 16.3  \\
    Means               & 144   & 54  & 42.9 & 11.1 & 17.6  \\
    Descriptor          & 527   & 172 & 35.3 & 14.0 & 20.0  \\
    Place               & 354   & 99  & 47.6 & 20.2 & 28.4  \\
    Figure              & 464   & 83  & 31.0 & 32.5 & 31.8  \\
    Cause               & 231   & 53  & 43.8 & 26.4 & 32.9  \\
    State\_of\_affairs  & 145   & 153 & 42.4 & 27.5 & 33.3  \\
    Use                 & 310   & 112 & 40.7 & 29.5 & 34.2  \\
    Hypothetical\_event & 215   & 57  & 50.0 & 28.1 & 36.0  \\
    \bottomrule
   \end{tabular}
  }
  \caption{Top 10 worst frame elements among the top 50 most frequent frame elements
  in dev set, ranked by $F_1$ score}
  \label{top10worstframelements}
 \end{minipage}
\end{table}

\subsection{Frame elements bearing targets}
\label{framelementsbearingtargets}
Looking closer at Table~\ref{top10bestframelements}, we see that a limited
number of frame elements, such as \fe{Number}, \fe{Substance}, \fe{Locale} and
\fe{Person}, have significantly higher $F_1$ scores,
systematically above 86\% when the overall $F_1$ on the top 50 most
frequent frame elements is at 56.2\% and the
average $F_1$ measure at 52.1\%. Those frame elements have in common to be
realized by their respective targets, in what is called
\emph{frame elements bearing targets} configurations, as in:
\begin{example}
 \label{examplefebar}
 These are surrounded by areas of limestone formations, scrub and grassland,
 coral cliffs, and fine \feanno{\target{sand}}{Substance} beaches
\end{example}
In example~\ref{examplefebar}, \emph{sand} is both the target evoking the
\semframe{Shapes} frame and the frame element \fe{Substance}.
Such configurations appear particularly easy for the parser to acquire
as it requires learning a correlation between a frame element, a frame
and a frame-evoking target without special considerations for the frame element
spans or syntactic realization.
To better account for such cases we introduce the \febar ratio, which,
for a given frame element, is defined as:
\begin{equation}
 \febar(f_e) = \frac{N_{febc}}{N_c}
\end{equation}
where $N_{febc}$ is the number of frame element bearing configurations for the
frame element $f_e$ and ${N_c}$ is the total number of configurations for the
frame element $f_e$.

As shown in Table~\ref{top10febyf1andfebar} and Table~\ref{top10febyfebar},
we find the \febar ratio to correlate nicely with the $F_1$ score
of a given frame element: the higher the \febar ratio, the more a given
frame element will be realized in a frame element bearing configuration, and
the easier it will be for the parser to predict it.
There are three notable exceptions which constitute interesting cases:
the \fe{Count} and \fe{Speaker} frame elements which have high
$F_1$ scores but low \febar ratios, meaning that they are easy to predict
despite being realized \emph{outside} of their respective targets;
and the \fe{Weapon}
frame element which has a high \febar ratio but a $F_1$ significantly lower
than frame elements with comparable \febar ratios, by almost 20 points of
$F_1$ score. Those examples actually illustrate interesting lexico-syntactic
patterns that we will discuss in the following sections.

\begin{table}[!htb]
 \begin{minipage}[t]{.48\linewidth}
  \centering
  \resizebox{\linewidth}{!}{
   \begin{tabular}{lrrrr}
    \toprule
    \multicolumn{5}{c}{\textsc{Top 10 FEs by $F_1$}}\\
    \midrule
              & febar        & P    & R    & F$_1$ \\
    \midrule
    Number    & 100.0        & 96.6 & 96.6 & 96.6  \\
    Substance & 95.2         & 91.7 & 88.8 & 90.2  \\
    Locale    & 96.5         & 87.3 & 89.0 & 88.1  \\
    Person    & 92.9         & 84.9 & 88.6 & 86.7  \\
    Count     & \textbf{1.2} & 73.5 & 74.4 & 73.9  \\
    Part      & 34.7         & 73.2 & 72.2 & 72.7  \\
    Unit      & 81.5         & 64.2 & 82.2 & 72.1  \\
    Speaker   & \textbf{1.5} & 72.0 & 71.6 & 71.8  \\
    Material  & 54.1         & 84.4 & 62.3 & 71.7  \\
    Cognizer  & 9.3          & 72.5 & 67.4 & 69.9  \\
    \bottomrule
   \end{tabular}
  }
  \caption{Top 10 best frame elements among the top 50 most frequent
   frame elements in dev set, with \febar ratio and ranked by $F_1$ score}
  \label{top10febyf1andfebar}
 \end{minipage}%
 \hspace{.04\linewidth}
 \begin{minipage}[t]{.48\linewidth}
  \centering
  \resizebox{\linewidth}{!}{
   \begin{tabular}{lrrrr}
    \toprule
    \multicolumn{5}{c}{\textsc{Top 10 FEs by febar}}\\
    \midrule
              & febar & P    & R    & F$_1$         \\
    \midrule
    Number    & 100.0 & 96.6 & 96.6 & 96.6          \\
    Locale    & 96.5  & 87.3 & 89.0 & 88.1          \\
    Substance & 95.2  & 91.7 & 88.8 & 90.2          \\
    Weapon    & 94.5  & 67.8 & 66.2 & \textbf{67.0} \\
    Person    & 92.9  & 84.9 & 88.6 & 86.7          \\
    Unit      & 81.5  & 64.2 & 82.2 & 72.1          \\
    Origin    & 67.6  & 56.4 & 62.0 & 59.1          \\
    Leader    & 63.0  & 64.3 & 66.7 & 65.5          \\
    Material  & 54.1  & 84.4 & 62.3 & 71.7          \\
    Text      & 45.1  & 71.1 & 52.9 & 60.7          \\
    \bottomrule
   \end{tabular}
  }
  \caption{Top 10 frame elements among the top 50 most frequent
   frame elements in dev set, ranked by \febar ratio}
  \label{top10febyfebar}
 \end{minipage}
\end{table}

\subsection{Lexical challenges}
\label{lexicalchallenges}

\subsubsection{Lexical diversity}
\label{lexicaldiversityerror}
As mentioned in Chapter~\ref{relatedwork}, lexical coverage and out-of-vocabulary
effects play a determining role in frame semantic parsing in general.
On argument identification, the impact of lexical coverage is probably best
exemplified by the \fe{Name} frame element, which covers mostly proper nouns,
as in Sentence~\ref{examplenamefe}, and as such is typically affected
by out-of-vocabulary effects.
\begin{example}
 \label{examplenamefe}
 Your Georgia O'Keeffe membership, please note, includes free admission to
 all art-related activities of the IMA's \feanno{Young Friends of Art}{Name}
 \target{group}
\end{example}
As shown in Table~\ref{top10worstframelements},
the \fe{Name} frame element has the lowest performances among all frame elements
predicted in the development set, with a $F_1$ score of 15.8\%, significantly below
both micro-averaged (56.2\%) and averaged (52.1\%) $F_1$ scores on all frame elements.
The task of identifying
\fe{Name} labels is actually a harder task than mere Named
Entity Recognition, considering that \fe{Name} frame elements can
be realized in various syntactic patterns, beyond noun phrases, as in:
\begin{example}
 The \target{island} \feanno{of Jamaica}{Name} will be near the top of the list for
 anyone planning an idyllic holiday getaway
\end{example}
The difficulty posed by the \fe{Name} frame element is encompassed in the
larger phenomenon of \emph{lexical diversity}: if a frame element is realized
in a wide range of lexical patterns, the frame element may prove harder to predict,
especially if it is realized in a lexical pattern that has not been previously
seen in the training set.
This is exactly what is happening for the \fe{Name} frame element: as
there is a multitude of possible lexical patterns
that can be labeled with \fe{Name}, they will prove almost impossible
to predict by the parser
if previously unseen in training, as there will be
no way for the parser to compensate for its lack of lexical knowledge.

\subsubsection{Compensating lexical diversity}
\label{compensatinglexical}
However, lexical diversity itself is not enough to characterize the
difficulty to predict a given frame element.
Indeed, high lexical diversity effects can be compensated by systematic
syntactic patterns. In other words, a frame element which is realized
lexically in a wide range of patterns may prove easy to predict
as long as it is realized in clear and systematic syntactic patterns.
This is typically the case of frame elements such as \fe{Agent} or \fe{Speaker},
which are overwhelmingly realized as subjects in nouns phrases (\vpattern{NP.Ext})
\textendash\
75\% of the time for \fe{Agent} and 85\% of the time for \fe{Speaker}
respectively.
Moreover, the \vpattern{NP.Ext} syntactic pattern has the benefit of
translating well into a single arc (\emph{subj}) between
the predicate and the
argument, in the dependency tree produced by the syntactic parser.
Therefore, the statistical classifier used for argument identification can
adequately learn a correlation between certain predicates and the arguments
connected to them by a \emph{subj} arc in order to properly predict
frame element labels such as \fe{Agent} or \fe{Speaker}. Such phenomena
explain why $F_1$ scores in frame elements such as \fe{Speaker} (71.8\%) can
be significantly higher than the micro-averaged and averaged $F_1$ measures
on all frame elements, despite their low respective \febar ratio.

Such compensating effects become all the more obvious when syntactic representations
between output dependencies and FrameNet valence units do not align.
In such cases, the statistical classifier clearly fails to properly
classify corresponding spans with the adequate frame element label, as in:
\begin{example}
 The Defense Department has awarded \feanno{BioPort Corporation}{Agent} a
 contract to
 manufacture, test, bottle and \target{store} \feanno{anthrax vaccine}{Theme},
 the company has announced.
\end{example}
In this particular case \emph{BioPort Corporation} is labeled by
the syntactic parser
as being in an \emph{obj} \textendash\ and not \emph{subj} \textendash\
dependency to the \emph{awarded}
\textendash\ and not
\emph{store} \textendash\
predicate,
while it is marked as \vpattern{NP.Ext} in FrameNet for the \target{store} target.
Consequently, the \rofames
parser fails to properly label it as the \fe{Agent}, a phenomenon that is
found to be systematic and can be grouped into the more general category
of \emph{syntactic misalignements}.
The question of syntactic challenges will be covered in greater
length in Section~\ref{syntacticchallenges}.

\subsubsection{Semantic considerations}
\label{semanticconsiderations}
When analyzing more carefully the case of the \fe{Weapon} frame element,
we found interesting cases where the appropriate annotation of the
frame element label required specific semantic considerations involving
deep lexical representations.
Indeed, in FrameNet, terms such as \emph{nuclear} will be solely labeled as
\fe{Weapon} in constructions such as \emph{nuclear war}, while it is usually
labeled as \fe{Weapon} together with the noun it qualifies in constructions such
as \emph{nuclear weapon} or \emph{nuclear missile}.
While the level of inter-annotator agreement on such cases is questionable,
we found the systematicity of the annotation rather consistent and motivated
by deep semantic considerations: a \emph{nuclear missile} is indeed a weapon,
but the \emph{nuclear} adjective here denotes a missile using nuclear energy.
While a \emph{nuclear war} implies a war involving \emph{nuclear weapons},
hence \emph{nuclear} here can be understood in its metonymic sense.
Given current statistical models for frame semantic parsing, it seems
hardly possible that those models would prove capable of capturing
such complex semantic phenomena, given the
depth of both lexical and semantic representations that such
phenomena would require to be extracted, all the more as they involve
information not readily available at the surface syntactic level.

\subsection{Syntactic challenges}
\label{syntacticchallenges}
In Section~\ref{compensatinglexical} were briefly mentioned
problems related to misalignements of syntactic representations between FrameNet
annotation and the output of the dependency parser.
In addition to those misalignement problems,
we found the vast majority of syntactic challenges to come from the
argument span identification algorithm, detailed in
Section~\ref{argidentificationmodel}. Table~\ref{scoresbysyntacticrealizations}
provides a detailed account of performances per syntactic realization of
the frame elements (PT.GF), which shows flagrant disparities of performances
across phrase types and grammatical functions.
\begin{table}[!htb]
 \centering
 \begin{tabular}{l@{\hskip 2cm}rrrr}
  \toprule
  \multicolumn{5}{c}{\textsc{Argument id. scores by PT.GF}}\\
  \midrule
  PT.GF         & dev             & P     & R    & F$_1$ \\
  \midrule
  Poss.Gen      & 41              & 100.0 & 85.4 & 92.1  \\
  Num.Quant     & 58              & 78.6  & 75.9 & 77.2  \\
  NP.Ext        & \numprint{1031} & 78.1  & 60.0 & 67.9  \\
  NP.Obj        & 542             & 69.6  & 60.0 & 64.4  \\
  Sfin.Dep      & 173             & 77.0  & 54.3 & 63.7  \\
  A.Dep         & 64              & 93.5  & 45.3 & 61.1  \\
  VPto.Dep      & 179             & 70.5  & 52.0 & 59.8  \\
  N.Head        & 252             & 63.1  & 54.4 & 58.4  \\
  PP.Dep        & 516             & 80.4  & 42.1 & 55.2  \\
  AVP.Dep       & 140             & 97.6  & 29.3 & 45.1  \\
  NP.Dep        & 19              & 66.7  & 31.6 & 42.9  \\
  N.Dep         & 243             & 50.4  & 24.3 & 32.8  \\
  AJP.Dep       & 170             & 91.2  & 18.2 & 30.4  \\
  VPbrst.Dep    & 81              & 35.1  & 16.0 & 22.0  \\
  Sfin.Head     & 65              & 21.2  & 10.8 & 14.3  \\
  NP.Appositive & 16              & 6.2   & 6.2  & 6.2   \\
  \bottomrule
 \end{tabular}
 \caption{Argument identification scores by syntactic realizations
 of the frame elements}
 \label{scoresbysyntacticrealizations}
\end{table}
The lowest scores reported, for PT.GF such as \vpattern{NP.Appositive},
\vpattern{Sfin.Head} or \vpattern{VPbrst.Dep} are almost exclusively
accounted for by approximations of the span identification algorithm and the
output representation of the dependency parser.
Indeed, the algorithm only considers as possible argument spans all continuous
spans in a sentence that contain a single word or comprise a valid subtree
of a word and all its descendants in the dependency parse produced by the
dependency parser. Therefore, the parser will systematically fail to recover
spans in the following configurations:
\begin{itemize}
 \item \textbf{appositive noun phrases} (\vpattern{NP.Appositive}) as in:
       \begin{example}
        In a statement, \target{President} \vuanno{Jacob Zuma}{Leader}{NP}{Appositive}
        said Mandela's condition was unchanged
       \end{example}
       where the possible spans output by the algorithm are either
       [President] [Jacob] [Zuma] taken
       individually or
       [President Jacob Zuma] as a whole, given that the dependency parser
       considers both
       \emph{President} and \emph{Jacob} to be children nodes of \emph{Zuma},
       which can therefore never produce [President] and [Jacob Zuma] as
       potential spans;
 \item \textbf{Head declarative finite complement clauses} (\vpattern{Sfin.Head}) as in:
       \begin{example}
        After a lifetime of trials,
        \vuanno{Donna not only earned her GED}{Figure}{Sfin}{Head} \target{at}
        Goodwill, she earned a job here
       \end{example}
       where the dependency parser considers \emph{at} to be a child node
       of the \emph{earned} predicate, which prevents separating \emph{at}
       from the rest of the clause in
       [Donna not only earned her GED at], necessary to output
       the correct span;
 \item \textbf{Dependent bare stem verb phrases} (\vpattern{VPbrst.Dep}) as in:
       \begin{example}
        Here's another story of success from
        \vuanno{what}{Hypothetical\_event}{NP}{Ext} \target{might}
        \vuanno{seem like an unlikely source}{Hypothetical\_event}{VPbrst}{Dep} \ldots
       \end{example}
       where again, due to the output of the dependency parser, the
       span identification algorithm will output [what might seem like] and
       [an unlikely source] as possible spans but never [seem like an unlikely source]
\end{itemize}

\subsection{Characterizing \emph{easy} frame elements}
In Section~\ref{lexicalchallenges} we described several properties of
frame elements that
conditioned their accurate identification by statistical classifiers. Those properties included
a low lexical diversity, or a potentially high lexical diversity combined
with consistent syntactic patterns.
Such properties illustrate why statistical classifiers
are actually less sensitive to frequency effects than to the strict and clear
correlation that may exist between a lexico-syntactic configuration and a given
frame element. This is best exemplified by cases which combine both
low lexical diversity and consistent syntactic patterns, as
does the \fe{Speaker} frame element when realized in a object
noun phrase (\vpattern{Speaker.NP.Obj}). Such a configuration
exclusively corresponds to
constructions involving \emph{according to} followed by an object noun phrase, as in:
\begin{example}
 \target{According to} \vuanno{John}{Speaker}{NP}{Obj}, Mary has already left
\end{example}
As expected, it is easily \emph{learnable} as demonstrated by the 100\% $F_1$ score
measured on its five occurences in the development set.

\subsection{Characterizing \emph{hard} frame elements}
\label{hardframeelements}
\emph{A contrario}, we can characterize \emph{hard} frame elements as
frame elements realized in very diverse lexical configurations not compensated
by systematic syntactic patterns which can be easily mapped to syntactic
representations produced by the dependency parser.
Such cases are actually best exemplified by \emph{non-core} frame elements.
Indeed, both \emph{peripheral} and \emph{extra-thematic} frame elements often
carry the particularity of being realized in potentially many different
frames. In a sense, one
could say that they are realized across various \emph{domains}, which
could explain why it turns out to be so challenging to extract systematicity
\textendash\ especially
lexical \textendash\ from their surface realizations.
Concretely, a \fe{Manner} frame element in the
\semframe{Facial\_expression} frame,
as in Sentence~\ref{mannerfacial}, will exhibit very different
lexico-syntactic patterns
than a \fe{Manner} frame element in the \semframe{Intentionally\_act} frame, as in
Sentence~\ref{mannerintentionnal}.
\begin{examples}
 \item \label{mannerfacial} They are equally direct in their dealings with visitors, too,
 so don't expect a \feanno{shy}{Manner} Jamaican \target{smile} as you walk by
 \item \label{mannerintentionnal} This kind of protection should be \target{done}
 \feanno{without drawing attention}{Manner} so that people inside the
 location would feel at ease
\end{examples}
Note that the difficulty posed by non-core frame elements explain the poor
results observed for adjectival and adverbial phrases in Table~\ref{scoresbysyntacticrealizations},
as \vpattern{AVP.Dep} and \vpattern{AJP.Dep} patterns are exclusively realized
in peripheral frame elements such as \fe{Time}, \fe{Manner} or \fe{Means}.

It could be possible to try and provide a systematic \emph{measure} of the
difficulty of a given frame element, constructed by quantifying the diversity
of its lexico-syntactic realization patterns, but we leave this to future work.

\section{Comparative analysis}
\label{comparativeanalysis}
The qualitative error analysis on both exemplar and paraphrastic approaches
yielded no conclusive results as to which patterns both approaches helped
better acquire, in light of what has been described in Section~\ref{errorbaseline}.
The error analysis of the paraphrastic approach confirmed that
the improvements observed were not statistically significant and more likely
incidental to the data distribution in the development set.
In the following sections we discuss two alternative hypothesis as to
what each approach contributed to: in Section~\ref{errorcoverage} we show that the
contribution of the exemplar approach is probably best explained and
characterized in terms of the positive impact it had on various
coverage metrics between the training and the development set.
In Section~\ref{errordistribution} we confirm the robustness of the statistical
classifier
to frequency effects and show that none of the aforementioned approaches'
contribution can be explained or characterized in terms of their
respective impact on data distribution in the training set.

\subsection{Coverage}
\label{errorcoverage}

The baseline, paraphrastic and exemplar approaches can
be first and foremost distinguished
in terms of \emph{coverage} between their respective training sets and the
development set against which argument identification is evaluated.
Indeed, the specificity of the paraphrastic approach is that it does not
bring any \emph{new} data to the training set, with respect to LU \textendash\ FE
\textendash\ VU \textendash\ VP
items, as shown in Table~\ref{coverageftmetrics} and Table~\ref{coveragepfnmetrics}.
The exemplar approach, on the contrary, does introduce information absent
from the baseline's fulltext training set, which significantly improves coverage metrics:
for argument identification, the percentage of \emph{unseen} frame element labels
goes from 9.6\% down to 2.9\%, and the percentage of \emph{unseen} valence unit
labels goes from 27.2\% down to 13.2\%.
Such improvements in coverage provide a strong hypothesis to account
for the robustness of the exemplar approach over the paraphrastic approach:
reducing the amount of unseen data reduces out-of-vocabulary effects for both
lexical units, frame elements and their syntactic realizations. This concretely
means that the parser will be less likely to be confronted to configurations
that it has not seen before, and therefore more likely to correctly predict
frame element labels, as shown in Section~\ref{measuringimpactexemplar}.
A concrete example of the benefits of the extended coverage of the
exemplar approach is provided by the \fe{Name} frame element, which is
affected almost exclusively by lexical coverage between training and
development sets,
as detailed in Section~\ref{lexicaldiversityerror}.
In this particular case, we found the exemplar approach to improve $F_1$ score by almost 6 points, when
the paraphrastic approach marginally improved it by about 0.6 point of $F_1$
score.

\begin{table}[h!tpb]
 \begin{minipage}[t]{.48\linewidth}
  \centering
  \resizebox{\linewidth}{!}{
   \begin{tabular}{lrrr}
    \toprule
    \multicolumn{4}{c}{\textsc{Coverage metrics on FT}}\\
    \midrule
                     & train           & dev             & overlap \\
    \midrule
    Lexical Units    & \numprint{3504} & \numprint{1778} & 74.2\%  \\
    Frame Elements   & 690             & 511             & 90.4\%  \\
    Valence Units    & \numprint{2964} & \numprint{1668} & 72.8\%  \\
    Valence Patterns & \numprint{5245} & \numprint{2165} & 33.6\%  \\
    \bottomrule
   \end{tabular}
  }
  \caption{Overlap between training and development sets, with a training
  set composed of FrameNet fulltext data}
  \label{coverageftmetrics}
 \end{minipage}%
 \hspace{.04\linewidth}
 \begin{minipage}[t]{.48\linewidth}
  \centering
  \resizebox{\linewidth}{!}{
   \begin{tabular}{lrrr}
    \toprule
    \multicolumn{4}{c}{\textsc{Coverage metrics on FT+EX}}\\
    \midrule
                     & train            & dev             & overlap \\
    \midrule
    Lexical Units    & \numprint{9953}  & \numprint{1778} & 88.2\%  \\
    Frame Elements   & \numprint{1031}  & 511             & 97.1\%  \\
    Valence Units    & \numprint{6468}  & \numprint{1668} & 86.8\%  \\
    Valence Patterns & \numprint{35131} & \numprint{2165} & 48.8\%  \\
    \bottomrule
   \end{tabular}
  }
  \caption{Overlap between training and development sets, with a training
  set composed of FrameNet fulltext and exemplar data}
  \label{coverageftexmetrics}
 \end{minipage}
\end{table}

\begin{table}[h!tpb]
 \centering
 \begin{tabular}{lrrr}
  \toprule
  \multicolumn{4}{c}{\textsc{Coverage metrics on FT + PFN}}\\
  \midrule
                   & train           & dev             & overlap \\
  \midrule
  Lexical Units    & \numprint{3504} & \numprint{1778} & 74.2\%  \\
  Frame Elements   & 690             & 511             & 90.4\%  \\
  Valence Units    & \numprint{2964} & \numprint{1668} & 72.8\%  \\
  Valence Patterns & \numprint{5245} & \numprint{2165} & 33.6\%  \\
  \bottomrule
 \end{tabular}
 \caption{Overlap between training and development sets, with a training
  set composed of FrameNet fulltext data and \pfn XP\#30 data}
 \label{coveragepfnmetrics}
\end{table}

\subsection{Distribution}
\label{errordistribution}

In light of the coverage results presented in Section~\ref{errorcoverage},
we also questioned whether the differences of performances
observed in Section~\ref{measuringimpactexemplar} could be explained in
terms of the
respective impact of both exemplar and paraphrastic
approaches on data distribution in the training set.

\begin{figure}[h!tpb]
 \begin{minipage}[t]{.48\linewidth}
  \centering
  \resizebox{\linewidth}{!}{
    \includegraphics{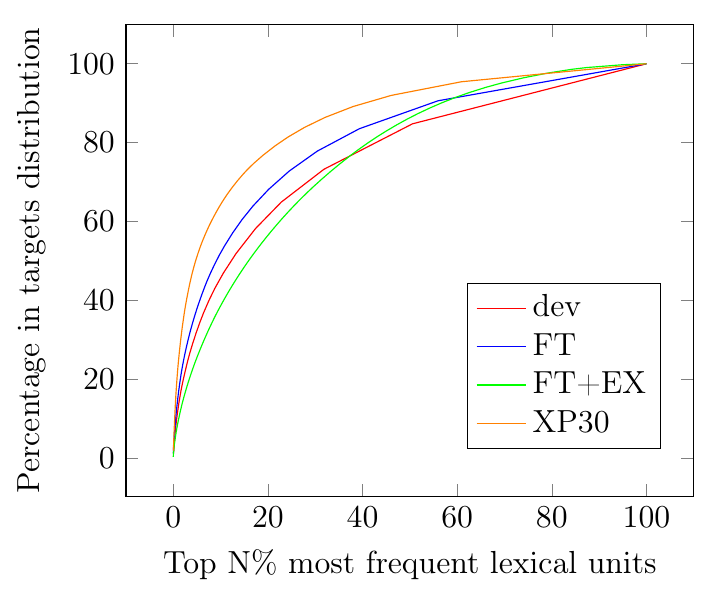}
  }
  \caption{Distribution of lexical units in development (dev) set, training
   set with fulltext (FT), with fulltext and exemplar (FT+EX) and with \pfn XP\#30 (XP30)}
  \label{distributionoflexicalunits}
 \end{minipage}%
 \hspace{.04\linewidth}
 \begin{minipage}[t]{.48\linewidth}
  \centering
  \resizebox{\linewidth}{!}{
    \includegraphics{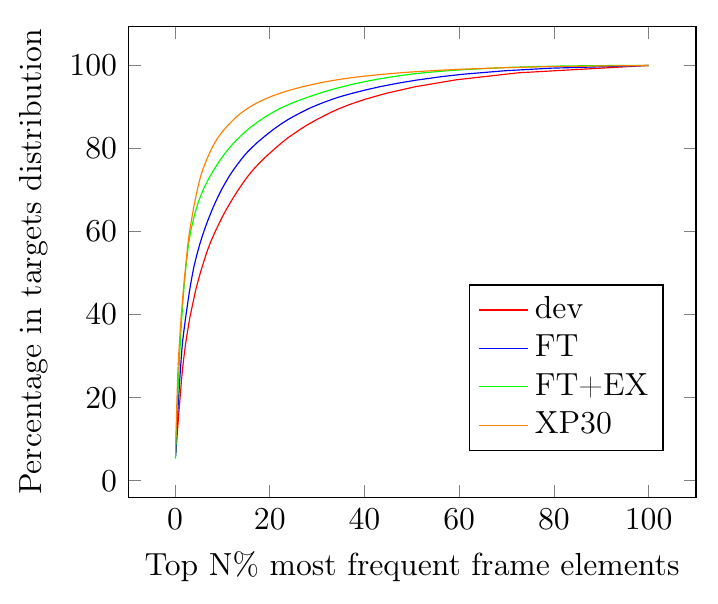}
  }
  \caption{Distribution of frame elements in development (dev) set, training
   set with fulltext (FT), with fulltext and exemplar (FT+EX) and with \pfn XP\#30 (XP30)}
 \end{minipage}
\end{figure}

\begin{figure}[h!tpb]
 \begin{minipage}[t]{.48\linewidth}
  \centering
  \resizebox{\linewidth}{!}{
    \includegraphics{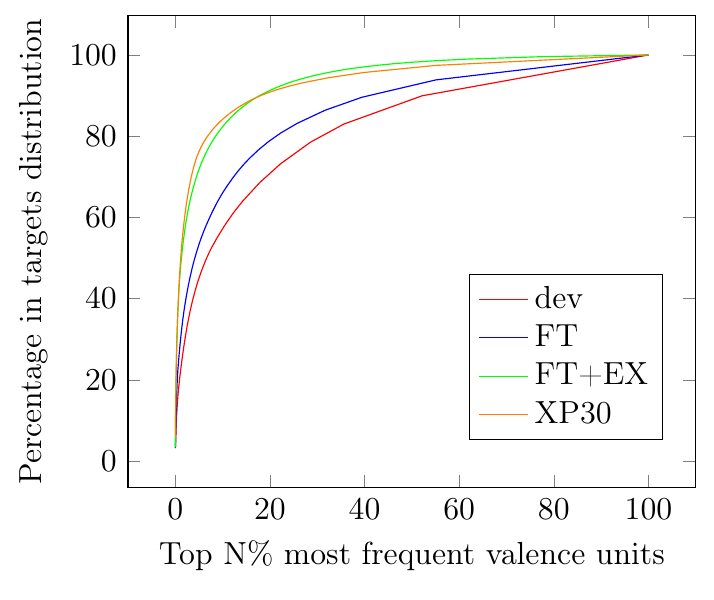}
  }
  \caption{Distribution of valence units in development (dev) set, training
   set with fulltext (FT), with fulltext and exemplar (FT+EX) and with \pfn XP\#30 (XP30)}
 \end{minipage}%
 \hspace{.04\linewidth}
 \begin{minipage}[t]{.48\linewidth}
  \centering
  \resizebox{\linewidth}{!}{
    \includegraphics{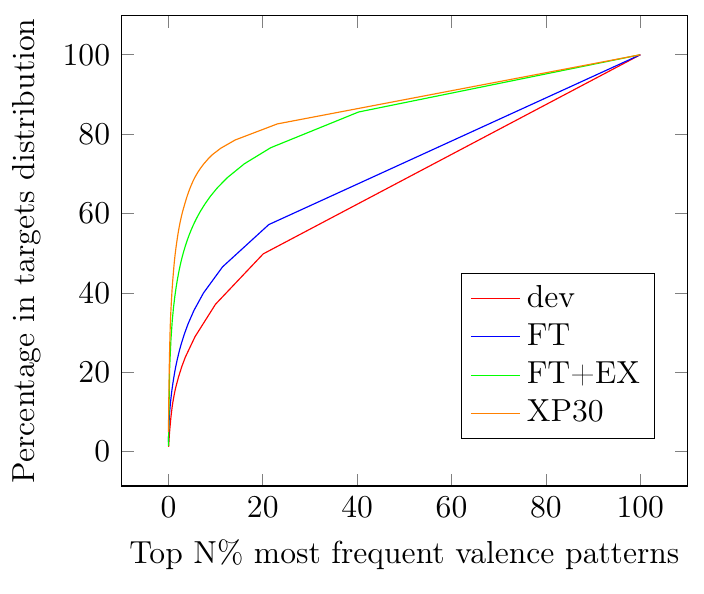}
  }
  \caption{Distribution of valence patterns in development (dev) set, training
   set with fulltext (FT), with fulltext and exemplar (FT+EX) and with \pfn XP\#30 (XP30)}
  \label{distributionofvalencepatterns}
 \end{minipage}
\end{figure}

Results, presented in Table~\ref{distributionoflexicalunits} to
Table~\ref{distributionofvalencepatterns} confirmed, in a
more systematic fashion than previously described in
Section~\ref{errorfrequencyeffects},
that the statistical
classifier could be considered robust to frequency effects in the data.
Indeed, those tables show that both exemplar and paraphrastic
approaches have a tendency to \emph{zipfianize} data distribution in the
training set, by making frequent elements exponentially more frequent. This
makes for data distributions in the training set
diverging significantly from that of the
development set. Nonetheless, those diverging data distributions did not
prevent the statistical classifier from extracting systematic lexico-syntactic
patterns, independent of LU - FE - VU - VP global frequencies, as shown
in Section~\ref{measuringimpactexemplar}.

\chapter{Discussion}
\label{discussion}

\section{Impact of syntactic preprocessing}
A major outcome of this work is the qualitative confirmation of the
predominant impact of syntactic preprocessing on the performances
of argument identification. Although the 80\% cap on recall has been
a well-known
limitation of the argument span identification algorithm
\citep{dasetal2014,tack2015}, question remained as to whether
syntactic challenges faced by frame semantic parsing
were circumscribed to the approximations of the algorithm.
Indeed, the underlying assumption of the argument span identification
algorithm \textendash\ to restrict multi-token
spans to all terminal nodes of a same parent node in a dependency tree
\textendash\ could
only be characterized as an \emph{approximation} once confronted to the
syntactic representation
produced by the dependency parser.
After all, if possible argument spans were
not retrieved by the heuristic algorithm, it was as much due to
the algorithm's presuppositions than to
the dependency representations output by the syntactic parser, failing to
match all the (syntactic) structures of the arguments in FrameNet.
Therefore, beyond the assumptions of the argument span identification
algorithm, the question of the core source of those
syntactic representation mismatches
turned out to be crucial: were syntactic representation misalignments due to
discrepancies between the underlying syntactic formalisms of FrameNet and
dependency parsers, or were they due to erroneous predictions
from those dependency parsers, generating syntactic representations
deviating from the expectations of their theoretical frameworks?
Our results tend to show that it is probably a bit of both.

In our replication study presented in Section~\ref{impactdepparser},
we showed that using
the state-of-the-art BIST dependency parser~\citep{kiperwasser2016}
improved argument identification
performances by at least 1 point of $F_1$ score over
previous approaches relying on the ten-years-older MSTParser~\citep{mst2006},
although both syntactic parsers relied on the same syntactic formalism and
the exact same training data.
Those results provide experimental support for hypothesizing strong correlations
between the quality of syntactic parsers' predictions and the ability
of frame semantic classifiers to extract useful patterns to better predict
argument roles and spans, independently
of the consistency of the syntactic formalisms used by FrameNet and dependency
parsers.

Additionally, in Section~\ref{syntacticchallenges} of
our error analysis, we showed that deep theoretical discrepancies persist
between what FrameNet and dependency parsers consider
to be syntactic constituents, and that, in several specific
syntactic configurations, those discrepancies clearly
prevent frame semantic classifiers from properly identifying argument spans
and extracting relevant patterns, regardless of the accuracy of
syntactic predictions. We notably showed that misalignments between
syntactic representations prove particularly detrimental to compensate
for high lexical diversity when classifiers cannot make use of syntactic cues
consistent with FrameNet annotation.

On both accounts, recent syntax-free and joint syntax-semantic
approaches to frame semantic parsing look promising \citep{swayamdipta2017}.
However, even most recent approaches using multitask learning with syntactic
scaffolding still rely on different syntactic formalisms than that of FrameNet.
As those models consistently yield better results than
syntax-free models, it may be worth exploring  in future work
joint learning of both FrameNet semantic and syntactic representations.

\section{Consistency of FrameNet annotation}
Another major outcome of this work deals with judging the
consistency of FrameNet annotation for the frame semantic parsing task.
In Chapter~\ref{results} we introduced the results of our paraphrastic
data augmentation system and showed that, contrary
to expectations, a frame semantic parser trained on an artificially
augmented dataset, containing data generated by a valence pattern matching
algorithm, significantly degraded performances when not relying
on any kind of paraphrastic candidates filtering. This result strongly
calls into question the consistency of FrameNet annotation, given that
frame semantics theory predicts that
a valence pattern matching algorithm should generate syntactically compatible
and semantically related clauses.
Additionally, we showed that the negative impact on performances
could be compensated by filtering paraphrastic candidates,
even randomly, to the point where we even observed marginal improvements on
argument identification,
though not statistically significant. At the time we proposed
two hypothesis to account for the impact of (random) paraphrastic candidates
filtering on argument identification: (1)
that there existed a latent variable, not necessarily
semantically motivated, or at least not in ways distributional models could
capture, that determined, among the set of unfiltered \pfn candidates, what
was noise from what was not; or (2) that data generated by \pfn were actually
always somehow noisy, but that the \rofames parser was robust enough
that it could compensate for the bias introduced by noisy artificial data,
provided that those noisy data remained limited in size.
The results of our error analysis, detailed in Section~\ref{hardframeelements}
and Section~\ref{compensatinglexical},
show once again that both hypothesis may hold some truth.

First of all, we demonstrated in Section~\ref{hardframeelements} that
frame elements in FrameNet encompassed different degrees of semantic
and syntactic specificity. Phrased differently, we could say that FrameNet
includes a great variety of \emph{classes} of different \emph{scope}. We showed,
in the same section, how, at the end of the spectrum, non-core frame elements
constituted some of the hardest classes to predict,
due to the great variety of \emph{domains} \textendash\
defined as specific syntactic and semantic configurations \textendash\
that those classes could cover.
Those considerations translate directly
into differences of scope at the scale of the valence pattern.
As such,
our valence pattern matching algorithm generates paraphrastic candidates which can
turn out to be quite \emph{far} from the original target. In short,
the \emph{scope} of a given valence pattern could constitute the
\emph{latent variable}
that determines whether or not a given lexical combination output by
our algorithm is \emph{likely} or not, and whether or not it should
be considered as \emph{noise}. Future work, experimenting on keeping
narrow-scope paraphrastic candidates only, could potentially shed light on this
hypothesis.

Second, we showed in Section~\ref{compensatinglexical} that the \rofames
parser was robust to noisy data in that it used syntactic cues to compensate for
high lexical diversity. This experimental results tend to demonstrate that
the noise generated by our paraphrastic data augmentation approach does not
evenly affect argument identification, and does so negatively only when
the parser
is unable to rely on syntactic patterns to compensate for the noisy data.

It may be worth noting that the paraphrastic data augmentation approach
introduced in this work
remained very preliminary, in that it was limited to introducing
lexical diversity at the level of predicate-argument combinations. Work from
\cite{hasegawaetal2011} suggest several ways for incorporating
syntactic diversity to our approach via FrameNet-internal knowledge,
through, e.g., voice and
perspective alternation, or NP\textendash PP\textendash VPing clauses
transformation. As such, it provides interesting developments for testing further
the robustness of FrameNet annotation.

\section{The more the merrier?}
In this work we confirmed two well-known facts of the semantic role labeling
literature:
(1) that adjuncts are usually significantly harder to predict that core
arguments; and (2) that core arguments predicted with high precision and recall
exhibit strong bias toward specific syntactic patterns\footnote{such
 as \fe{agents} frequently realized as subjects of noun phrases}
\citep[for both, see Table 13 in][]{propbank2005}.
Our work suggests that the high number of classes in FrameNet may actually
be an asset, as it could provide just the right number of classes for a
statistical classifier to make
interesting generalizations without overfitting or being unable to
extract discriminant systematic patterns from too generic classes.

The poor performances
of the \rofames parser on non-core frame elements suggest that it may even be
needed to \emph{increase} the number of those frame elements by splitting
them into subclasses better able to characterize semantic and syntactic phenomena
in specific domains. This intuition goes against past work in frame semantic
parsing, notably that of \cite{matsubayashi2014}, which attempted to reduce
the complexity of the frame semantic parsing task by reducing the number
of frame elements to be predicted.

Rather than manually increasing the number of classes in FrameNet, it may
prove beneficial to make use of the literature on latent variable
grammars \citep[e.g.][]{petrovetal2006,petrov2010} for learning
latent sub-classes of the gold meta-classes in an unsupervised fashion via
split-and-merge algorithms.

However, increasing the number of classes, even latently, raises the question
of whether FrameNet data would contain enough representative samples for
a statistical model to
appropriately extract generic representations for each class.
The question is all the more relevant that recent neural
network models for frame semantic parsing are designed to learn multiple
complex representations at once, while relying on a limited set of
features. Would those models prove capable of learning
infrequent phenomena and compensating for the lack of syntactic cues provided
by dependency parsers? Beyond the question of the sensitivity of deep
learning models to frequency
effects lies the question of their ability to learn from tiny data.
The fact that the syntax-free model proposed
by \cite{swayamdipta2017} achieves comparable results than the previous model
of \cite{dasetal2014}, while not being bounded by the approximations of
the argument span identification algorithm, suggests that the model may
fail to properly extract certain patterns when dispensing with syntactic
preprocessing.
To further argue on the matter, more research is needed to qualitatively
understand the performances of recent
neural network models for frame semantic parsing.

\section{Learning conceptual representations from limited data}
Recent neural network approaches to frame semantic parsing
\citep{fitzgerald2015,roth2016,swayamdipta2017,yangmitchell2017} rely
on a similar framing of the statistical learning process:
from a limited set of distributional input features,
a neural network creates and enriches, via each layer,
distributional representations of frame semantic structures.
Given the hard constraint posed by the size of the training data and
the information bear by input features, the challenge
for the model is to converge to distributional representations rich enough
to produce accurate predictions of frame semantic structures. Considering
the learning problem at hand, we propose two conceptual approaches to
improving the model:
one is to \emph{reduce the distance} between input and output
representations by relying on more informative input features, acquired
independently of FrameNet annotation and potentially closer
to the desired output representations; two is to \emph{improve the learning rate}
by guiding the model and constraining its representations \emph{at each
layer} of the network.

The first approach follows from observations made throughout this work:
in order to appropriately learn frame semantic structure representations,
statistical models need to be able to capture a whole range
of linguistic phenomena and handle a great variety of specific
problems. Those include, but are not restricted to,
being able to: handle out-of-vocabulary and out-of-domain
lexical items (see Section~\ref{lexicaldiversityerror}),
model semantic properties which
condition the realization of arguments (such as \emph{animateness}, see
Section~\ref{thecaseforcase}), operate subtle semantic distinctions
(see Section~\ref{semanticconsiderations})
or appropriately resolve coreference when needed
(see Section~\ref{necessaryapproximations}), not
to mention being able to model
the constraints which condition the syntactic realization of the
arguments of predicates (see Section~\ref{syntacticchallenges}).
If the distributional representations of words based on contextual information
on which rely most neural network
models for frame semantic parsing have proven quite successful at
handling out-of-vocabulary items \citep{rothlapata2015}, they still fall
short of encompassing the depth of information covered by the aforementioned
cases. The literature suggests many possible leads to enrich those contextual
representations, be it via the modeling of proper names \citep[e.g.][]{herbelot2015}
useful for identifying named entities and their (semantic) properties,
or via the incorporation of syntactic information \citep[e.g.][]{levygoldberg2014},
the integration of those representations to models of
formal semantics \citep[e.g.][]{erk2016} or the modeling of semantic relations
between those representations \citep[e.g.][]{hendersonpopa2016,rollererk2016},
all of which could contribute to reducing the amount of information statistical
models of frame semantic parsing have to extract from limited gold data alone.

The second approach follows from the observation than motivated this
work in the very first place: as previously detailed in Section~\ref{richannotation},
The FrameNet taxonomy is a well-structured resource
which contains rich annotation not necessarily readily available
at the surface (sentence) level, but which is nonetheless required for
statistical models to learn appropriate generalization principles.
Consider the following example from \citep{hasegawaetal2011}:
\begin{examples}
 \item \feanno{They}{Authorities} are going to \target{incarcerate} \feanno{him}{Prisoner}
 \item \feanno{They}{Agent} are going to \target{confine} \feanno{him}{Theme} \feanno{to prison}{Holding\_location}
\end{examples}
Those two sentences, which are considered as paraphrase of each other,
exemplify many semantic distinctions:
the \lexunit{incarcerate}{v} lexical unit of the \semframe{Imprisonment} frame
is more specific than the
\lexunit{confine}{v} lexical unit of the the \semframe{Inhibit\_movement} frame,
in that it implies that the \fe{Holding\_location}
is a \fe{Prison}. The \fe{Prison} frame element
not being \emph{required} to form a minimal clause in the
\semframe{Imprisonment} frame,
is hence characterized as \emph{non-core}.
However, acquiring the kind of combinatorial constraints such
as \semframe{Imprisonment} $=$ \semframe{Inhibit\_movement} $+$ \fe{Holding\_location}
$as$ \fe{Prison} characterized in FrameNet by frame and frame element relations
has proven particularly hard to acquire from textual data alone, and more
specifically from distributional (contextual) representations of words
\citep{botschenetal2017}. However, such informative constraints could be
incorporated higher-up in the neural network layered architecture, by
retrofitting intermediate representations of frame semantic structures
following \cite{faruquietal2015}, or designing hybrid models constraining
intermediate representations via hard-coded rules.

\chapter{Conclusion}
\label{conclusion}

In this work, we have reported several important results which
we hope will contribute to more systematic error analysis of statistical
models for automatic frame semantic structure extraction.

Our first contribution is the \emph{replication} of several past studies on
frame semantic
parsing, relying this time on the most recent FrameNet 1.7 data release.
We proposed more robust and larger development and testing sets containing
no duplicate annotation sets, and showed that, when tested on those new data sets,
the benefits of past features,
such as exemplar or hierarchy, are not as significant as originally
reported in \citep{kshirsagaretal2015} as they contribute to a 2 $F_1$ gain
rather that 4 $F_1$ gain on argument identification with gold frames.
Additionally, we showed that preprocessing toolkits play a crucial role
in the performance of statistical models for argument identification,
and that differences in preprocessing setups can lead to nearly 3 $F_1$
points differences on argument identification scores. Such variations, being
of the same order of magnitude than the gains reported by recent statistical
models for frame semantic parsing \citep{roth2016}, strongly call into question
the real benefits of those models which rely on different preprocessing
toolkits than their baselines.

Our second contribution deals with the \emph{smartness} and
\emph{robustness} of the \rofames statistical
classifier, originally proposed by \citep{dasetal2014}.
In order to properly
test the smartness of the classifier, defined as its ability to extract all
useful information for frame semantic structure extraction when exposed to
a given set of data, we proposed to train it
on an artificially augmented dataset, generated using a rule-based model
combining valence pattern matching and lexical substitution on the original
FrameNet gold training set.
We showed that the \rofames parser proves relatively \emph{smart}, as
we saw no statistically significant increase in performances
in our best data augmentation setup,
but not necessarily \emph{robust} to noisy data,
as performances significantly decreased when the parser was exposed to
unlikely lexical configurations of
predicate-argument structures.
We showed, moreover, that the quality of syntactic preprocessing
plays a major
role in the performances of the classifier on argument identification,
especially for argument span identification. We discussed how the syntactic
representation mismatches between FrameNet and dependency parsers are as
much due to erroneous predictions from the dependency parsers than to
divergence in FrameNet and dependency parsers' underlying syntactic formalisms.
Additionally, we hypothesized that the varying syntactic and semantic scopes
of the different classes of frame elements could act as a latent variable
for determining noisy paraphrastic data, and demonstrated that classes
which prove hardest to predict usually exhibit a great variety of
semantic and syntactic realizations.\\
\noindent
We had originally asked the following questions in introduction:
\begin{enumerate}
 \item Are prediction failures due to a
       \emph{lack of annotation}, which translates to the absence in the
       training data of linguistic patterns necessary for \emph{any}
       machine learning model to properly predict the frame semantic
       structures included in the evaluation data?
 \item Does failure to capture structural generalization
       principles necessary to correctly predict frame semantic structures
       lie with the probabilistic models used so far?
\end{enumerate}
We can now conclude with the following answers:
\begin{enumerate}
 \item If the lack of annotation in FrameNet does account for certain
       failures in frame semantic structure predictions,
       especially
       for out-of-domain named entities and lexical items, strong limitations
       are also posed to
       machine learning models by FrameNet's formalism and its definition of
       large-scope classes, especially for non-core frame elements,
       which prevent classifiers from extracting systematicity for those
       specific classes.
 \item Strong conceptual limitations do indeed prevent statistical models
       for frame semantic parsing to properly extract useful generalization
       principles,
       especially for identifying argument spans or processing
       fine-grained semantic distinctions at the argument level.
\end{enumerate}

Our work suggests possible new concrete leads for improving statistical models
of frame semantic parsing: it seems clearer and clearer that frames and
arguments should be predicted jointly, and that so should (frame) semantic
and syntactic representations. Recent approaches relying on neural network
architectures also suggest that statistical classifiers should rely on richer
input representations of words, better able to characterize deep semantic
properties, coreferences, semantic relations and syntactic constraints, and
on latent representations of subclasses, better able at characterizing
the variety of syntactic and semantic configurations of non-core
frame elements.

\begin{appendices}
 \chapter{Frame Semantic Parsing History: Detailed Overview}
 \label{detailedoverviewfsparsing}

 In the following sections we provide a detailed
 historical overview of
 supervised and semi-supervised methods for frame semantic parsing.
 For practical reasons, we focus on work which relied on the
 FrameNet 1.5 dataset and postdated the seminal work of~\cite{dasetal2014}.
 We make two important exceptions:
 we mention the work of \cite{srlgildeajurafsky2002},
 given that they were the very first to propose a statistical approach
 to frame semantic parsing, although they focused exclusively on argument
 identification given gold targets and gold frames and relied on a
 very small FrameNet dataset which predated the 1.3 release.
 We also mention the work of \cite{johansson2007},
 as their system achieved the best results on the
 SemEval 2007 shared task 19 and was the first to provide a full frame
 semantic parsing pipeline comprising target, frame and argument identification.

 \section{\cite{srlgildeajurafsky2002}}

 \cite{srlgildeajurafsky2002} pioneered the task of argument
 identification\footnote{which they referred to as \emph{semantic role labeling}
 and included both arguments span identification and arguments role labeling}
 with FrameNet, given gold targets and gold frames.
 They used a pre-1.3 FrameNet dataset
 which comprised about 50,000 annotated sentences.
 Their system relied on discriminative models which made use of both lexical
 and syntactic features such as tokens, part-of-speech tags, dependency paths,
 tokens positions, and voice. Most subsequent work on semantic role labeling
 in general made use of their original features set, often enriched with
 additional features.
 \cite{srlgildeajurafsky2002} achieved 82\% accuracy on
 labeling pre-identified constituents and a
 63\% $F_{1}$ score on the full argument identification task measured on
 individual frame elements. Per sentence, the system achieved 38\% (0.38) accuracy,
 significantly lower than the 0.66 \textendash\ 0.82 inter-annotator agreement
 (which usually varies depending
 on the predicate).\footnote{\cite{srlgildeajurafsky2002} also reported
  the kappa statistic, but it has been shown
  to be inadequate for judging inter-annotator agreement on frame semantic annotation,
  as, contrary to the usual setting where the kappa statistic is used,
  FrameNet annotators do not have to choose among
  a fixed pool of label for each annotated instance. See~\citep[note n.2]{burchardt2006}
 for details}
 This difference between parsers accuracy scores and inter-annotator agreement
 metrics makes semantic role labeling in general, and frame semantic parsing
 more specially, a \emph{hard} task in computational linguistics and natural
 language processing.

 \section{\cite{johansson2007}}

 \cite{johansson2007} provided the first full system for frame semantic parsing to
 include target identification, frame identification and argument identification.
 They achieved the best performance at the SemEval 2007 shared
 task 19.

 For target identification, their system consisted of a set a heuristic
 rules.

 For frame identification, they used a SVM classifier to disambiguate
 polysemous lemmas and assign the correct frame when multiple options where
 possible. They also extended the vocabulary of frame-evoking words with WordNet
 to handle unknown lexical units. They then used a collection of separate SVM
 classifiers \textendash\ one for each frame \textendash\ to predict a single
 evoked frame for each occurrence of a word in the extended vocabulary.

 For argument identification, they divided the task into two subtasks and solved
 the two tasks sequentially. They first identified candidate spans by classifying
 them using SVMs. They then assigned frame element labels to the identified
 arguments using SVMs as well.

 On the SemEval 2007 test set, they achieve 76.10\% $F_1$ score at target
 identification,
 57.34\% $F_1$ score at exact frame identification with predicted targets,
 and 42.01\% $F_1$ score at exact argument identification with predicted targets
 and frames.

 \section{\cite{dasetal2014}}

 \cite{dasetal2014} improved the results of \cite{johansson2007}
 on all tasks of the SemEval shared task, and provided a long-standing
 state-of-the art system on FrameNet 1.5 data, which was used as a baseline
 by most subsequent work on frame semantic parsing.

 On target identification, they modified the set of heuristic rules of
 \cite{johansson2007} and improved the $F_1$ score by about 3 points.

 On frame identification, they used a discriminative probabilistic
 conditional log-linear model based on latent variables incorporating
 lexical-semantic features. They expanded this model with a similarity graph
 computed with WordNet to handle frame identification for unseen lexical units.
 They improved the joint score on target and frame identification of
 \cite{johansson2007} by about 4 points of $F_1$ score.\footnote{Note that,
  as \cite{johansson2007}
  did not provide scores for frame identification with gold targets, it is difficult
  to evaluate the contribution of the frame identification system of \cite{dasetal2014}
 in itself}

 On argument identification, they treated jointly the two tasks or argument
 span identification and frame element labeling. They derived the set of spans
 with a high-recall rule-based
 algorithm that looks at the dependency syntactic context of the predicate word.
 They then used a conditional
 log-linear model over spans for each role of each evoked frame, which was
 trained using maximum conditional log-likelihood. Their model used a significant
 number of features ranging from lexical items combinations to syntactic
 dependency paths (See Figure~\ref{semaforfeatures} for a full account of their features).
 They used beam search for
 decoding to prevent arguments overlap. They also proposed a system which
 modeled argument identification as constrained optimization solved dynamically
 with dual decomposition \citep{martinsdualdecomp2011}, but which did not yield
 significant improvements.\footnote{
  Their approach did form however the basis of subsequent successful work, see
  \citep{hermann2014,tack2015}}
 They improved the full score of argument identification with predicted frames
 and predicted targets by about 4.5 points of $F_1$ score compared to
 \cite{johansson2007}.

 \section{\cite{hermann2014}}

 \cite{hermann2014} proposed a model for frame identification based on distributed
 representations of predicates and their syntactic contexts.
 They used the WSABIE algorithm \citep{wsabie2011} to map input and
 frame representations
 to a common latent space during training, where the distance between the latent
 representations of the target predicate and the correct frame is minimized while
 the distance between the latent representations of the predicate target and all
 other frames is maximized.
 Decoding is performed using the maximal cosine similarity between candidate
 frame representations
 and the latent representation of the target predicate.
 They improved frame identification scores with gold targets by about 5 points
 of $F_1$ score compared to the baseline of \cite{dasetal2014}.

 For argument identification they relied on \cite{dasetal2014}'s log-linear
 model but formalized the problem as constrained optimization with three ILP
 constraints:
 \begin{enumerate}
  \item each span could only have one role;
  \item each core role could be present only once;
  \item all overt arguments had to be non-overlapping.
 \end{enumerate}
 They solved the problem with an off-the-shelf ILP solver,
 which is not specified, and which performances could explain why they achieve
 better results than \cite{dasetal2014}'s original ILP-based system
 and \cite{kshirsagaretal2015}'s replication results when
 using \cite{hermann2014}'s frame identification system.
 They similarly improved argument identification scores with predicted frames
 by about 5 points of $F_1$ score over the baseline of \cite{dasetal2014}.

 \section{\cite{tack2015}}

 \cite{tack2015} used the same ILP formulation of argument identification as
 \cite{hermann2014}, but proposed to solve it with a custom dynamic programming
 algorithm instead of an off-the-shelf ILP solver.
 They achieved better performance compared to the previous approaches, with an
 improvement of about 0.4 $F_1$ score with predicted frames compared to
 \cite{hermann2014}.

 \section{\cite{kshirsagaretal2015}}

 \cite{kshirsagaretal2015} were the first to try and train a parser based on
 that of \cite{dasetal2014}
 on both fulltext and exemplar data since \cite{dasetal2014} reported that
 training on both fulltext and exemplar sentences hurt performances on the
 SemEval dataset.
 They used the model of \cite{dasetal2014}, extracting candidate spans with
 heuristics, enforcing non-overlapping constraints on arguments and decoding
 with beam search.

 They actually showed that, on the FrameNet 1.5 dataset,
 training on both fulltext and exemplars improved $F_1$ score
 on argument identification with gold frames by nearly 3 points.
 Following the remark of \cite{dasetal2014} which argued that the damage in
 performance could be due to the difference of domains between fulltext and
 exemplar data, they also experimented training on both fulltext and
 exemplars with \emph{domain adaptation} \citep{daume2007}. This allowed
 features to be conditioned on the domain (here, fulltext or exemplar),
 as, in domain adaptation,
 each feature has a domain-specific weight and a global weight, which allows
 systems to better adjust to specific domains while not loosing generalization
 power. They found however a marginal 0.3 points $F_1$ score
 improvement over training on both fulltext and exemplar annotations without
 domain adaptation.

 They also proposed to incorporate \emph{hierarchy features}, in this case
 Inheritance and SubFrame relations, in order to capture statistical generalizations
 about the kinds of arguments seen in frame elements.
 Finally, they suggested using \emph{guide features}, in this case PropBank
 annotated data, to overcome the limitations posed by the sparcity of FrameNet
 annotated data. Using Propbank as a guide feature consisted in using
 predictions done on the FrameNet dataset by a model trained on PropBank data
 as additionnal features when training the model on FrameNet data.
 Guide features, however, did not yield significant results.

 Their best system, incorporating the hiearchy feature and trained on both
 fulltext and exemplar data, achieved a 4 $F_1$ gain over
 the baseline of \cite{dasetal2014}.

 \section{\cite{rothlapata2015}}

 \cite{rothlapata2015} were the first to incorporate sentence
 and discourse context
 features motivated by linguistic considerations inspired by~\cite{fillmore1982}.
 They proposed to incorporate the following key features:
 \begin{enumerate}
  \item a set of features modeling document-specific aspects of word
        meaning using distributional semantics. They relied on distributional word
        representations adapted to documents content;
  \item a feature modeling previous role assignment mentioned in discourse,
        acquired by resolving coreferences on frame element semantic types with the Stanford
        Coreference Resolution system \citep{leeetal2015};
  \item a feature derived from automatically computed
        coreference chains, which helped determine which frame element labels
        were likely to be assigned to new entities. For example, the \fe{Result}
        of a \fe{Causation} is more likely to be discourse-new than the \fe{Effect}
        that caused it;
  \item a set of constraints on roles, as defined in previous studies,
        but exploited using a re-ranking algorithm.
 \end{enumerate}
 Their system achieved nearly 1 point $F_1$ gain in argument identification
 with gold frames and 0.7 $F_1$ point gain in argument identification with
 predicted frame.

 \section{\cite{fitzgerald2015}}

 \cite{fitzgerald2015} proposed the first neural model for frame semantic parsing.
 Their model embedded candidate arguments and frame elements for a given frame
 into a shared vector space.
 They used a feed-forward neural
 network to learn correlations between embedding
 dimensions in order to create argument and role representations.
 They then used the dot product between role representations to score possible
 roles for candidate arguments. The decoding process relied on
 a constrained graphical model which jointly modeled the assignment of
 semantic roles to all arguments of a predicate.
 They achieved their best results by training their model in a multitask
 setting on both FrameNet and PropBank data, made possible by the decoupling
 of span and frame-role representations of their model.
 Note that they used the same preprocessing as \cite{tack2015} for fair replication.
 They achieved 70.9\% $F_1$ on argument identification with predicted frames,
 a 0.6 $F_1$ gain over their baseline of \cite{tack2015}.

 \section{\cite{roth2016}}

 \cite{roth2016} proposed a neural frame semantic parser based on \citep{rothlapata2016}.
 They modeled the semantic relationships between a predicate and its arguments by
 analyzing the dependency path between the predicate word and each
 argument head word. They considered lexicalized paths, which they decomposed into
 sequences of individual items, namely the words and dependency relations on a
 path. They defined the embedding of a dependency path to  be  the  final
 memory output state of a LSTM layer that took a path as input,
 with each input step representing a binary indicator for a part-of-speech tag, a
 word form, or a dependency relation.
 They used a state-of-the-art
 dependency parser \citep{kiperwasser2016} for preprocessing, and achieved
 70.0\% $F_1$ on argument identification with predicted frames, 0.9 $F_1$ points
 short from \cite{fitzgerald2015}.

 \section{\cite{swayamdipta2017}}
 \cite{swayamdipta2017} proposed four different neural models for frame semantic
 parsing.

 They first proposed a syntax-free model called the \emph{softmax-margin SegRNN},
 based on the \emph{SegRNN} of \cite{kongetal2016},
 and which uses a combination of
 bidirectionnal RNNs with a semi-Markov CRF with a slight modification to
 favor recall over precision.
 Their \emph{softmax-margin SegRNN} model learns representations of targets,
 lexical units, frames and
 frame elements. It achieves 66.4\% $F_1$ on argument identification
 with gold frames, and 69.9\% $F_1$ on argument identification with
 predicted frames using the system of \cite{hermann2014}.

 They then proposed three alternative models which make use
 of syntactic information.
 Their first syntactic model adds \emph{dependency features} derived
 from a dependency parsers \citep{andors2016}. It achieves 67.8\% $F_1$
 on argument identification
 with gold frames and 70.6\% on argument identification with predicted frames.
 Their second syntactic model adds \emph{phrase-structure features} derived from
 a state-of-the-art phrase-structure parser \citep{dyeretal2016}.
 It achieves 68.9\% $F_1$
 on argument identification with gold frames and 70.9\% $F_1$ on
 argument identification with predicted frames.

 Finally, authors proposed a model that incorpored their previous syntax-free model
 into a multitask setting where the second task was to identify unlabeled constituents,
 a task which they call \emph{syntactic scaffolding}. Their model achieves 68.3\% $F_1$
 on argument identification with gold frames and 70.7\% $F_1$ on argument identification
 with predicted frames.

 \section{\cite{yangmitchell2017}}
 \cite{yangmitchell2017} proposed multiple neural network models for performing
 both frame identification and argument identification sequentially and jointly.

 For frame identification, they proposed a multi-layer neural model which
 learns to predict a relation between a predicate and a frame given the
 predicate-frame pair and the sentence containing the predicate. Their system
 achieves 88.2\% $F_1$ on frame identification with gold targets,
 comparable to the 88.4\% $F_1$ score of \citep{hermann2014}. However, it
 achieves 75.7\% $F_1$ on ambiguous predicates, a 2.5 $F_1$ gain over
 the baseline.

 For argument identification, they proposed an integrated model which combined
 a \emph{sequential neural model} and a \emph{relational neural model}.
 The sequential neural model formalizes argument identification as a
 word-by-word labeling task where frame elements are encoded using the IOB
 tagging scheme and where learning is performed by a CRF layer on top of
 a DB-LSTM. The benefit of using a neural network architecture is again
 that the LSTM requires limited input features. Those includes only, for each
 word in the sentence, the current word, the predicate, and a position mark that
 denoted whether the current word was in the neighborhood of the predicate.
 The relational neural model formalizes argument identification as multi-class
 classification over pre-identified argument spans. The network learns to
 predict a relation between a predicate and an argument given the predicate-argument
 pair and the sentence that contains it. The inputs of the network are discrete
 features such as words within the argument span, the dependents of
 the argument's head, and their dependency label, predicate word(s), its
 dependents and their dependency labels. These features are mapped to a low
 dimensional space where each input feature is computed into an embedding that
 concatenates average of each feature's embeddings. The feature embeddings
 are then passed on to a non-linear hidden layer and training operates by
 minimizing the negative conditional log-likelihood of the training examples,
 with the conditional probability given by a specific potential function.
 The integrated model is a relational neural model that is learning using the
 knowledge of the sequential model, which learns probabilities for argument
 labels over words instead of spans.

 Joint inference involves jointly performing frame and argument identification.
 To do so, authors relied on standard structural constraints for semantic role
 labeling, which involves avoiding non-overlapping argument spans and repeated
 core roles for each frame. They also introduced two new constraints:
 \begin{enumerate}
  \item one which encodes the compatibility between frame types and semantic
        roles, based on which frame elements belong to each frame;
  \item one which encodes type consistencies of frame element fillers of
        different frames, e.g. the same named entity cannot play both a \fe{Person}
        and a \fe{Vehicule} role. They identify 6 mutually exclusive entity types:
        \fe{Person}, \fe{Location}, \fe{Weapon}, \fe{Vehicule}, \fe{Value} and
        \fe{Time}.
 \end{enumerate}
 They then solved the constrained optimization problem with $AD^3$ \citep{martins2015}.
 Preprocessing, which included part-of-speech tagging and dependency parsing,
 was done with the Stanford CoreNLP toolkit \citep{manningetal2014}.

 On argument identification with gold frames, their integrated model achieves
 65.5\% $F_1$, a 2 points gain over the baseline of \cite{kshirsagaretal2015} at
 63.1\% $F_1$, but still
 lower than the score of \cite{swayamdipta2017} at 68.9\% $F_1$.
 On argument identification with predicted frame, they achieve state-of-the-art
 results with joint inference at 76.6\% $F_1$, way above the previous
 baseline of \cite{fitzgerald2015} at 70.9\% $F_1$.

\end{appendices}

\backmatter

\clearpage

\bibliographystyle{aclnat}
\bibliography{memoir}

\end{document}